\documentclass[11pt, oneside]{article}	
\usepackage{geometry}                		
\usepackage{graphicx}				
\usepackage{authblk}

\usepackage{amssymb}
\usepackage{amsmath}
\usepackage{amsthm}
\usepackage{bm}
\usepackage{hyperref}
\usepackage{color}
\usepackage{float}

\newcommand{\V}[1]{\bm{#1} } 

\newcommand{\mR}{\mathbb{R}}

\newcommand{\lb}{\left(}
\newcommand{\rb}{\right)}

\newcommand{\Req}[1]{eq.\ (\ref{eq:#1})}
\newcommand{\BReq}[1]{Eq.\ (\ref{eq:#1})}
\newcommand{\NReq}[1]{\eqref{eq:#1}}
\newcommand{\Reqs}[2]{eqs.\ (\ref{eq:#1},\ref{eq:#2})}

\newcommand{\NReqs}[2]{\eqref{eq:#1} and \eqref{eq:#2}}

\newcommand{\Reqss}[2]{eqs.\ (\ref{eq:#1}-\ref{eq:#2})}

\newcommand{\Rfig}[1]{Fig.\ \ref{fig:#1}}

\newcommand{\Rfigs}[2]{Figs.\ \ref{fig:#1} and \ref{fig:#2}}

\newcommand{\Lfig}[1]{\label{fig:#1}}
\newcommand{\Leq}[1]{\label{eq:#1}}

\newcommand{\Rsec}[1]{sec.\ \ref{sec:#1}}
\newcommand{\Lsec}[1]{\label{sec:#1}}
\newcommand{\Rapp}[1]{app.\ \ref{app:#1}}
\newcommand{\Lapp}[1]{\label{app:#1}}
\newcommand{\be}{\begin{eqnarray}}
\newcommand{\ee}{\end{eqnarray}}
\newcommand{\subbe}{\begin{subequations}}
\newcommand{\subee}{\end{subequations}}
\newcommand{\no}{\nonumber}

\newcommand{\Ave}[1]{\left\langle {#1} \right\rangle} 
\newcommand{\bs}{\backslash}

\makeatletter
  \newcommand{\subsubsubsection}{\@startsection{paragraph}{4}{\z@}%
    {1.0\Cvs \@plus.5\Cdp \@minus.2\Cdp}%
    {.1\Cvs \@plus.3\Cdp}%
    {\reset@font\sffamily\normalsize}
  }
  \makeatother
\title{Analysis of High-dimensional Gaussian Labeled-unlabeled Mixture Model via Message-passing Algorithm}
\author[1]{Xiaosi~Gu\thanks{Email: gu.xiaosi.26m@st.kyoto-u.ac.jp}}
\author[1]{Tomoyuki~Obuchi\thanks{Email: obuchi@i.kyoto-u.ac.jp}}
\affil[1]{Graduate School of Informatics, Kyoto University, Kyoto 606-8501, Japan}
\begin{document}
\maketitle


\begin{abstract}
Semi-supervised learning (SSL) is a machine learning methodology that leverages unlabeled data in conjunction with a limited amount of labeled data. Although SSL has been applied in various applications and its effectiveness has been empirically demonstrated, it is still not fully understood when and why SSL performs well. Some existing theoretical studies have attempted to address this issue by modeling classification problems using the so-called Gaussian Mixture Model (GMM). These studies provide notable and insightful interpretations. However, their analyses are focused on specific purposes, and a thorough investigation of the properties of GMM in the context of SSL has been lacking. In this paper, we conduct such a detailed analysis of the properties of the high-dimensional GMM for binary classification in the SSL setting. To this end, we employ the approximate message passing and state evolution methods, which are widely used in high-dimensional settings and originate from statistical mechanics. We deal with two estimation approaches: the Bayesian one and the $\ell_2$-regularized maximum likelihood estimation (RMLE). We conduct a comprehensive comparison between these two approaches, examining aspects such as the global phase diagram, estimation error for the parameters, and prediction error for the labels. A specific comparison is made between the Bayes-optimal (BO) estimator and RMLE, as the BO setting provides optimal estimation performance and is ideal as a benchmark. Our analysis shows that with appropriate regularizations, RMLE can achieve near-optimal performance in terms of both the estimation error and prediction error, especially when there is a large amount of unlabeled data. These results demonstrate that the $\ell_2$ regularization term plays an effective role in estimation and prediction in SSL approaches.
\end{abstract}
\noindent\textbf{Keywords:} Statistical inference, Message-passing algorithms, Machine Learning
\section{Introduction}\Lsec{Introduction}
Modern machine learning tasks often achieve high predictive accuracy when large labeled datasets are available. However, in domains like healthcare, specifically in the field of medical diagnosis, the process of collecting labeled data poses challenges as it requires domain expertise and tends to be costly. Conversely, unlabeled data is often abundant in such tasks. Leveraging the information contained in unlabeled data can thus help alleviate the difficulties of obtaining labeled data. Semi-supervised learning (SSL) \cite{zhu2005semi,books/mit/06/CSZ2006} is such a learning framework to leverage unlabeled data in conjunction with a limited number of labeled data. SSL has been already applied in various applications such as image classification \cite{yalniz2019billion} and speech recognition \cite{zhang2020pushing}. Although the effectiveness of SSL has been demonstrated empirically~\cite{berthelot2019mixmatch} and theoretically~\cite{lelarge2019asymptotic} to some extent, it is not fully understood when and why SSL works well.

\par Gaussian mixture model (GMM) \cite{bishop2006pattern} is a parametric probability density function which is widely used in machine learning and statistical analysis, serving as a prototype for classification problems~\cite{lelarge2019asymptotic,tanaka2013statistical,permuter2006study,mai2019high,mignacco2020role}. In classical statistics limit where the sample size tends to be large, a widely adopted approach for estimating the GMM's parameters is maximum likelihood estimation (MLE) \cite{bishop2006pattern}. By utilizing MLE, one can obtain an unbiased and optimal solution, making it a typical strategy for the GMM parameter estimation. However, MLE is not an optimal solution and is biased \cite{sur2019modern} in the high-dimensional statistics limit where the sample size is comparable to or even smaller than the number of parameters. This situation naturally emerges in many contexts, such as imaging \cite{oliver2018realistic} and genomics \cite{libbrecht2015machine}. Recent studies by \cite{belkin2019reconciling} and \cite{mei2022generalization} have discovered that the generalization performance in high-dimensional statistics is different from the predictions of the classical ones. Additionally, in this important limit, GMM is not an overly strong assumption for modeling the aforementioned semi-supervised tasks, because a certain universality occurs under some reasonable conditions~\cite{gerace2022gaussian,montanari2022universality}. These naturally motivate us to study the SSL performance in the GMM framework. 

A number of methods are available for the GMM inference, and one of the important methods is the Bayesian approach. When the assumed model and prior are correct, the Bayesian approach provides the Bayes-optimal (BO) setting where the best performance in terms of the estimation error is obtained, thus providing a natural baseline to be compared with other inference methods. The BO solution is indeed nice, especially in the classification problems based on GMM~\cite{lelarge2019asymptotic,tanaka2013statistical}, but the assumption that the correct model and prior are known, including the hyperparameter values, is unrealistic for many applications. Furthermore, the Bayesian approach frequently suffers from computational complexity since it requires a high-dimensional integration. These present challenges in the Bayesian approach in high-dimensional settings, and an alternative approach, which is tractable in high-dimensional settings but still exhibits near-BO performance, is desired.

Such alternative approaches have been discussed in a number of recent articles~\cite{mignacco2020role,zhang2013non,saade2016clustering,aubin2020generalization}. An interesting observation among them in the context of classification based on GMM is that commonly used convex loss function under appropriate regularizations can yield near-BO performances~\cite{mignacco2020role}. This finding is interesting because near-optimal performance can be obtained by a more practical method than the Bayesian approach. Similar phenomena are observed in different models and problems~\cite{zhang2013non,saade2016clustering,aubin2020generalization}. These analyses are restricted to the case with labeled data or unlabeled data only, and hence it is interesting to see whether these phenomena happen in the case with both labeled and unlabeled data.

With these motivations, in this paper, we conduct a theoretical analysis of the classification problem based on two-class GMM with both labeled and unlabeled data in the high-dimensional setting. A particular focus is put on finding a method/situation achieving
the near-BO performance, but a more comprehensive analysis will be presented. There are some different ways to implement SSL in this problem setting. We concentrate on an $\ell_2$-regularized MLE (RMLE) approach that can naturally treat unlabeled data by regarding the labels as hidden variables. In relation to the Bayesian approach, RMLE can be regarded as a maximum a posteriori (MAP) estimation in the same family of probabilistic models with appropriate priors. The comparison between the Bayesian approach and RMLE thus becomes a comparison of different estimation methods in the common probabilistic structure. Furthermore, a wider estimation framework including MLE, called M-estimation, has recently attracted renewed interest and has been shown to give near-optimal performances in many different contexts~\cite{mignacco2020role,aubin2020generalization,bean2013optimal,donoho2016high,advani2016equivalence,thrampoulidis2018precise}. Hence, RMLE becomes a reasonable choice to consider our problem.

To treat the Bayesian approach and RMLE in a unified manner, we introduce
\be
	p_{\beta}(\bm{w} \mid \mathcal{D}) = \frac{1}{Z_{\beta}}p^{\beta}(\mathcal{D}\mid \bm{w}) p^{\beta}(\bm{w}),\Leq{eq1}
\ee
where $Z_{\beta}$ is a normalization constant, $p(\mathcal{D}|\bm{w})$ is the likelihood describing the data generating process, and $p(\bm{w})$ is the prior distribution for $\bm{w}\in\mR^N$. We call $p_{\beta}(\bm{w}|\mathcal{D})$ $\beta$-posterior which is used in several literatures such as compressed sensing~\cite{montanari2012graphical}. The $\beta$-posterior becomes the original posterior when $\beta=1$ and yields RMLE when $\beta=\infty$, respectively. Handling the $\beta$-posterior is usually difficult in practice, but an efficient algorithm computing the marginals, named approximate message-passing (AMP)~\cite{donoho2009message}, is applicable in our problem setting and is considered to be exact in the high dimensional limit. The AMP performance in this limit is known to be characterized by a simple scalar recursion called state evolution (SE)~\cite{donoho2010message,bayati2011dynamics} that enables to characterize the performance in a clear way. We rely on the $\beta$-posterior, AMP, and SE as the technical basis of our theoretical analysis. 

The remaining part of the paper is organized as follows. In the next section, we summarize the literature of related work, to clarify the difference of the present paper from them\textcolor{black}{, and briefly summarize our main results.} In \Rsec{Formulation}, we present our problem setting: we first define the $\beta$-posterior and introduce the factor graph representation, to employ the BP algorithm; the BP algorithm is approximated by employing the central limit theorem \textcolor{black}{(CLT)} for avoiding the computational difficulty and the AMP algorithm, one of our main workhorses, is derived; the associated SE iteration is also derived. In \Rsec{theoretical_results}, we perform a comprehensive theoretical analysis based on SE: phase transitions characterizing the drastic change of the estimators are investigated and the comparison between the BO and RMLE performances is conducted, to find that RMLE with the optimally tuned regularization yields the very closed performance to the BO one.
The last section is devoted to conclusion.

\section{\textcolor{black}{Related work and main results}}\Lsec{Related}
\subsection{Related work}
Message-passing algorithms enable efficient computations of approximate marginal probabilities for high-dimensional probabilistic models. AMP is a specific message-passing algorithm primarily used for solving high-dimensional estimation problems, and has been especially used in the context of compressed sensing \cite{donoho2009message} and low-rank matrix estimation \cite{matsushita2013low,deshpande2014information}. Notably, AMP is shown to give the optimal performance among many first-order methods in high-dimensional regression and low-rank matrix estimation under a random design assumption~\cite{celentano2020estimation}. The roots of AMP can be traced back to notions of belief propagation (BP)~\cite{yedidia2003understanding} and approximate belief propagation (ABP)~\cite{kabashima2003cdma,tanaka2005approximate}. The key ideas behind AMP were developed in the so-called cavity method in statistical physics \cite{Thouless01031977,mezard1987spin}. The macroscopic dynamics of  AMP algorithms are known to be generally characterized by an iterative equation for a few number of macroscopic parameters that is called SE. The origin of SE can be traced back to density evolution \cite{richardson2001design} that has played a central role in analyzing the performance of BP in low-density parity-check codes.

The unsupervised classification based on GMM in the high dimensional limit has been first studied in~\cite{barkai1994statistical} in the case of two clusters. This analysis was generalized to the case of more clusters in~\cite{lesieur2016phase} using AMP and SE, and a detailed quantitative result revealing the computational and information theoretical limits was obtained. \textcolor{black}{The asymptotic performance of supervised multiclass classification problems based on GMM has been studied in \cite{thrampoulidis2020theoretical,loureiro2021learning}. \cite{thrampoulidis2020theoretical} focused on the square loss with $\ell_2$ regularization, and \cite{loureiro2021learning} extended to arbitrary convex loss and regularization, and examined the impact of regularization on performance.} The case with both labeled and unlabeled data was analyzed in~\cite{lelarge2019asymptotic,tanaka2013statistical} in the BO setting. \cite{tanaka2013statistical} employed the replica method in the case of equal-weight mixtures, uncovering the coexistence of multiple solutions involving a first-order phase transition. \cite{lelarge2019asymptotic} adopted the cavity method to study the same situation as \cite{tanaka2013statistical} and quantified the estimator's performance based on generalization error (GE), finding that a significant accuracy enhancement can be obtained from the unlabeled data. \textcolor{black}{\cite{takahashi2024role} studied a two-class labeled-unlabeled GMM model similar to ours and employed the replica method to derive sharp asymptotic results for self-training, proposing a pseudo-label refinement approach that yields performance close to that of fully supervised learning.} Our study is different from all these results in that a more comprehensive analysis is conducted: both the labeled and unlabeled data are treated in the general imbalance ratio and a comparison between the BO setting and RMLE is performed.

\subsection{\textcolor{black}{Main results}}
\textcolor{black}{The contribution of this paper can be summarized as follows. We derive the AMP iteration formulas for both the RMLE and Bayesian methods and analyze AMP's macroscopic behavior using SE. We analyze SE to construct phase diagrams, examining how various parameters influence these diagrams and their corresponding statistics. We also study the microscopic dynamics of AMP, derive the instability boundary of the AMP algorithm, and present the related phase. Moreover, we compare the two inference methods in terms of mean squared error (MSE) and GE, finding that under the optimal regularizer, the RMLE results nearly match those of the BO, and that the optimal regularizer value becomes finite unlike the fully supervised case \cite{mignacco2020role}. Finally, we investigate the quantitative relationship between the optimal regularizer, the labeled unlabeled ratio, and the class imbalance parameter.}

\section{Formulation}\Lsec{Formulation} 

\subsection{Problem setting}\Lsec{sub_Problem}
Let $\mathcal{D}^l = \{(\bm{x}_\mu,y_\mu)\}_{\mu = 1}^{M_l}$ be the set of labeled data points: the label $y_\mu$ is encoded as $y_\mu \in \{-1,1\}$ and the feature vector $\bm{x}_\mu$ is assumed to be a real $N$-dimensional vector $\bm{x}_\mu \in \mathbb{R}^N$\textcolor{black}{, and the sample size for the labeled data is denoted as $M_l$.} We assume each label is independently and identically distributed (i.i.d.) from the following distribution:
\be
    p(y_\mu) = \rho \delta_{y_\mu,+1} + (1-\rho) \delta_{y_\mu,-1},\Leq{eq2}
\ee
where $\delta_{a,b}$ denotes Kronecker delta and the parameter $\rho\in[0,1]$ controls the balance of the two labels. Given $y_\mu$, the feature vector $\bm{x}_\mu$ is generated by
\be
    \bm{x}_\mu = y_\mu \frac{\bm{w}_0}{\sqrt{N}} + \bm{\xi}_\mu,\Leq{eq3}
\ee
where $\bm{w}_0\in \mathbb{R}^N$ characterizes the cluster center, and $\bm{\xi}_\mu\in\mathbb{R}^N$ is a Gaussian noise vector i.i.d. from $\mathcal{N}(\bm{0}_N, \sigma_0^2 I_N)$ where $\bm{0}_N$ and $I_N$ denote the $N$-dimensional zero vector and identity matrix, respectively. Similarly, let $\mathcal{D}^{u}=\{\bm{x}_\nu\}_{\nu = 1}^{M_u}$ denotes the unlabeled dataset, each component $\bm{x}_{\nu}$ of which is i.i.d. from the same generative model as the labeled one, except that the labels are not observed. Therefore, the total dataset $\mathcal{D} = \mathcal{D}^l \cup \mathcal{D}^u$, the size of which is denoted as $M=M_l+M_u$, obeys the following distribution:
\be
p(\mathcal{D} \mid \bm{w}_0;\sigma_0^2) = &\Bigg[\prod_{\mu\in \mathcal{D}^l}\frac{1}{\sqrt{2\pi \sigma_0^2}}\exp\bigg\{{-\frac{1}{2\sigma_0^2}\Big\Vert \bm{x}_\mu -\frac{y_\mu \boldsymbol{w}_0}{\sqrt{N}}\Big\Vert_2^2}\bigg\}\Bigg]\nonumber
\\
&\times\Bigg[\prod_{\nu \in \mathcal{D}^u}\frac{1}{\sqrt{2\pi\sigma_0^2}}\bigg(\rho\exp\bigg\{-\frac{1}{2\sigma_0^2}\Big\Vert\boldsymbol{x}_{\nu} - \frac{\bm{w}_0}{\sqrt{N}}\Big\Vert_2^2\bigg\}\nonumber \\ & + (1-\rho)\exp\bigg\{-\frac{1}{2\sigma_0^2}\Big\Vert\bm{x}_{\nu} + \frac{\bm{w}_0}{\sqrt{N}}\Big\Vert_2^2 \bigg\}\bigg)\Bigg].\Leq{eq4}
\ee

Based on this generative model, we deal with a problem to estimate $\bm{w}_0$ from the given dataset $\mathcal{D}$. To this end, as described in \Rsec{Introduction}, we consider two approaches: the Bayesian approach and RMLE. For the former one, we introduce a zero-mean Gaussian as the prior distribution:
\be
p(\V{w};\lambda_0)=\sqrt{\frac{\lambda_0}{2\pi}}e^{-\frac{1}{2}\lambda_0 \Vert\V{w}\Vert_2^2}.\Leq{eq5}
\ee
The posterior distribution is thus given by
\be
p(\V{w} \mid \mathcal{D};\sigma^2,\lambda)\propto p(\mathcal{D} \mid \V{w};\sigma^2)p(\V{w};\lambda).\Leq{eq6}
\ee
This becomes the central object for the Bayesian approach. To simplify the situation, we assume that the model parameter $\sigma_0^2$ is known and $\sigma^2=\sigma_0^2$ holds throughout the paper. \textcolor{black}{Since $\lambda_0$ represents the precision of signal and $\sigma^2$ denotes the noise strength, the ratio $1/(\lambda_0 \sigma^2)$ can be interpreted as the signal-to-noise ratio (SNR).} Here $\V{w}$ denotes the parameter variables while $\V{w}_0$ represents the true parameter values; this subscript rule is applied to other parameters such as $\lambda$. For the latter RMLE one, we have a point estimator for $\V{w}_0$ defined by
\be
	\hat{\bm{w}}\lb \mathcal{D};\sigma^2,\lambda \rb = \operatorname*{arg\,min}_{\bm{w}}
	\Bigg[
	-\log p(\mathcal{D} \mid \bm{w};\sigma^2)+\frac{1}{2}\lambda \Vert\V{w}\Vert_2^2
	\Bigg]
=
 \operatorname*{arg\,min}_{\bm{w}}
	\Bigg[ 
	-\log p(\V{w} \mid \mathcal{D};\sigma^2,\lambda)
	\Bigg].\Leq{eq7}
\ee
The aim of this paper is to analyze the property of this estimator and to compare its performance with that of the Bayesian approach. \textcolor{black}{To evaluate the estimator, we consider both MSE
\begin{align} 
\mathcal{E}_{\mathrm{MSE}} \equiv \frac{1}{N}\left\lVert\hat{\bm{w}} - \bm{w}_0\right\rVert_2^2, 
\end{align} 
and GE defined as
\begin{align} 
    \mathcal{E}_{GE}(\bm{x}_{new}) = \frac{1}{4}\mathbb{E}_{y_{new},\bm{x}_{new}, \mathcal{D}^l,\mathcal{D}^u}\Big[(y_{new}-\hat{y}_{new})^2\Big]\Leq{ge},
\end{align}
where $y_{new}$ is the true label of unseen data $\bm{x}_{new}$, $\mathcal{D}^l$ and $\mathcal{D}^u$ are the training datasets, and $\hat{y}_{new}$ is a prediction or estimator for $y_{new}$.}
These two approaches can be treated in a unified manner by the $\beta$-posterior:
\be 
p_{\beta}(\bm{w}\mid \mathcal{D};\sigma^2,\lambda) = \frac{1}{Z_{\beta}}p^{\beta}(\mathcal{D}\mid \bm{w};\sigma^2) p^{\beta}(\bm{w};\lambda).\Leq{eq8}
\ee
The posterior distribution is reproduced at $\beta=1$ and the limit $\beta\to \infty$ yields the point-wise measure at RMLE as $p_{\beta}(\bm{w}\mid \mathcal{D};\sigma^2,\lambda) \to \delta\lb \V{w}-\hat{\V{w}}\lb \mathcal{D};\sigma^2,\lambda \rb\rb$. Hence we may concentrate on computing the expectation over the $\beta$-posterior in general $\beta$ instead of treating those two approaches separately. 

\textcolor{black}{The above estimation problem is analyzed in a high dimensional limit where $N, M_l, M_u \to \infty$ while keeping  $\alpha_l = M_l/N = \mathcal{O}(1)$  and  $\alpha_u = M_u/N = \mathcal{O}(1)$; this limit is recently called proportional limit. In this limit, AMP and SE are expected to become exact, giving us precise asymptotic information. The analytical results are characterized by a small number of quantities known as order parameters, and examining their parameter dependence provides detailed insights into the system. The technical strategy of this paper is to analyze these order parameters in a unified manner using the  $\beta$-posterior distribution, covering both  $\beta = 1$  and  $\beta = \infty$. In the following subsections, we explain how to handle the  $\beta$-posterior distribution in detail.}
\subsection{Factor graph representation}\Lsec{sub_Factor}
To handle the $\beta$-posterior, we employ the framework of graphical modeling and message-passing algorithms. For this, we represent the $\beta$-posterior as the following factorized form:
\be
p_{\beta}(\bm{w}\mid \mathcal{D};\sigma^2,\lambda) = \frac{1}{Z_\beta} \prod_{\mu \in \mathcal{D}^l} \Phi_\mu^l(\bm{w},\bm{x}_\mu,y_\mu)\prod_{\nu\in\mathcal{D}^u}\Phi_\nu^u(\bm{w},\bm{x}_\nu)\prod_{i=1}^N \psi_{i}(w_i),
    \Leq{factorized_form}
\ee
where
\begin{align}
&\Phi_{\mu}^l(\bm{w},\bm{x}_\mu,y_\mu)= \exp\bigg\{{-\frac{\beta}{2\sigma^2}\Big\Vert\bm{x}_\mu -\frac{y_\mu \bm{w}}{\sqrt{N}}\Big\Vert_2^2}\bigg\},\Leq{eq10}\\
   & \Phi_{\nu}^u(\bm{w},\bm{x}_\nu) =\bigg(\rho\exp\bigg\{-\frac{1}{2\sigma^2}\Big\Vert\bm{x}_{\nu} - \frac{\bm{w}}{\sqrt{N}}\Big\Vert_2^2\bigg\}+ (1-\rho)\exp\bigg\{-\frac{1}{2\sigma^2}\Big\Vert\bm{x}_{\nu} + \frac{\bm{w}}{\sqrt{N}}\Big\Vert_2^2 \bigg\}\bigg)^\beta,\Leq{eq11}\\
   & \psi_{i}(w_i) \propto \exp\bigg\{-\frac{\beta}{2}\lambda w_i^2 \bigg\}.\Leq{eq12}
\end{align}
We call these functions $\Phi^l,\Phi^u,\psi$ potential functions according to the standard terminology of graphical models \cite{jordan1999introduction,wainwright2008graphical}. A factor graph representation of \Req{factorized_form} is introduced as the left panel of \Rfig{factor_graph}.
\begin{figure}[htbp]
\centering
\includegraphics[width=0.49\textwidth]{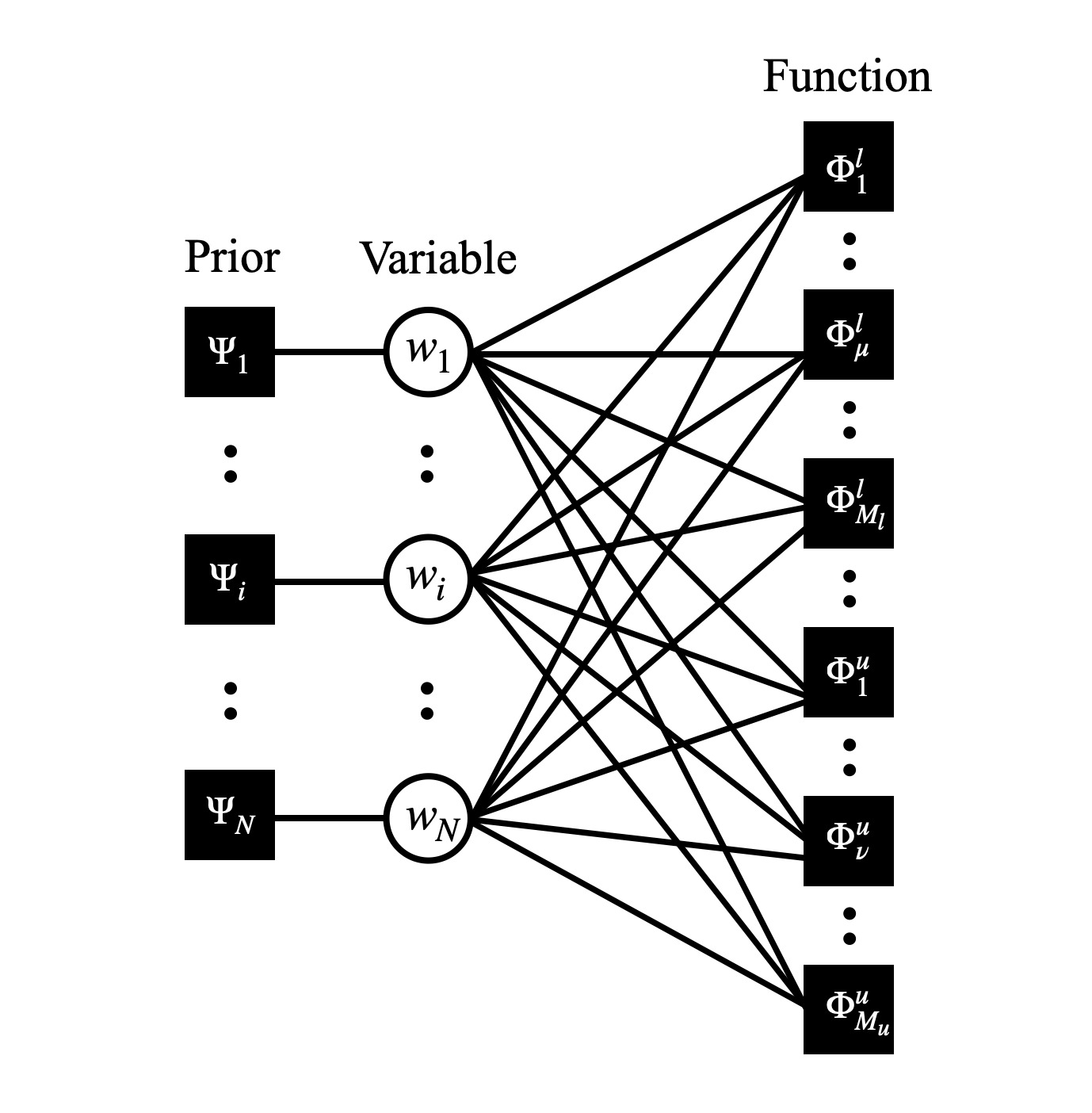}
\includegraphics[width=0.49\textwidth]{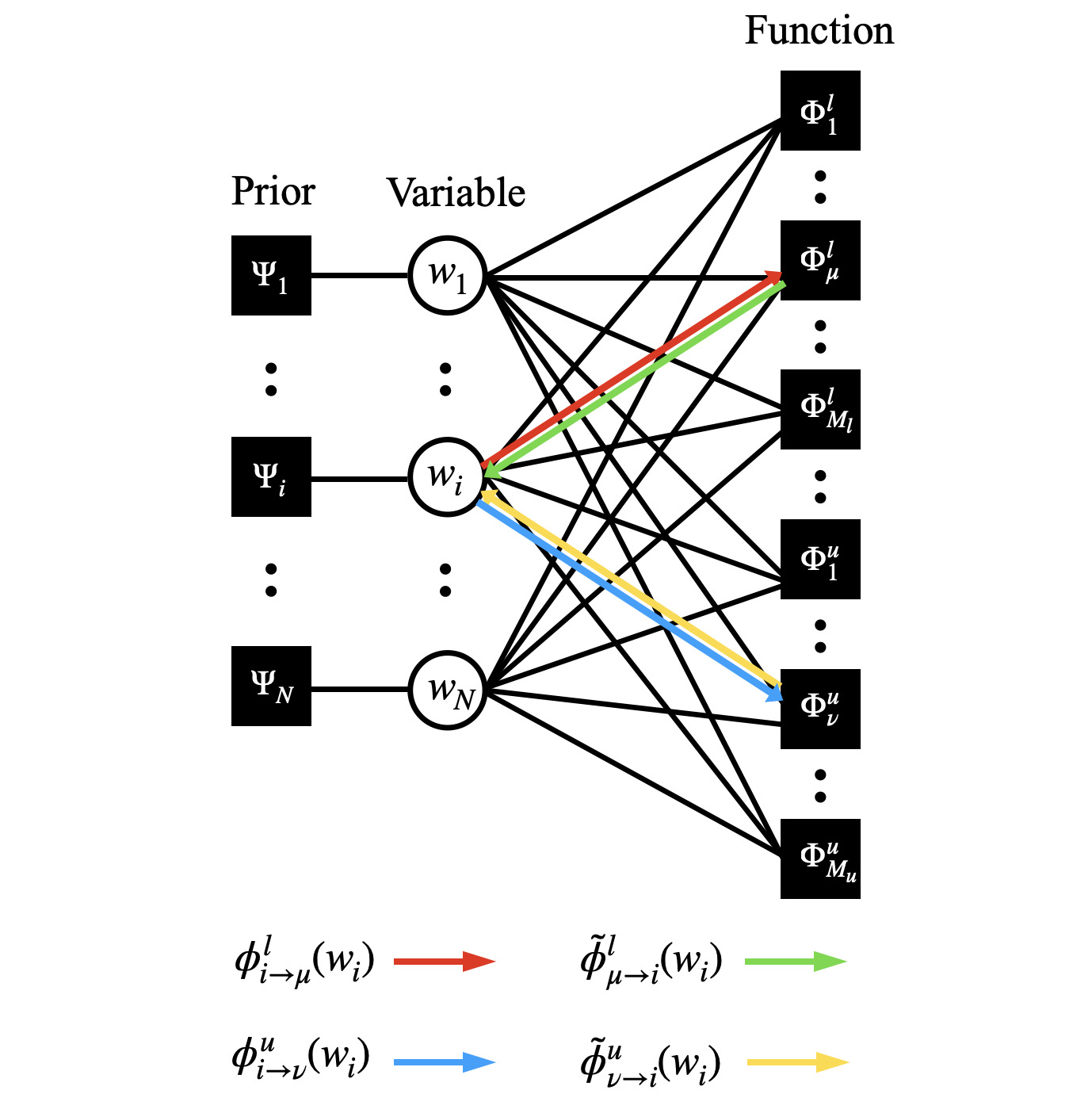}
\caption{(Left) Factor graph corresponding to \Req{factorized_form} (Right) Visualization of messages passing on the graph.}
\Lfig{factor_graph}
\end{figure}
On this representation, we compute the marginal distributions $\{ p_{\beta,i}(w_i)=\int \lb \prod_{j(\neq i)}dw_j\rb p_{\beta}(\V{w})\}_{i=1}^{N}$ using message-passing algorithms. 

For notational simplicity, we denote the average over the $\beta$-posterior as
\be
\Ave{\cdots}=\int d\V{w}~
p_{\beta}(\bm{w}\mid \mathcal{D};\sigma^2,\lambda)\lb \cdots \rb.\Leq{eq13}
\ee
Besides, we consider another distributions $p_{\beta}^{\bs \mu}$ representing the $\beta$-posterior without a specific potential function $\Phi_{\mu}^{u}$ in \Req{factorized_form}; the average over $p_{\beta}^{\bs \mu}$ is denoted by $\Ave{\cdots}^{\bs \mu}$.

\subsection{Belief propagation and approximate belief propagation}
According to the usual prescription of BP \cite{yedidia2003understanding,kschischang2001factor}, we consider the messages passing from function node to variable node $\big\{\tilde{\phi}_{\mu\to i}^l(w_i), \tilde{\phi}_{\nu\to i}^u(w_i)\big\}$, and from variable node to function node $\big\{\phi_{i\to\mu}^l(w_i), \phi_{i\to\nu}^u(w_i)\big\}$. These messages are connected to each other through the following iterative equations:
\subbe
\be
    \tilde{\phi}_{\mu\to i}^l(w_i)&\propto& \int \lb \prod_{j(\neq i)}dw_j\rb \Phi_\mu^l(\bm{w},\boldsymbol{x}_\mu,y_\mu)\prod_{j (\neq i)}^N \phi_{j\to\mu}^l(w_j),\Leq{BP_l_mu2i}\\
    \phi_{i\to\mu}^l(w_i)&\propto& \psi(w_i)\prod_{\omega (\neq \mu)}^{M_l}\tilde{\phi}_{\omega\to i}^l(w_i)\prod_{\nu=1}^{M_u}\tilde{\phi}_{\nu\to i}^u(w_i),\Leq{BP_l_i2mu}\\
    \tilde{\phi}_{\nu\to i}^u(w_i)&\propto& \int \lb \prod_{j(\neq i)}dw_j\rb \Phi_\nu^u(\bm{w},\bm{x}_\nu)\prod_{j (\neq i)}^N \phi_{j\to\nu}^u(w_j),\Leq{BP_u_nu2i}\\
    \phi_{i\to\nu}^u(w_i)&\propto& \psi(w_i)\prod_{\mu =1}^{M_l}\tilde{\phi}_{\mu\to i}^l(w_i)\prod_{m(\neq \nu)}^{M_u}\tilde{\phi}_{m\to i}^u(w_i).\Leq{BP_u_i2nu}
\ee
\Leq{BP}
\subee
A visualization of the messages is given as the right panel of \Rfig{factor_graph}. The solution of \Req{BP} provides an access to the true marginal distribution $p_{\beta,i}(w_i)$ as
\be
p_{\beta,i}(w_i)\propto \psi(w_i)
\prod_{\mu \in \mathcal{D}^{l}}\tilde{\phi}^{l}_{\mu \to i}(w_i)
\prod_{\nu \in \mathcal{D}^{u}}\tilde{\phi}^{u}_{\nu \to i}(w_i).\Leq{eq15}
\ee
Unfortunately, \BReq{BP} involves a computational difficulty coming from the high-dimensional integration in \Reqs{BP_l_mu2i}{BP_u_nu2i}. ABP avoids this difficulty by replacing the high-dimensional integration with a one-dimensional Gaussian integration. Such a replacement is possible because in \Reqs{BP_l_mu2i}{BP_u_nu2i} the potential functions' dependence on $\V{w}$ only appears through the form $\frac{1}{\sigma^2 \sqrt{N}}\sum_j^N x_{\mu j} w_j$ that is the extensive sum of random variables and can be regarded as a Gaussian variable thanks to \textcolor{black}{CLT}. 

Let us focus on how to compute \Req{BP_l_mu2i} according to the above idea. Denoting $d_{\mu}^{l}=\frac{1}{\sigma^2 \sqrt{N}}\sum_j^N x_{\mu j} w_j$ and isolating the $i$th term, we can write  
\be
d_\mu^l = \frac{1}{\sigma^2\sqrt{N}}x_{\mu i}w_i + \frac{1}{\sigma^2\sqrt{N}}\sum_{j(\neq i)}^N x_{\mu j} w_j  = \frac{1}{\sigma^2\sqrt{N}}x_{\mu i}w_i + \tilde{p}_{\mu\to i}^{l} + \sqrt{\frac{S_\mu^{l}}{\beta}}z,
\Leq{d_CLT}
\ee
where $\tilde{p}_{\mu\to i}^{l}$ and $\frac{S_\mu^{l}}{\beta}$ are the mean and variance of $\frac{1}{\sigma^2\sqrt{N}}\sum_{j(\neq i)}^N x_{\mu j} w_j$ over the distribution $\prod_{j (\neq i)}^N \phi_{j\to\mu}^l(w_j)$, and $z$ is the zero-mean unit-variance  Gaussian random variable representing the fluctuation around the mean. The explicit forms of the mean and variance are
\be
\tilde{p}_{\mu\to i}^{l} &=& \frac{1}{\sigma^2\sqrt{N}}\sum_{j(\neq i)}^N x_{\mu j}\hat{w}_{j\to\mu}^{l},\Leq{eq17}\\
S_\mu^{l} &=& \frac{1}{\sigma^4 N}\sum_j^N x_{\mu j}^2\chi_{j\to \mu}^{l},\Leq{eq18}
\ee
where 
\be 
 \hat{w}_{j\to \mu}^{l} &\equiv& \int dw_{j} \ w_j\prod_{j(\neq i)}^N \phi_{j \to \mu}^{l}(w_j) \equiv \langle w_j \rangle^{\bs \mu},\Leq{eq19}\\
\chi_{i\to\mu}^{l} &\equiv& 
\beta \bigg\{ \big\langle w_i^2\big\rangle^{\bs\mu} - \Big( \big\langle w_i \rangle^{\bs\mu}\Big)^2 \bigg\}
\Leq{eq20}.
\ee
Note that the average over $\prod_{j (\neq i)}^N \phi_{j\to\mu}^l(w_j)$ is here identified with the average over $p_{\beta}^{\bs \mu}$ and is denoted as $\Ave{\cdot }^{\bs \mu}$. This is the direct consequence of the BP's assumption which can be justified in the large $N$ limit with the current generative model of $\V{x}_{\mu}$ yielding no correlation between different components. Another noteworthy remark is that in the variance formula \NReq{eq20} the $i$th contribution $\frac{1}{\sigma^4 N}x_{\mu i}^2\chi_{i\to \mu}^{l}$ is included though it should not be there when treating $\frac{1}{\sigma^2\sqrt{N}}\sum_{j(\neq i)}^N x_{\mu j} w_j$ literally. This correction is added to make the following calculations simpler by lightening the notation and does not affect the result because the correction is sufficiently small ($\mathcal{O}(1/N)$) and negligible in the large $N$ limit.

The first term in the \textcolor{black}{right-hand} side of \Req{d_CLT} is small thanks to the factor $1/\sqrt{N}$, justifying an expansion of the potential function with respect to (w.r.t.) the term up to the second order. This implies that the message $\tilde{\phi}^{l}_{\mu \to i}(w_i)$ takes a Gaussian form. The same holds for $\tilde{\phi}^{u}_{\mu \to i}(w_i)$, and resultantly all the messages $\{\tilde{\phi}^{l}_{\mu \to i},\phi^{l}_{i \to \mu},\tilde{\phi}^{u}_{\mu \to i},\phi^{u}_{i \to \mu} \}$ become Gaussian. Hence, the functional update \NReq{BP} can be reduced to iterative equations for the parameters characterizing those Gaussians. In this way, the computationally intractable formulas \NReq{BP} are converted to a feasible problem.

The iterative equations of the Gaussian parameters of the messages depend on the value of $\beta$. Here, we only quote the resultant formulas at \textcolor{black}{$\beta=\infty$ and $\beta=1$}. For readers interested in the derivation details, we state them in \Rapp{subsub_RMLE} and \Rapp{subsub_BO}. 

The ABP formulas for RMLE are 
\subbe
\Leq{ABP_RMLE}
\be
    \tilde{p}_{\nu\to i}^{(t)} &=& \frac{1}{\sigma^2\sqrt{N}}\sum_{j(\neq i)}^N x_{\nu j}\hat{w}_{j\to\nu}^{(t)}, \Leq{bp1r}\\
    S_\nu^{(t)} &=& \frac{1}{\sigma^4 N}\sum_{i}^N x_{\nu i}^2 \chi_{i\to \nu}^{(t)},\Leq{bp2r}\\
    \chi_{i\to \nu}^{(t+1)} &=& \Bigg(\lambda+\frac{\alpha}{\sigma^2} - \frac{1}{\sigma^4 N}\sum_{m(\neq\nu)}^{M_u} x_{m i}^2  T\Big(\tilde{p}_{m\to i}^{(t)},S_m^{(t)}\Big)\Bigg)^{-1},\Leq{bp3r}\\
    \hat{w}_{i\to\nu}^{(t+1)} &=& \frac{\chi_{i\to \nu}^{(t+1)}}{\sigma^2 \sqrt{N}}\Bigg(\sum_{\mu=1}^{M_l}y_\mu x_{\mu i} + \sum_{m(\neq\nu)}^{M_u}x_{m i}F\Big(\tilde{p}_{m\to i}^{(t)},S_m^{(t)}\Big)\Bigg).\Leq{bp4r}
\ee
\subee
The functions $F(p,t)$ and $T(p,t)$ are introduced as follows:
\be
g(y|p,t) &=& \ln\Big(\rho\exp\big\{p+\sqrt{t}y\big\}+(1-\rho)\exp\big\{-\big(p+\sqrt{t}y\big)\big\}\Big),\Leq{eq22}
\\
G(y|p,t) &=& -\frac{y^2}{2} + g(y|p,t),\Leq{eq23}
\\
y^*(p,t)&\equiv& \operatorname*{arg\,max}_{y}G(y|p,t),\Leq{eq24}
\\
F(p,t) &\equiv& \frac{\partial}{\partial p}G(y^*(p,t)|p,t),\Leq{eq25}
\\
T(p,t) &\equiv& \frac{\partial^2}{\partial p^2}G(y^*(p,t)|p,t).\Leq{eq26}
\ee
The function $g(y|p,t)$ corresponds to the logarithm of the potential function $\Phi_{\nu}^u(\bm{w},\bm{x}_\nu)$, and the maximization w.r.t. $y$ comes from the saddle-point condition when taking the limit $\beta \to \infty$. On the other hand, the Bayesian approach provides the following ABP formulas:
\subbe
\Leq{ABP_Bayes}
\be
\tilde{p}_{\nu\to i,B}^{(t)} &=& \frac{1}{\sigma^2\sqrt{N}}\sum_{j(\neq i)}^N x_{\nu j}\hat{w}_{j\to\nu,B}^{(t)},\Leq{bp1b}\\
\chi^{(t+1)}_{i\to\nu,B} &=& \Bigg(\lambda+\frac{\alpha}{\sigma^2}-\frac{1}{\sigma^4 N}\sum_{m\neq\nu}^{M_\mu}x_{m i}^2\tilde{T}\Big(\tilde{p}_{m\to i,B}^{(t)}\Big)\Bigg)^{-1},\Leq{bp2b}\\
\hat{w}_{i\to\nu,B}^{(t+1)} &=& \frac{\chi^{(t+1)}_{i\to\nu,B}}{\sigma^2 \sqrt{N}}\Bigg(\sum_{\mu=1}^{M_l}y_\mu x_{\mu i} + \sum_{m(\neq\nu)}^{M_u}x_{m i}\tilde{F}\Big(\tilde{p}_{m\to i,B}^{(t)}\Big)\Bigg),\Leq{bp3b}
\ee
\subee
where 
\be 
&&
\tilde{F}(p) \equiv \frac{\rho \exp\{p\}-(1-\rho)\exp\{-p\}}{\rho\exp\{p\}+(1-\rho)\exp\{-p\}},\Leq{eq28}
\\
&& 
\tilde{T}(p) \equiv \frac{\partial}{\partial p} \tilde{F}(p).\Leq{eq29}
\ee
Here, to make the algorithmic aspect explicit, we put the superscript $(t)$ to denote the time index within the iteration. Bayesian results are subscripted with ``$B$" to make them easier to distinguish from the results in RMLE. For the derivation, please refer to \Rapp{subsub_BO}.

In \Req{ABP_RMLE}, the dependence on the superscripts $l$ and $u$ of the quantities $\big\{\tilde{p}_{\nu\to i}, S_\nu, \chi_{i\to \nu}, \hat{w}_{i\to\nu}\big\}$ is neglected. To understand the reason why this is justified, we compare the following two quantities:
\be
\hat{w}_{i\to\mu}^{l(t+1)} &=&\frac{\chi_{i\to \mu}^{l(t+1)}}{\sigma^2 \sqrt{N}}\Bigg(\sum_{\omega(\neq \mu)}^{M_l}y_\omega x_{\omega i} + \sum_{\nu = 1}^{M_u}x_{\nu i}F\Big(\widetilde{p}_{\nu\to i}^{u(t)},S_\nu^{u(t)}\Big)\Bigg),\Leq{eq30}\\
\hat{w}_{i\to\nu}^{u(t+1)} &=&\frac{\chi_{i\to \nu}^{u(t+1)}}{\sigma^2 \sqrt{N}}\Bigg(\sum_{\mu=1}^{M_l}y_\mu x_{\mu i} + \sum_{m(\neq\nu)}^{M_u}x_{m i}F\Big(\widetilde{p}_{m\to i}^{u(t)},S_m^{u(t)}\Big)\Bigg).\Leq{eq31}
\ee
The former is the posterior mean of $w_i$ in the system without the labeled potential function $\Phi_{\mu}^{l}$ while the latter one is the counterpart when the unlabeled potential function $\Phi_{\nu}^{u}$ is absent. The first important observation is that the difference between these two quantities is small and can be neglected. The second important one is the nontrivial parameters $\{ \widetilde{p}_{m\to i}^{u(t)}\}_{\mu,i}$ requires only the unlabeled mean $\{ \hat{w}_{i\to\nu}^{u(t)}\}_{i,\nu}$ for their computations. Hence, it is sufficient for closing the equations to only compute $\{ \hat{w}_{i\to\nu}^{u(t)}\}_{i,\nu}$, justifying to treat $\{ \hat{w}_{i\to\nu}^{u(t)}\}_{i,\nu}$ only and to remove the superscript $u$ from the respective quantities.

\subsection{\textcolor{black}{Approximate message-passing}} 
The ABP equations can be further approximated because the posterior mean under the absence of the potential function $\Phi_{\nu}^{u}$, $\hat{w}_{i\to \nu}$, is well approximated by the full posterior mean $\hat{w}_i=\Ave{w_i}$, since the difference is due to the only one potential function and is small. Rewriting all $\hat{w}_{i\to \nu}$ by $\hat{w}_i$ in the ABP equations yields another set of iterative equations in terms of $\{\hat{w}_i\}_i$. This is simpler because the number of variables tracked in the equations is reduced from $NM$ to $N$, and resultantly the computational cost is reduced from $\mathcal{O}(NM^2)$ to $\mathcal{O}(NM)$. The algorithm derived according to this idea is called AMP. Here we summarize the AMP algorithms for the RMLE and Bayesian cases.

AMP for the RMLE is given by
\subbe
\Leq{AMP_RMLE}
\begin{align}
    \tilde{p}_\nu^{(t)} &= \frac{1}{\sigma^2\sqrt{N}}\sum_j^N x_{\nu j}\hat{w}_j^{(t)} - \frac{\chi^{(t)}}{\sigma^4 N}\sum_{j}^N x_{\nu j}^2 F\Big(\tilde{p}_\nu^{(t-1)},\frac{\chi^{(t-1)}}{\sigma^2}\Big),\Leq{amp1r}\\
    \chi^{(t+1)} &= \Bigg(\lambda+\frac{\alpha}{\sigma^2} - \frac{1}{\sigma^4 N}\sum_{\nu=1}^{M_u}  x_{\nu i}^2 T\Big(\tilde{p}_{\nu}^{(t)},\frac{\chi^{(t)}}{\sigma^2}\Big)\Bigg)^{-1},\Leq{amp2r}\\
    \hat{w}_i^{(t+1)} &= \frac{\chi^{(t+1)}}{\sigma^2 \sqrt{N}}\Bigg(\sum_{\mu = 1}^{M_l}y_\mu x_{\mu i} + \sum_{m = 1}^{M_u}x_{m i} F\Big(\tilde{p}_{m}^{(t)},\frac{\chi^{(t)}}{\sigma^2}\Big) - \frac{\hat{w}_{i}^{(t)}}{\sigma^2\sqrt{N}}\sum_{m = 1}^{M_u} x_{m i}^2 T\Big(\tilde{p}_m^{(t)},\frac{\chi^{(t)}}{\sigma^2}\Big)\Bigg).\Leq{amp3r}
\end{align}
\subee
While AMP for the Bayesian approach is
\subbe
\Leq{AMP_BA}
\begin{align}
    \tilde{p}_{\nu,B}^{(t)} &= \frac{1}{\sigma^2\sqrt{N}}\sum_j^N x_{\nu j}\hat{w}_{j,B}^{(t)} - \frac{\chi_{B}^{(t)}}{\sigma^4 N}\sum_j^N x_{\nu j}^2 \tilde{F}{\Big(\tilde{p}_{\nu,B}^{(t-1)}\Big)},\Leq{amp1b}\\
    \chi^{(t+1)}_{B} &= \Bigg(\lambda + \frac{\alpha}{\sigma^2} - \frac{1}{\sigma^4 N}\sum_{\nu=1}^{M_u}x_{\nu i}^2\tilde{T}\Big(\tilde{p}_{\nu,B}^{(t)}\Big)\Bigg)^{-1},\Leq{amp2b}\\
    \hat{w}_{i,B}^{(t+1)} &= \frac{\chi^{(t+1)}_{B}}{\sigma^2\sqrt{N}}\Bigg(\sum_{\mu=1}^{M_l}y_{\mu}x_{\mu i}+\sum_{m=1}^{M_u} x_{m i}\tilde{F}{\Big(\tilde{p}_{m,B}^{(t)}\Big)} - \frac{\hat{w}_i^{(t)}}{\sigma^2\sqrt{N}}\sum_{m=1}^{M_u}x_{m i}^2\tilde{T}\Big(\tilde{p}_{m,B}^{(t)}\Big)\Bigg).\Leq{amp3b}
\end{align}
\subee
The derivations of these AMP algorithms are deferred to \Rapp{AMP}.

\subsection{\textcolor{black}{State evolution}}\Lsec{state_evolution}
The macroscopic behavior of BP can be analyzed by a theoretical framework called SE \cite{donoho2010message,bayati2011dynamics}. In the large $N$ limit, we may assume that the estimator $\hat{w}_{i\to\nu}$ follows a Gaussian distribution $\mathcal{N}\left(k_t w_{0i},v_t\right)$ where the stochasticity comes from the data generation process given $\V{w}_0$, \Req{eq3}. The two macroscopic parameters $k_t$ and $v_t$ describe the overlap with the true parameter $\boldsymbol{w}_0$ and the variance, respectively. The iterative equations that these two parameters obey can be derived from the BP iteration, and those equations are called SE equations. Here we summarize the SE equations of the RMLE and Bayesian cases.

The SE equations of RMLE are given by 
\subbe
\Leq{SE_RMLE}
\be
    \chi_{t+1} &=& \Bigg(\lambda+\frac{\alpha}{\sigma^2} - \frac{\alpha_u}{\sigma^2} \int Dz\ \bigg(\rho T\Big(\mathcal{P},\frac{\chi_t}{\sigma^2}\Big)+(1-\rho)T\Big(\mathcal{Q} ,\frac{\chi_t}{\sigma^2}\Big) \bigg)\Bigg)^{-1},\Leq{se1r}\\
    k_{t+1} &=& \chi_{t+1}\Bigg(\frac{\alpha_l}{\sigma^2}+\frac{\alpha_u}{\sigma^2}\int Dz\ \bigg(\rho F\Big(\mathcal{P} ,\frac{\chi_t}{\sigma^2}\Big)-(1-\rho) F\Big(\mathcal{Q} ,\frac{\chi_t}{\sigma^2}\Big)\bigg)\Bigg),\Leq{se2r}\\
    v_{t+1} &=& \chi_{t+1}^2\Bigg(\frac{\alpha_l}{\sigma^2} + \frac{\alpha_u}{\sigma^2} \int Dz\ \bigg(\rho F^2\Big(\mathcal{P}, \frac{\chi_t}{\sigma^2}\big) +(1-\rho) F^2\Big(\mathcal{Q}, \frac{\chi_t}{\sigma^2}\Big)\bigg)\Bigg),\Leq{se3r}
\ee
\subee
where $Dz = e^{-z^2/2}/\sqrt{2\pi}\ dz$, $\alpha_l = M_l/N,\ \alpha_u = M_u/N$, and
\be
&&
v_s = \frac{1}{N}\sum_{j(\neq i)}^N w_{0 j}^2,\Leq{eq33} 
\\ &&
\tilde{v}_t = k_t^2 v_s + v_t,\Leq{eq34}
\\ &&
\mathcal{P} = \frac{k_t v_s}{\sigma^2} + \sqrt{\frac{\tilde{v}_t}{\sigma^2}} z,\Leq{eq35}
\\ &&
\mathcal{Q} = -\frac{k_t v_s}{\sigma^2} + \sqrt{\frac{\tilde{v}_t}{\sigma^2}} z.
\Leq{eq36}
\ee
In the limit $N\to \infty$, $v_s$ becomes equal to the true signal variance $1/\lambda_0$ in \Req{eq33}. Henceforth, we assume $v_s=1/\lambda_0$ hereafter. Those for the Bayesian approach are 
\subbe
\Leq{SE_BA}
\be
    \chi_{t+1,B} &=& \Bigg(\lambda+\frac{\alpha}{\sigma^2} - \frac{\alpha_u}{\sigma^2} \int Dz\ \bigg(\rho \tilde{T}\big(\mathcal{P}_{B}\big)+(1-\rho)\tilde{T}\big(\mathcal{Q}_{B}\big) \bigg)\Bigg)^{-1},\Leq{se1b}\\
    k_{t+1,B} &=& \chi_{t+1,B}\Bigg(\frac{\alpha_l}{\sigma^2}+\frac{\alpha_u}{\sigma^2}\int Dz\ \bigg(\rho \tilde{F}\big(\mathcal{P}_{B}\big)-(1-\rho) \tilde{F}\big(\mathcal{Q}_{B}\big)\bigg)\Bigg),\Leq{se2b}\\
    v_{t+1,B} &=& \chi_{t+1,B}^2\Bigg(\frac{\alpha_l}{\sigma^2} + \frac{\alpha_u}{\sigma^2} \int Dz\ \bigg(\rho \tilde{F}^2\big(\mathcal{P}_{B}\big)+(1-\rho) \tilde{F}^2\big(\mathcal{Q}_{B}\big)\bigg)\Bigg).\Leq{se3b}
\ee
\subee
The three parameters $\chi_t,k_t,v_t$ ($\chi_{t, B},k_{t, B},v_{t, B}$) characterize the macroscopic behavior of BP or ABP iteration. We call these parameters $\textit{order parameters}$ according to physics terminology.  The derivations of \Reqs{SE_RMLE}{SE_BA} are given in \Rapp{state_evolution}.

These order parameters enable us to compute many quantities of interest. For example, the MSE between the true parameter $\V{w}_0$ and the estimator $\hat{\V{w}}$ can be expressed by using the above overlap $k$ and the variance $v$ as
\begin{align}
    \mathcal{E}_{MSE} &\equiv\frac{1}{N}\big\lVert\hat{\bm{w}} - \boldsymbol{w_0}\big\rVert_2^2
    = (k-1)^2/\lambda_0+v\Leq{mse1}.
\end{align}
It is also possible to track how MSE evolves along the algorithm steps from the SE-evaluated $k_t$ and $v_t$, which precisely corresponds to the MSE evolution computed from the BP or ABP algorithm. \textcolor{black}{The AMP algorithm is essentially the same as the ABP one except for the computational cost, and hence SE again can fully characterize the AMP behavior in the large $N$ limit.} In this way, SE provides a simple way to analyze the macroscopic dynamical behavior of the algorithms.


\subsection{Convergence control and the $\lambda$-$\chi$ correspondence}\Lsec{sub_Conv}
As far as we numerically examined, the AMP iteration \NReq{AMP_RMLE} tends to be badly converging or even diverging. We also observed that this convergence issue can be partly attributed to the non-negativity and the oscillating tendency of $\chi^{(t)}$. To control the convergence, we thus avoid updating $\chi^{(t)}$ and instead fix it to a certain appropriate value during the iteration. Our numerical experiments show that this strategy works very well and the convergence is greatly improved. Hence hereafter we adopt this strategy both in running AMP and SE iterations.

Fortunately, this strategy does not prevent us from our primary objective of analyzing RMLE given $\lambda$, $\hat{\V{w}}(\lambda)$. This is thanks to the fact that the regularization coefficient $\lambda$ only appears in the $\chi$-related equations in the AMP/SE iterations \NReqs{AMP_RMLE}{SE_RMLE}. This implies that there is a certain simple correspondence between $\chi$ and $\lambda$, and we can find an appropriate value of $\lambda$ on which the estimator $\hat{\V{w}}(\lambda)$ is (asymptotically) equal to the convergent solution of AMP given $\chi$, $\V{w}^*(\chi)=\lim_{t\to \infty}\V{w}^{(t)}(\chi)$. 

The above equality $\hat{\V{w}}(\lambda)=\V{w}^*(\chi)$ defines the correspondence between $\lambda$ and $\chi$, and the correspondence can be computed by using SE. The procedure is simple: fix $\chi$ to a certain appropriate value; conduct the SE iteration \NReq{SE_RMLE} until the convergence on the fixed $\chi$; solve \Req{se1r} w.r.t. $\lambda$ on the convergent solution with setting $\chi_{t+1}$ and $\chi_{t}$ to the fixed value of $\chi$. In this way, we can obtain a relation between $\lambda$ and $\chi$, and the example plots for several $\alpha_u$ are shown as the left panel of \Rfig{lambda and chi}. The correspondence between $\lambda$ and $\chi$ is expected to be one-to-one, and our SE-based result confirms this reasonable expectation. Another interesting finding from this figure is that there exists a cusp at a certain value of $\chi$ against a fixed $\alpha_u(>0)$. This signals the emergence of a phase transition which is extensively discussed in \Rsec{theoretical_results}. 

For a given $\lambda$, the above relation enables us to choose the appropriate $\chi$ value, thus allowing us to handle AMP and SE with good convergence given $\chi$.  All the results by AMP and SE in the following sections are obtained in this way. The same is true for the Bayesian case \NReqs{SE_BA}{AMP_BA}, and the example plots of the $\lambda$-$\chi$ relation are given as the right panel of \Rfig{lambda and chi}.
\begin{figure}[htbp]
\centering
\includegraphics[width=0.49\textwidth]{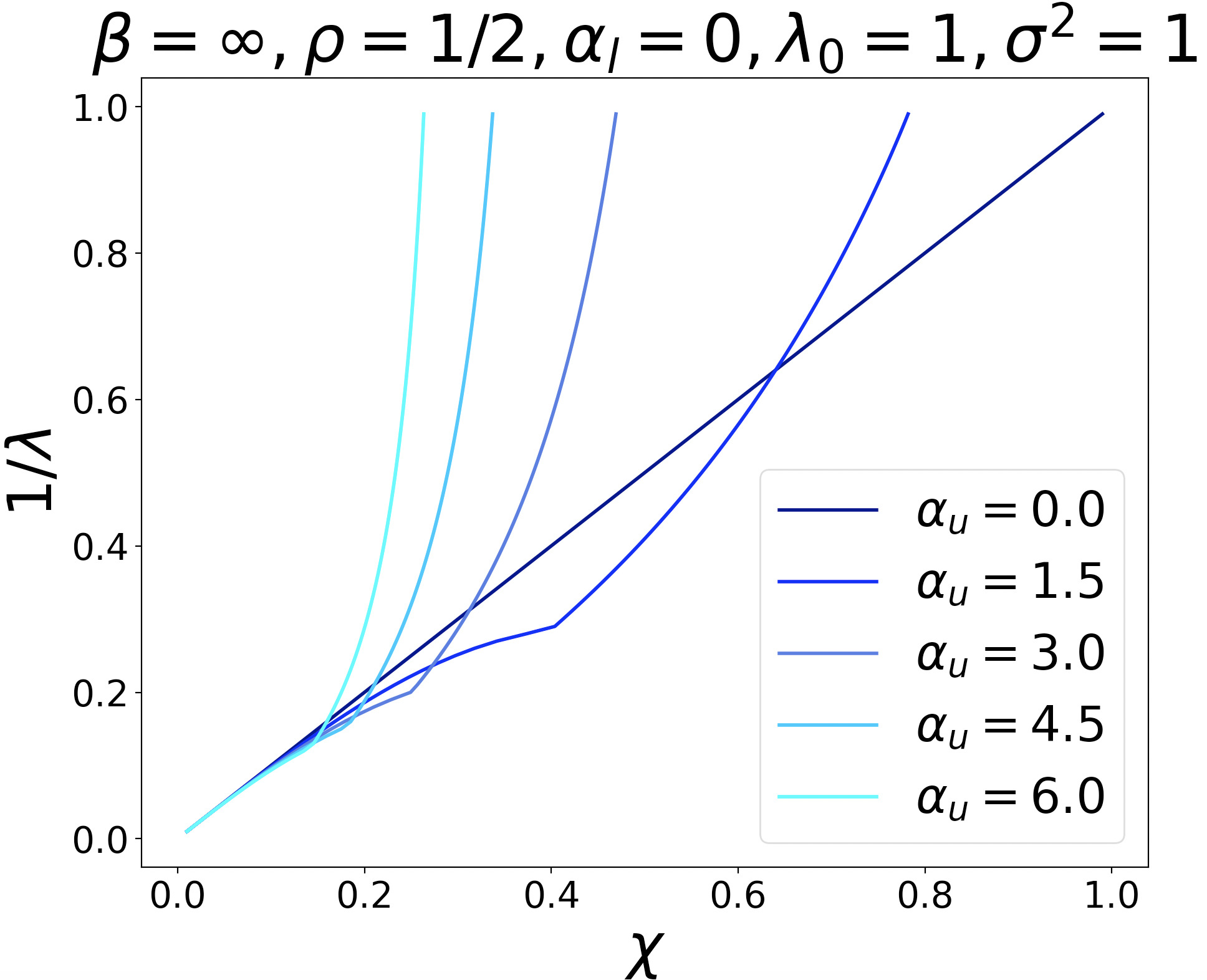}
\includegraphics[width=0.49\textwidth]{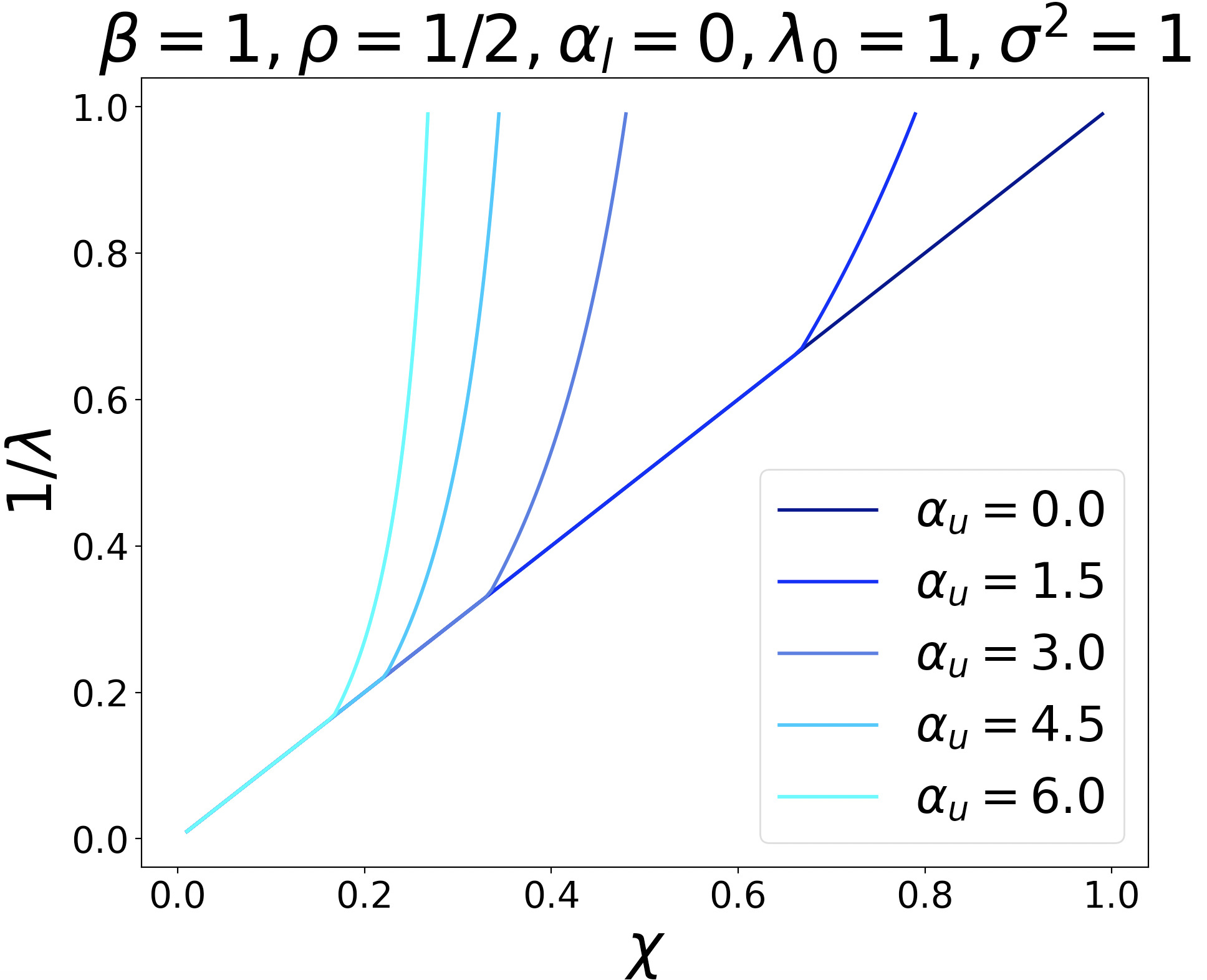}
\caption{\label{fig:fg} The relation between $\chi$ and $1/\lambda$ for RMLE (left) and the Bayesian approach (right) at $\rho=1/2,\alpha_l=0,\lambda_0=1,\sigma^2=1$. The straight black lines depict the relation $1/\lambda=\chi$, which is the correct trivial relation at $\alpha_u=\alpha_l=0$.}
\Lfig{lambda and chi}
\end{figure}

\textcolor{black}{We also compare the consistency between SE and AMP, as well as between AMP and the gradient-based algorithm. The results are presented in \Rapp{numerical_validation}.}
\section{Theoretical results}\Lsec{theoretical_results} 
In this section, we summarize theoretical findings derivable through SE. The stationary solution of SE, $k_*=\lim_{t\to \infty}k_t,v_*=\lim_{t\to \infty}v_t$, is focused and relevant quantities are computed from it. This stationary solution can exhibit some drastic change at certain regions when the parameters $\sigma^2(=\sigma_0^2), \lambda_0, \lambda, \alpha_l, \alpha_u$ are varied. This change is connected to phase transitions in physics, and we start by deriving the associated phase diagrams for RMLE in \Rsec{sub_RMLE} and for the Bayesian approach in \Rsec{sub_BO}. Then, the performances of those two approaches are quantitatively compared in \Rsec{sub_CoRB}.

\subsection{Phase diagram of RMLE}\Lsec{sub_RMLE}  
Here we summarize the phase diagrams for RMLE. The study proceeds in the following manner. We first study the case $\rho = 1/2$ and $\alpha_l=0$. This case has been rather intensively investigated in many previous studies~\cite{barkai1994statistical,rose1990statistical,biehl1993statistical, barkai1993scaling,  watkin1994optimal}, and contains all phases appearing in the other parameter values; thus, we regard this case as the baseline. Then, we change the value of $\alpha_l$ and $\rho$ to see how the labeled data and the imbalance in the label distribution influence the estimation. 

\subsubsection{The baseline case}\Lsec{sub_UEC}
This case has a symmetry between the two label values $y=\pm 1$, and resultantly phase transitions connected to the breaking of this symmetry emerge. Let us examine this point. 

Inserting $\rho=1/2,~\alpha_l=0$ into \Req{SE_RMLE}, we have
\subbe
\Leq{SE_RMLE_spe}
\be
    \chi_{t+1} &=& \Bigg(\lambda+\frac{\alpha_u}{\sigma^2} - \frac{\alpha_u}{\sigma^2}\int Dz\ T\bigg(\frac{k_t}{\lambda_0\sigma^2} + \sqrt{\frac{k_t^2/\lambda_0 + v_t}{\sigma^2}} z,\frac{\chi_t}{\sigma^2}\bigg)\Bigg)^{-1},\Leq{se_spe1}\\
    k_{t+1} &=& \chi_{t+1} \frac{\alpha_u}{\sigma^2} \int Dz\ F\bigg(\frac{k_t}{\lambda_0\sigma^2} + \sqrt{\frac{k_t^2/\lambda_0 + v_t}{\sigma^2}} z, \frac{\chi_t}{\sigma^2}\bigg),\Leq{se_spe2}\\
    v_{t+1} &=& \chi_{t+1}^2 \frac{\alpha_u}{\sigma^2} \int Dz\ F^2\bigg(\frac{k_t}{\lambda_0\sigma^2} + \sqrt{\frac{k_t^2/\lambda_0 + v_t}{\sigma^2}} z,\frac{\chi_t}{\sigma^2}\bigg).\Leq{se_spe3}
\ee
\subee
As explained in \Rsec{sub_Conv}, we use $\chi$ as a control parameter, and it is reasonable to focus on the fixed point $(k_*,v_*)$ given $\chi$ of \Req{SE_RMLE_spe}. A number of solutions emerge as the fixed point depending on the parameter values. We first enumerate those solutions. 

A trivial solution $k_* = v_* = 0$ can be easily found since $F(0,t)=0$ holds by definition. The presence of this solution is the direct consequence of the above symmetry. This condition $k_* = v_* = 0$ means that $\hat{\boldsymbol{w}}$ becomes the zero vector and hence no meaningful estimate of $\boldsymbol{w}_0$ is available. This solution is nothing but the paramagnetic solution in physics, but we here call it \textit{undetected solution} or \textit{undetected phase} from our problem context. The range of $\chi$ in which this undetected solution exists should be clarified. Inserting this solution into \Req{se_spe1} and solving it w.r.t. $\chi$, we have
\be
\chi=\frac{1}{2}\frac{1+\lambda\sigma^2 \pm \sqrt{(1+\lambda\sigma^2)^2 - 4(\alpha_u/\sigma^2+\lambda)\sigma^2}}{\alpha_u/\sigma^2+\lambda}.
\Leq{chi_undetected}
\ee
The positive branch is reasonable since it is connected to the trivial solution $\chi=1/\lambda$ at $\alpha_u=0$. This defines the existence region of the undetected solution as well as the correspondence between $\chi$ and $\lambda$. 

\BReq{SE_RMLE_spe} has two other solutions: the one with $k_*=0,~v_*\neq 0$ and the other one with $k_*\neq 0,~v_*\neq 0$. The former implies that our estimator $\hat{\bm{w}}$ becomes a non-zero vector but is orthogonal to the true parameter $\bm{w}_0$, meaning that our estimate of $\bm{w}_0$ is not different from random guess; this motivates us to call this $\textit{random phase}$. The latter one has a non-zero overlap enabling us to have a nontrivial estimate, and thus we call this $\textit{detected phase}$. In the detected phase, the overlap $k_*$ can be both positive or negative. When $\alpha_l$ is small and $\alpha_u$ is large, a metastable state with $k_* < 0$ has been reported for $\beta = 1$ \cite{tanaka2013statistical}, and our analysis based on SE confirms a similar solution in RMLE. However, since the $k_* < 0$ solution has the higher free energy and AMP can converge to $k_* > 0$ with appropriate initial conditions obtainable with $\alpha_l>0$, the $k_* < 0$ solution is irrelevant. Thus, we focus solely on the solution with $k > 0$ in the remainder of this paper.

Assuming the transitions between the phases are continuous, we can derive compact formulas of critical conditions determining the phase boundary.  The undetected--random critical condition can be obtained as follows. Expanding \Req{se_spe3} w.r.t. $v_t$ from the undetected phase, we get
\be
    v_{t+1} = \frac{\alpha_u \chi^2}{\sigma^2}\int Dz\ \bigg( \frac{\partial F(p,\chi/\sigma^2)}{\partial p}\bigg|_{p=0}\bigg)^2\frac{v_t}{\sigma^2} z^2 = \frac{\alpha_u \chi^2}{\sigma^4} T^2\Big(0,\frac{\chi}{\sigma^2}\Big) v_t.\Leq{eq49}
\ee
The condition where the solution $v_{t+1}=v_{t}=0$ becomes unstable is
\be
    1 = \frac{\alpha_u \chi^2}{\sigma^4} T^2\Big(0,\frac{\chi}{\sigma^2}\Big)\Leftrightarrow \chi = \sigma^2\big(1+\sqrt{\alpha_u}\big)^{-1}.
\Leq{CC_U-R}
\ee
This gives the critical condition between the undetected and random phases. \textcolor{black}{Note that the perturbative discussion here is solely for determining the phase boundary and is not related to {\it replica symmetry breaking} (RSB). However, from the linear stability analysis of AMP discussed in \Rapp{sub_Micr}, it becomes clear that the random phase is accompanied by RSB, and the critical condition from the undetected to random phases accidentally coincides with the instability condition leading to RSB, as seen in \Req{at-line}}. In a similar way, the undetected--detected critical condition is computed by an expansion of \Req{se_spe2} w.r.t. $k_t$ from the undetected phase, yielding the critical condition
\be
    1 = \frac{\alpha_u \chi}{\lambda_0\sigma^4}T\Big(0,\frac{\chi}{\sigma^2}\Big) \Leftrightarrow \chi = \sigma^2\Big(1+\frac{\alpha_u}{\lambda_0\sigma^2}\Big)^{-1}.\Leq{eq52}
\Leq{CC_U-D}
\ee
The critical condition between the detected and random phases can also be derived from the expansion of \Req{se_spe2} w.r.t. $k_t$ from the random solution. The result is
\be
    1 =\frac{\alpha_u \chi}{\lambda_0}  \int Dz \ T\bigg(\sqrt{\frac{v_*}{\sigma^2}} z, \frac{\chi}{\sigma^2}\bigg).\Leq{eq54}
\Leq{CC_D-R}
\ee
The order parameter value $v_*$ comes from the random solution obtained by solving \Req{se_spe3} w.r.t. $v$ under the condition $k=0$. 

Due to the presence of the nontrivial order parameter $v_*(\neq 0)$, the critical condition \NReq{CC_D-R} is less trivial compared to \Reqs{CC_U-R}{CC_U-D} both of which are directly determined from the parameters. Besides, this condition becomes even imprecise due to the phenomenon called RSB. Once RSB occurs, the SE prediction is known to be inaccurate. The critical condition of the RSB region is detected as the dynamical instability of BP (thus of AMP)~\cite{kabashima2003cdma}, and it takes the following form in the present problem:
\be
    1=\frac{\alpha_u \chi^2}{\sigma^4} \int Dz\ T^2\bigg(\frac{k_*}{\lambda_0\sigma^2} + \sqrt{\frac{k_*^2/\lambda_0+v_*}{\sigma^2}} z,\frac{\chi}{\sigma^2}\bigg).
 \Leq{at-line}
\ee
The detailed derivation is deferred to \Rapp{sub_Micr}. This condition is reduced to \Req{CC_U-R} if we assume the undetected solution $k_*=v_*=0$, implying that the random phase involves RSB or the dynamical instability of BP. Once this instability occurs, the BP becomes not convergent, and hence it becomes not possible to obtain the estimator by BP. This implies the RSB region should be avoided from the viewpoint of estimation. 

\begin{figure}[htbp]
\centering
\includegraphics[width=0.49\textwidth]{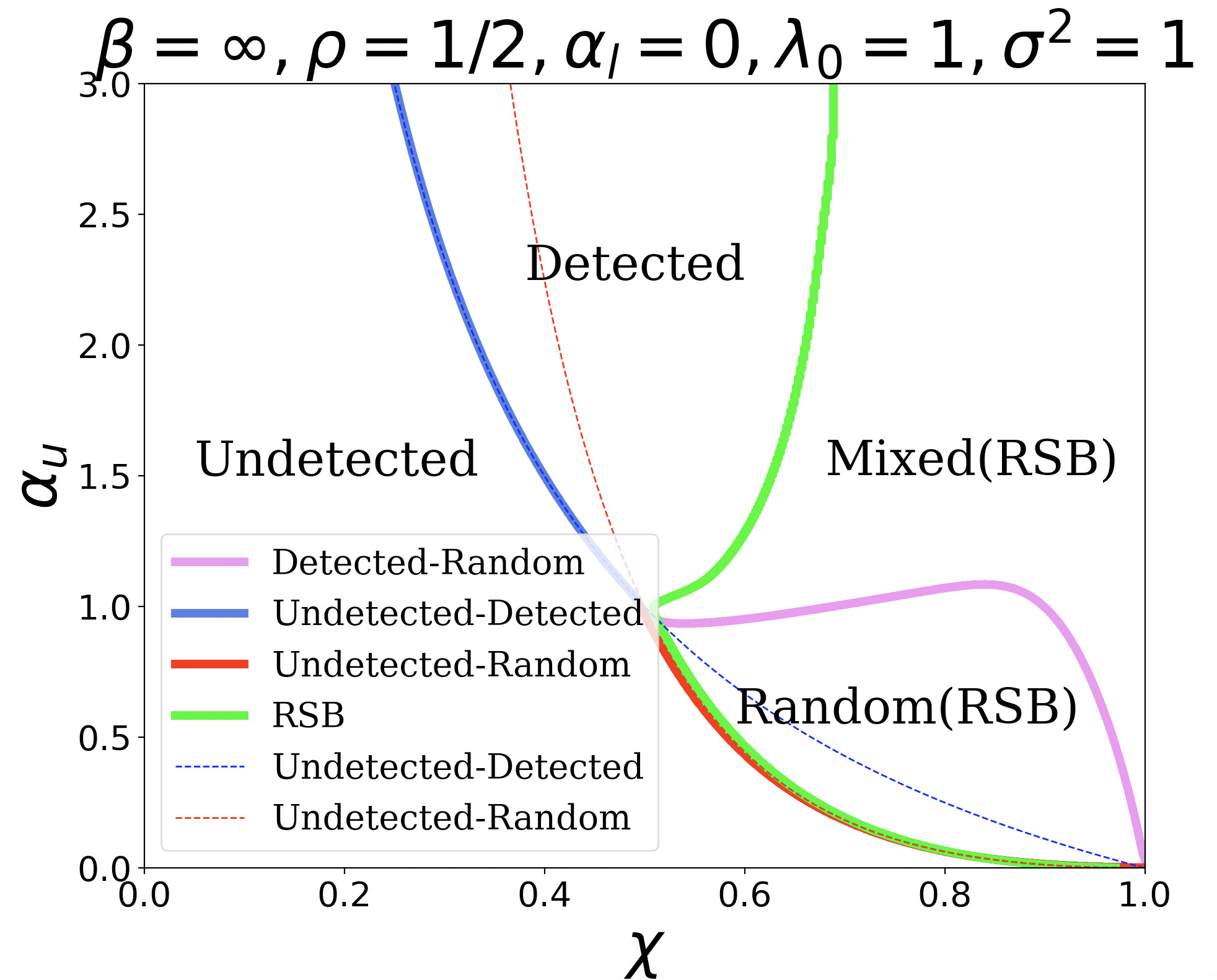}
\includegraphics[width=0.49\textwidth]{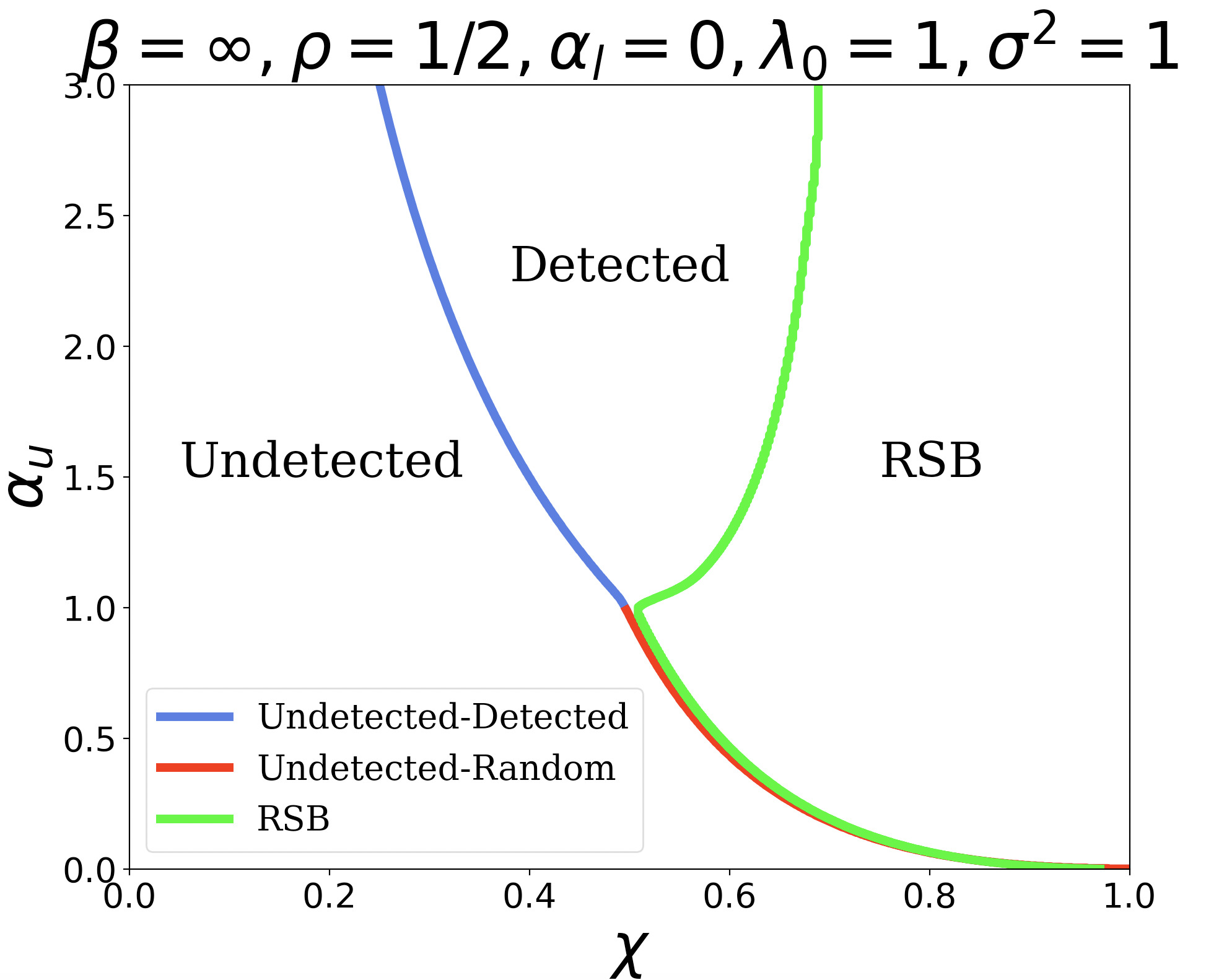}
\caption{Phase diagram of RMLE at $\rho=1/2, \alpha_l=0$ with $\lambda_0=\sigma^2=1$.}
\Lfig{phase transition baseline}
\end{figure}

Summarizing the analysis so far, we can draw the phase diagram in the $\chi$--$\alpha_u$ plane as shown in \Rfig{phase transition baseline}. In the left panel, the bottom-left region corresponds to the undetected phase. The phase boundary from the undetected phase to the detected phase is given by the blue solid line, while that to the random phase is depicted by the red solid line; the dashed blue and red curves correspond to \Req{CC_U-D} and \Req{CC_U-R}, respectively, and are given as a guide to the eye. The green solid line corresponds to \Req{at-line}, and the right area to this line exhibits RSB. It can be observed that there is a transition from the detected phase to another phase exhibiting RSB, and we call it the mixed phase since it is expected to have nonzero $k_*$ even with involving RSB. The solid pink line corresponds to \Req{CC_D-R} but this is not a precise boundary since RSB is not correctly taken into account in the RSB region in our analysis as we mentioned above. Hence, hereafter we neglect the difference between the mixed and random phases and regard those regions as a single RSB region. Accordingly, we rewrite the left panel to the right one. This right panel is set to be our baseline phase diagram and will be compared to phase diagrams at other parameter values.

\subsubsection{Impact of labeled data, label imbalance, and SNR}\Lsec{sub_IPPD} 
The baseline phase diagram has three distinct phases. We here investigate how these phases change when increasing $\alpha_l$ from zero and change $\rho$ from the balanced value $1/2$. 

Let us start by investigating the effect of the labeled dataset by introducing finite $\alpha_l$. Although our formulation allows us to study any $\alpha_l$ case, we are particularly interested in the small $\alpha_l$ region because it is a plausible situation in SSL that only a limited number of labeled data points are available.

In the left panel of \Rfig{phase transition rho and alpha_l}, we give a simultaneous plot of phase boundaries with different values of $\alpha_l$; the other parameters are set to be the same as \Rfig{phase transition baseline} for a clear comparison with the baseline. 

\begin{figure}[htbp]
\centering
\includegraphics[width=0.49\textwidth]{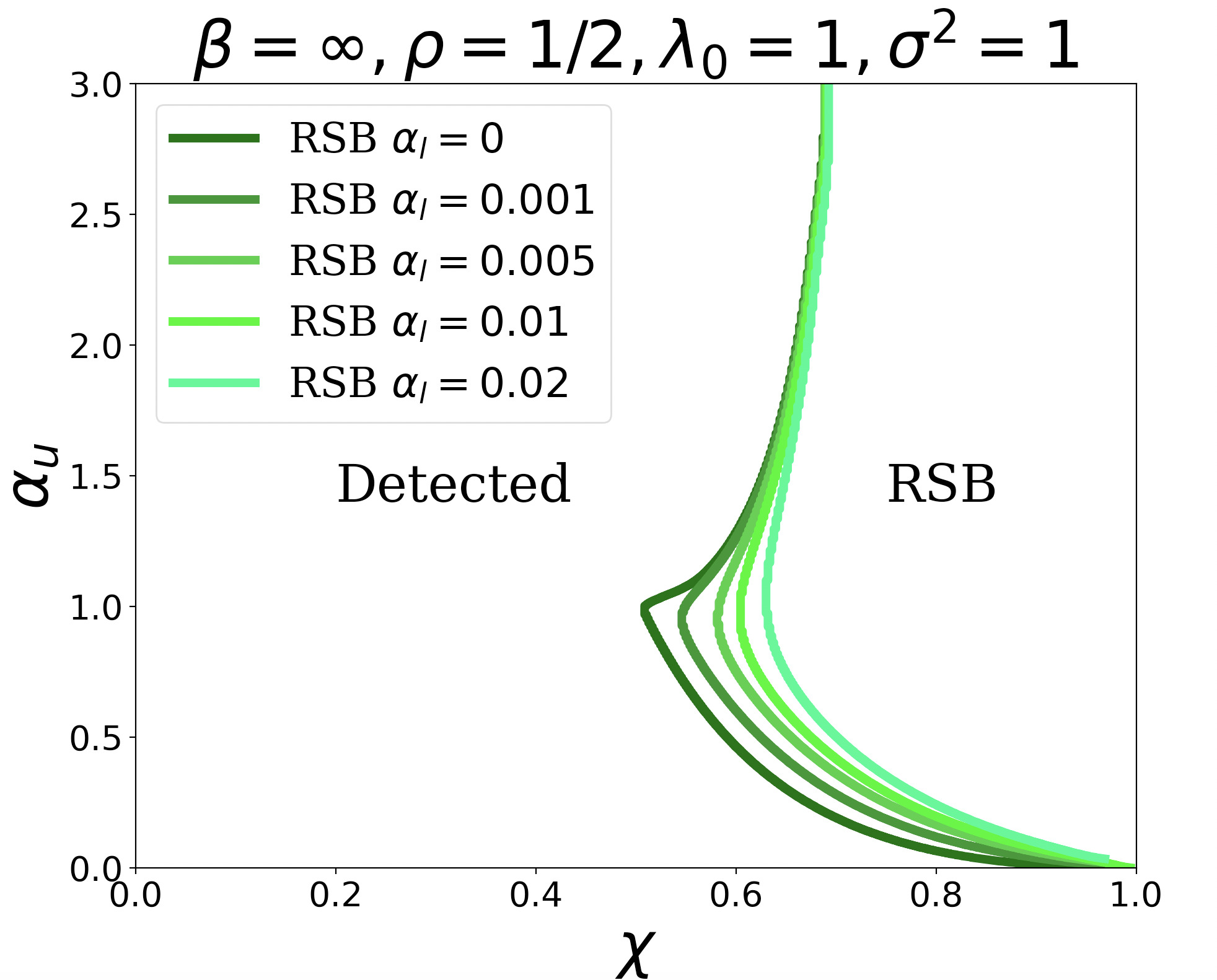}
\includegraphics[width=0.49\textwidth]{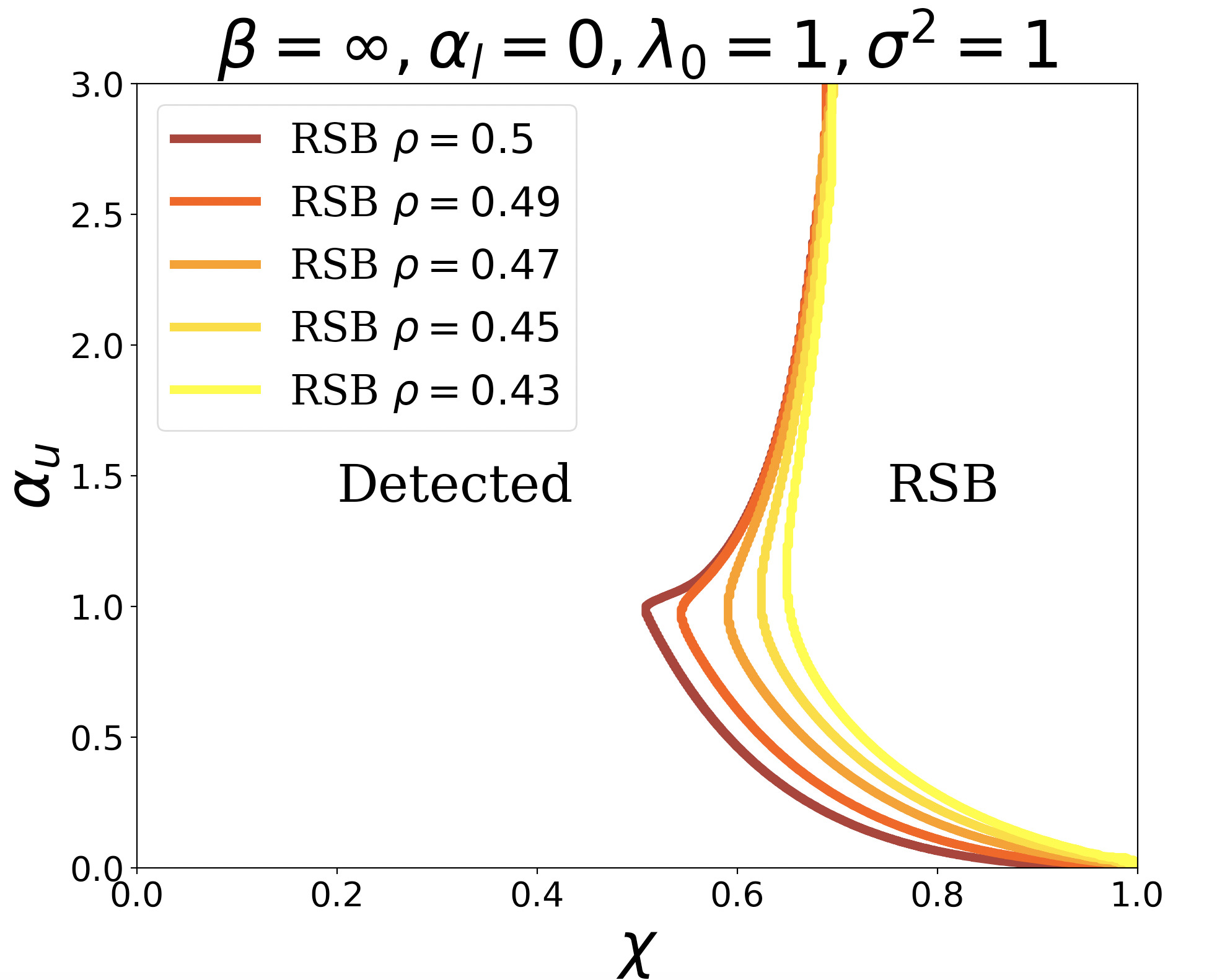}
\caption{\label{fig:fg} Impact of labeled data $\alpha_l$, label imbalance $\rho$ on the phase diagram}
\Lfig{phase transition rho and alpha_l}
\end{figure}

The undetected phase disappears by introducing finite $\alpha_l$. This is natural because the labeled data can break the symmetry between the two label values. It is also observed that the RSB region tends to be significantly reduced as $\alpha_l$ grows. This is also nice news since the RSB phase is not preferable from the viewpoint of estimation as discussed above. These highlight the significant positive effect of labeled data, even if its size is small. 

Meanwhile, the right panel of \Rfig{phase transition rho and alpha_l} shows the counterpart of the left panel when changing $\rho$ instead of $\alpha_l$. A deviation of $\rho$ from $1/2$ makes the undetected phase disappear, which is again reasonable because the label imbalance breaks the symmetry. The larger imbalance reduces the RSB region more, which is very akin to the impact of $\alpha_l$.

\begin{figure}[htbp]
\centering
\includegraphics[width=0.49\textwidth]{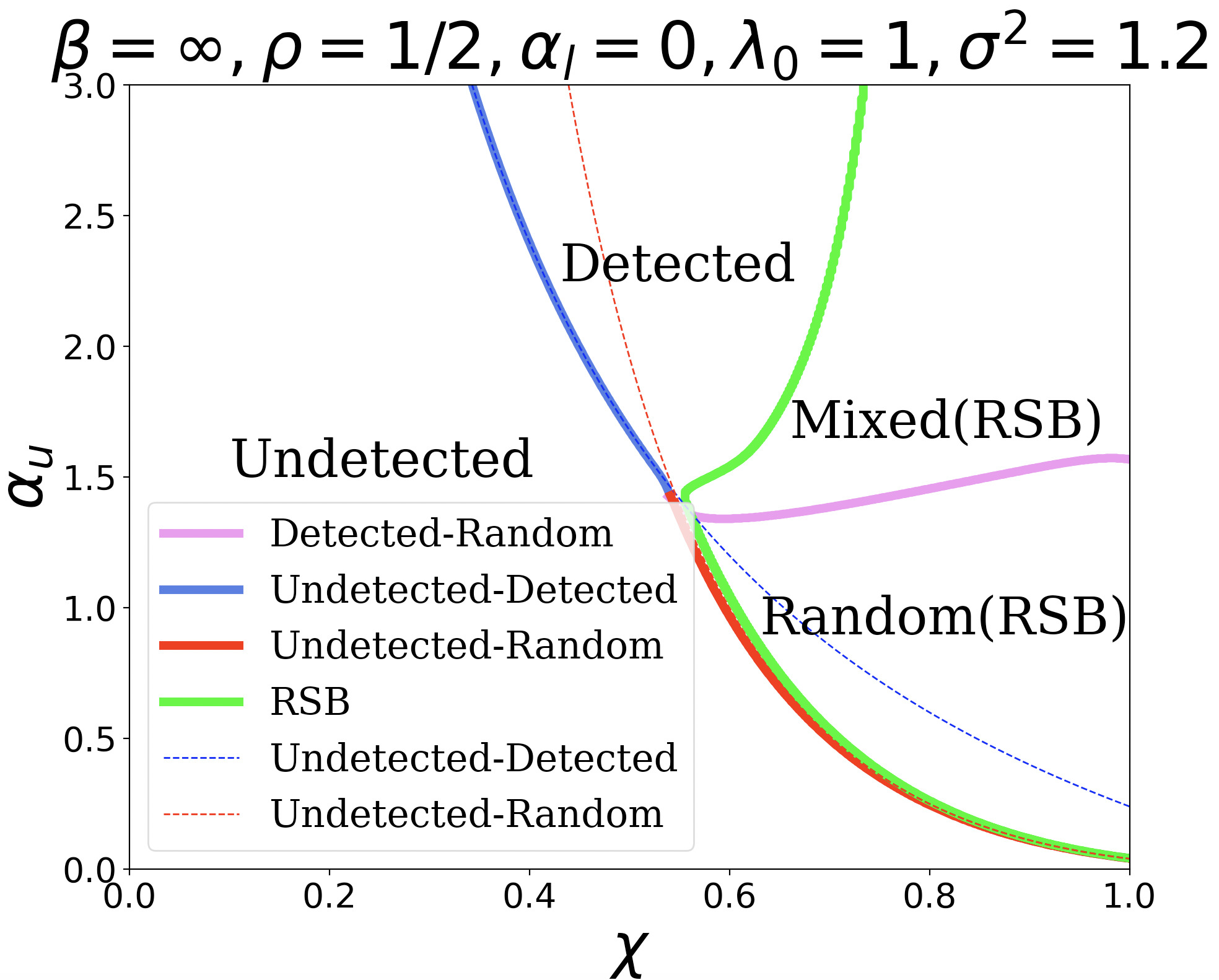}
\includegraphics[width=0.49\textwidth]{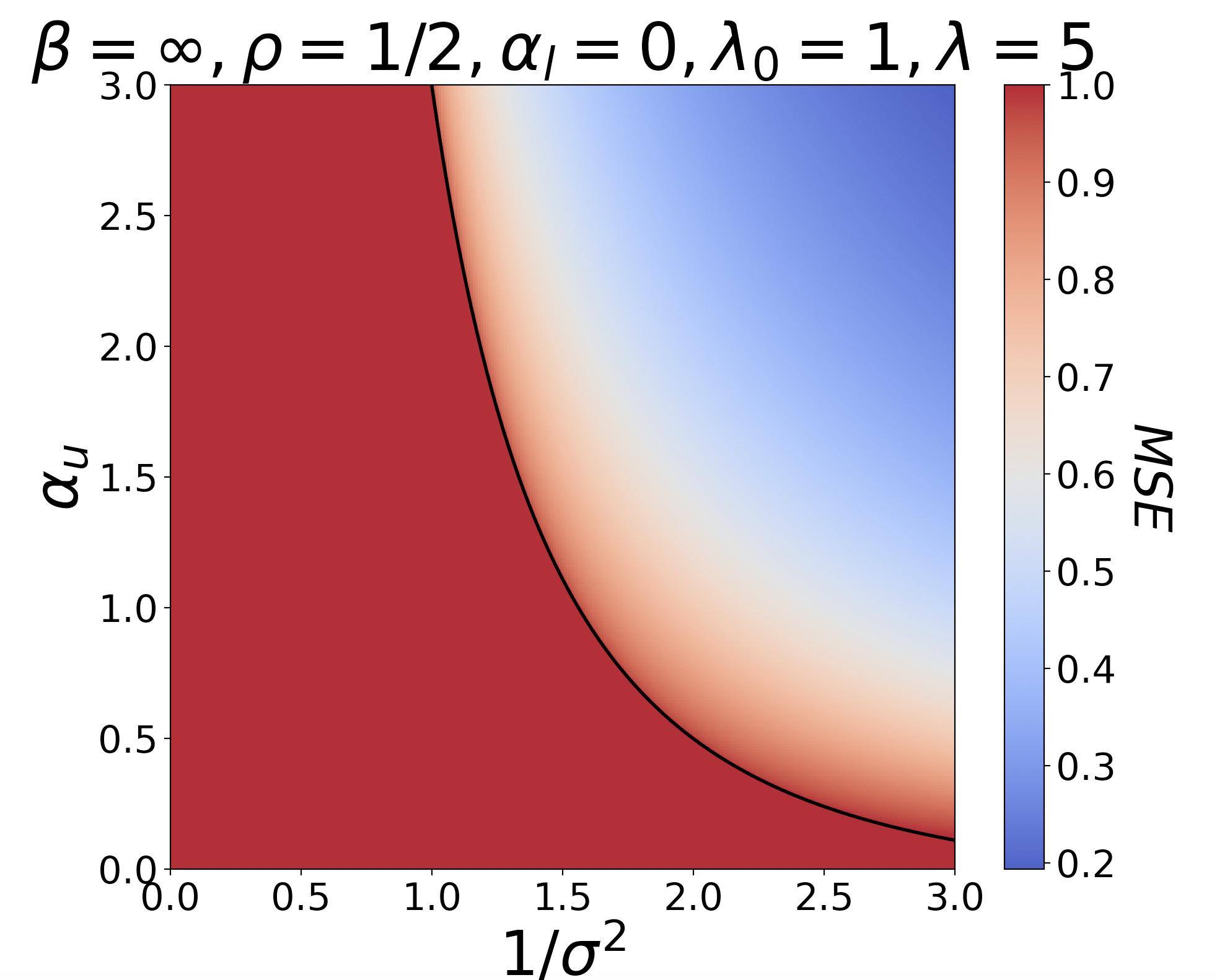}
\caption{\label{fig:fg} Impact of SNR $1/(\lambda_0 \sigma^2)$ on the phase diagram}
\Lfig{phase transition SNR}
\end{figure}

As a further investigation, in \Rfig{phase transition SNR}, we present phase diagrams illustrating the impact of change of $\sigma^2$. The left panel is the phase diagram corresponding to the left panel of baseline diagram \Rfig{phase transition baseline}, though $\sigma^2$ is increased from the baseline. \textcolor{black}{The ratio $1/(\lambda_0 \sigma^2)$ represents SNR,} and hence this increase of $\sigma^2$ corresponds to the reduction of SNR. By this SNR reduction, the undetected phase becomes larger while the detected and RSB phases get smaller. In the right panel of the same figure, we depict the relationship between $\alpha_u$ and SNR: the parameters are set to be $\alpha_l=0,\ \rho=1/2,\ \lambda_0=1$ and $\lambda=5$. The red region corresponds to the undetected phase on which MSE is kept unity, while the right upper region showing a gradation from red to blue is the detected phase. The phase boundary is depicted by the black curve whose analytical expression is given by
\be
\alpha_u =\Big(\big(\lambda-\lambda_0\big)\sigma^2-1\Big)\lambda_0\sigma^2.\Leq{eq66}
\ee
This is derived by combining the positive branch of \Req{chi_undetected} and \Req{CC_U-D}. These provide a clear visualization of how these parameters influence the estimation.

\subsection{Phase diagram of the Bayesian approach}\Lsec{sub_BO} 
In this section, we explore the phase diagram for the Bayesian approach. In parallel to the RMLE case, we start from the baseline case with $\rho=1/2$ and $\alpha_l=0$ and then investigate how labeled data, label imbalance, and SNR influence the phase diagram. 

\subsubsection{The baseline case}\Lsec{sub_UEC_BA} 
By a similar analysis to \Rsec{sub_UEC}, the phase diagram for the Bayesian case can be derived. Let us start by summarizing the critical conditions below.
 
\begin{figure}[htbp]
\centering
\includegraphics[width=0.5\textwidth]{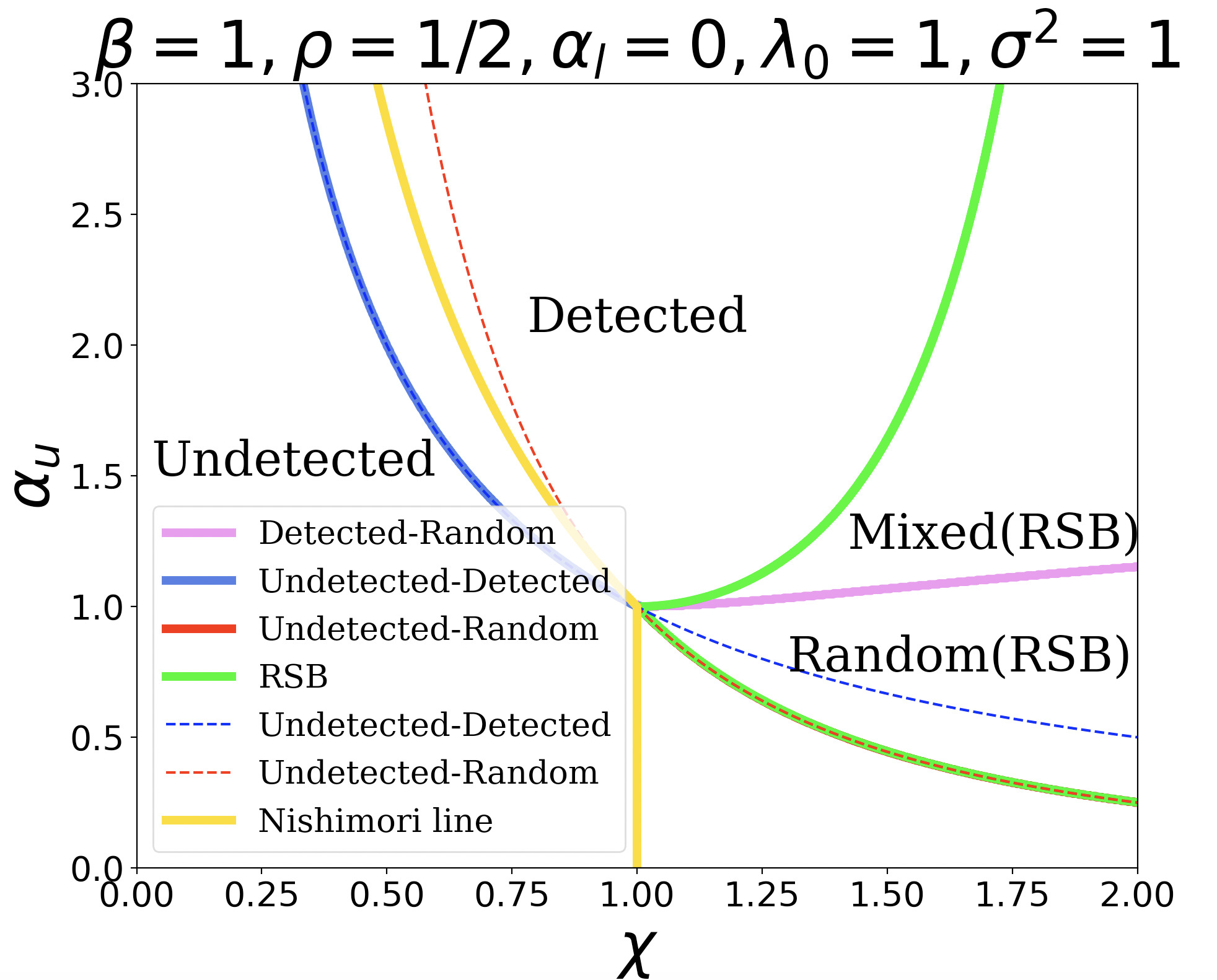}
\caption{\label{fig:fg} Phase diagram of the Bayesian approach at $\rho=1/2, \alpha_l=0$. The lines are depicted in the same rule as the left panel of \Rfig{phase transition baseline}, though there is an additional line (yellow solid curve) describing the BO subspace or Nishimori line.}
\Lfig{phase transition BA}
\end{figure}

The three critical conditions, between the undetected and detected phases, the undetected and random phases, and the detected and random phases, are given by
\be
&&
\chi = \frac{\sigma^4\lambda_0}{\alpha_u},\Leq{u_d_BO}
\\ &&
\chi=\frac{\sigma^2}{\sqrt{\alpha_u}},
\\ &&
1 =\frac{\alpha_u \chi}{\lambda_0}  \int Dz \ \tilde{T}\bigg(\sqrt{\frac{v_*}{\sigma^2}} z\bigg),
\ee
respectively. The RSB critical condition can also be derived as the way of deriving \Req{at-line}. The result is
\be
1 = \frac{\alpha_u \chi^2}{\sigma^4}\int Dz\ \mathrm{sech}^4\bigg(\frac{k_*}{\lambda_0\sigma^2} + \sqrt{\frac{k_*^2/\lambda_0+v_*}{\sigma^2}} z\bigg).\Leq{BA_AT_line}
\ee
Summarizing these results, in \Rfig{phase transition BA}, we draw a phase diagram corresponding to the left panel of \Rfig{phase transition baseline}.

In the Bayesian case, we are particularly interested in the BO setting $\lambda=\lambda_0$. The relation $\lambda=\lambda_0$ is implicitly encoded into $\chi$ by the strategy explained in \Rsec{sub_Conv}, and forms a nontrivial specific subspace in the phase diagram. In \Rfig{phase transition BA}, the BO subspace is depicted by the yellow solid line, and we call this line {\it Nishimori line} according to the physics terminology~\cite{iba1999nishimori,nishimori2001statistical}.

\subsubsection{Impact of labeled data, label imbalance, and SNR}\Lsec{sub_IPPD_BA} 
We next investigate the influence of the labeled data, the label imbalance, and the SNR change in the Bayesian approach. The resultant phase diagrams corresponding to \Rfigs{phase transition rho and alpha_l}{phase transition SNR} are summarized in \Rfig{phase transition BA-2}.

\begin{figure}[htbp]
\centering
\includegraphics[width=0.32\textwidth]{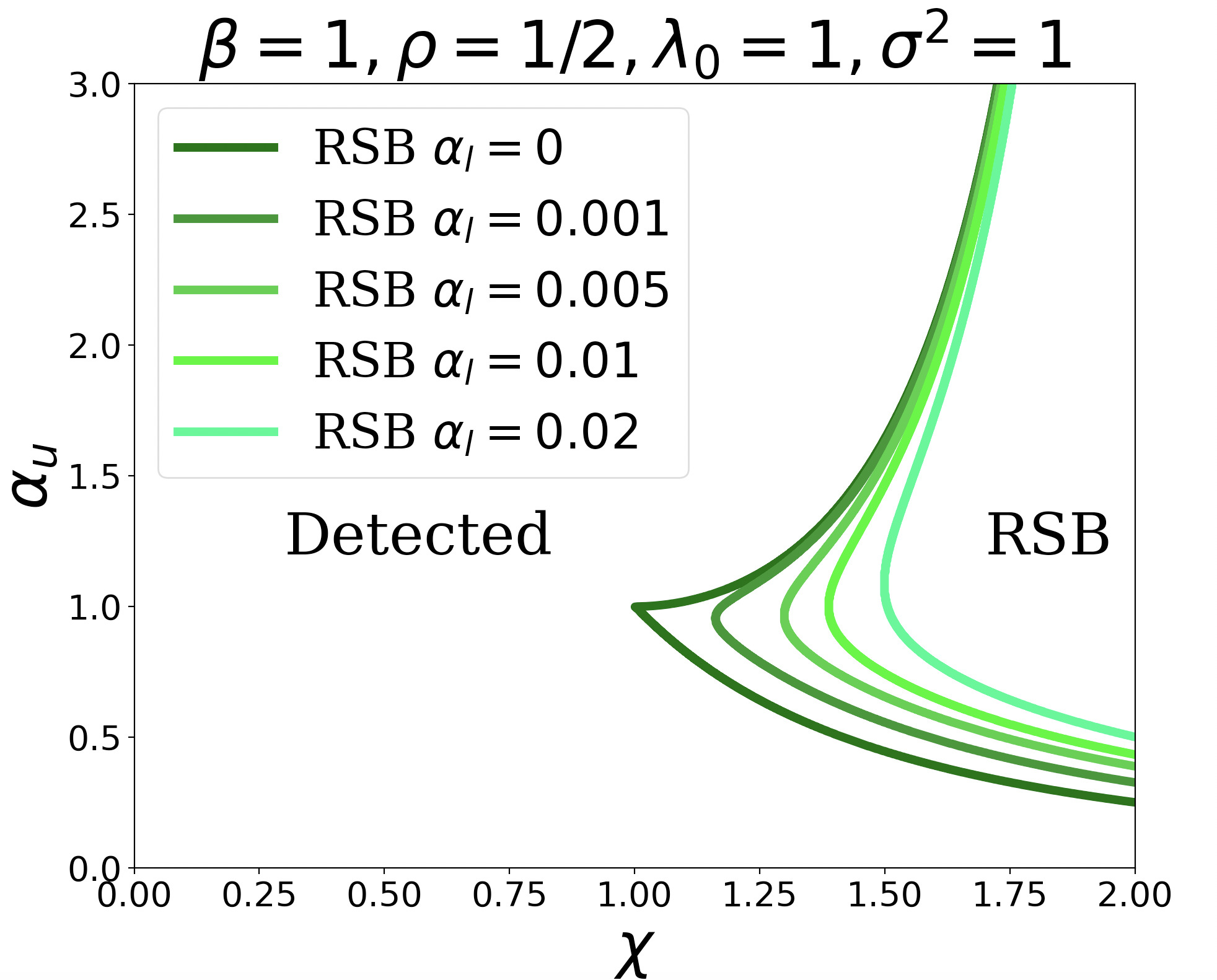}
\includegraphics[width=0.32\textwidth]{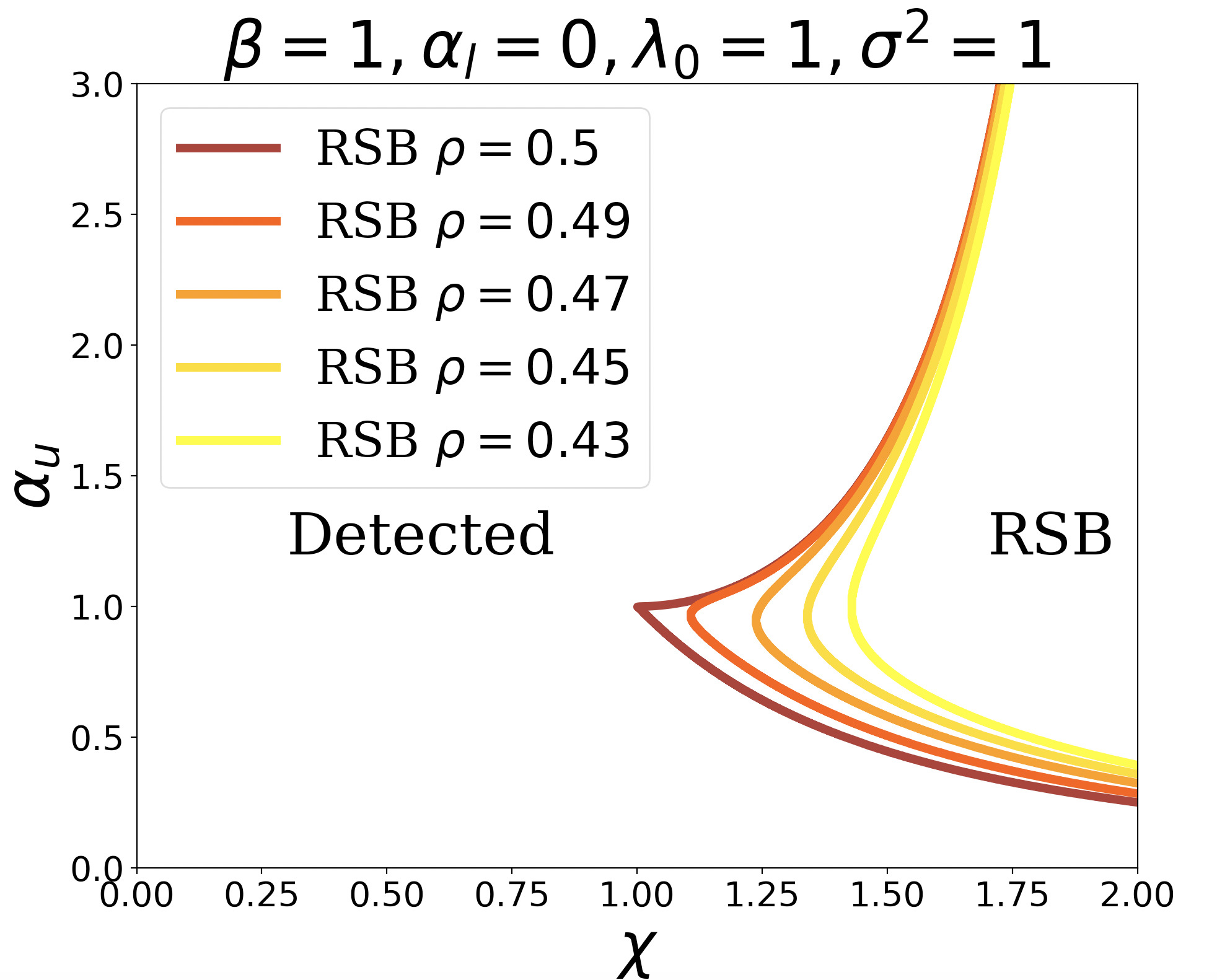}
\includegraphics[width=0.32\textwidth]{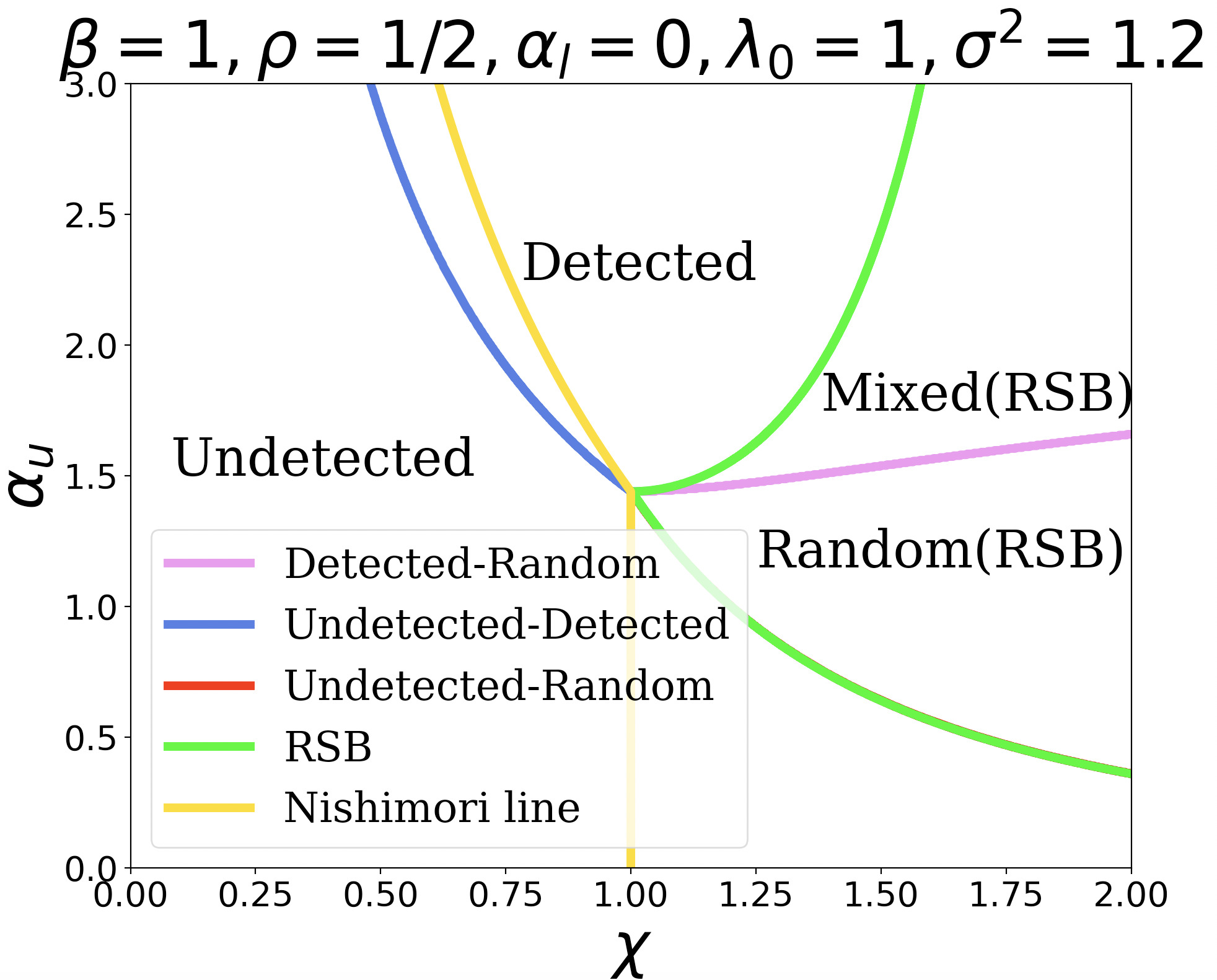}
\caption{\label{fig:fg} Impact of labeled data $\alpha_l$, label imbalance $\rho$, and SNR on the phase diagram. The left and middle panels are counterparts of \Rfig{phase transition rho and alpha_l}. The right panel shows a smaller SNR case compared to the baseline case in \Rfig{phase transition BA}.} 
\Lfig{phase transition BA-2}
\end{figure}

The introduction of labeled data with finite $\alpha_l$ and deviations in label imbalance $\rho$ from $1/2$ in the Bayesian approach lead to the disappearance of the undetected phase and a reduction of the RSB region, as shown in the left and middle panels. This emphasizes the positive impact of even a small amount of labeled data or the label imbalance on estimation as in the RMLE case. In the right panel, we investigate the effects of SNR. The SNR reduction (an increase of $\sigma^2$) enlarges the undetected phase while shrinking the detected and RSB phases, again aligning with the observations in the RMLE case.

\subsubsection{The BO and model mismatch cases}\Lsec{sub_RSB_MM} 
The BO setting is a special case in the Bayesian approach and has been extensively studied in the literature \cite{ lelarge2019asymptotic,tanaka2013statistical, tanaka2002statistical} since it yields the best performance and can thus be a nice reference for other formulations/algorithms. In this subsection, we follow this direction and investigate MSE over a wide range of parameters in the BO setting. In addition, we also examine some non-BO situations ($\lambda \neq \lambda_0$) to see the difference from the BO setting. It is known that RSB, which affects the convergence of AMP, can occur only in the non-BO case, and we are particularly interested in how RSB appears in the present model under the Bayesian treatment.

\begin{figure}[htbp]
\centering
\includegraphics[width=0.32\textwidth]{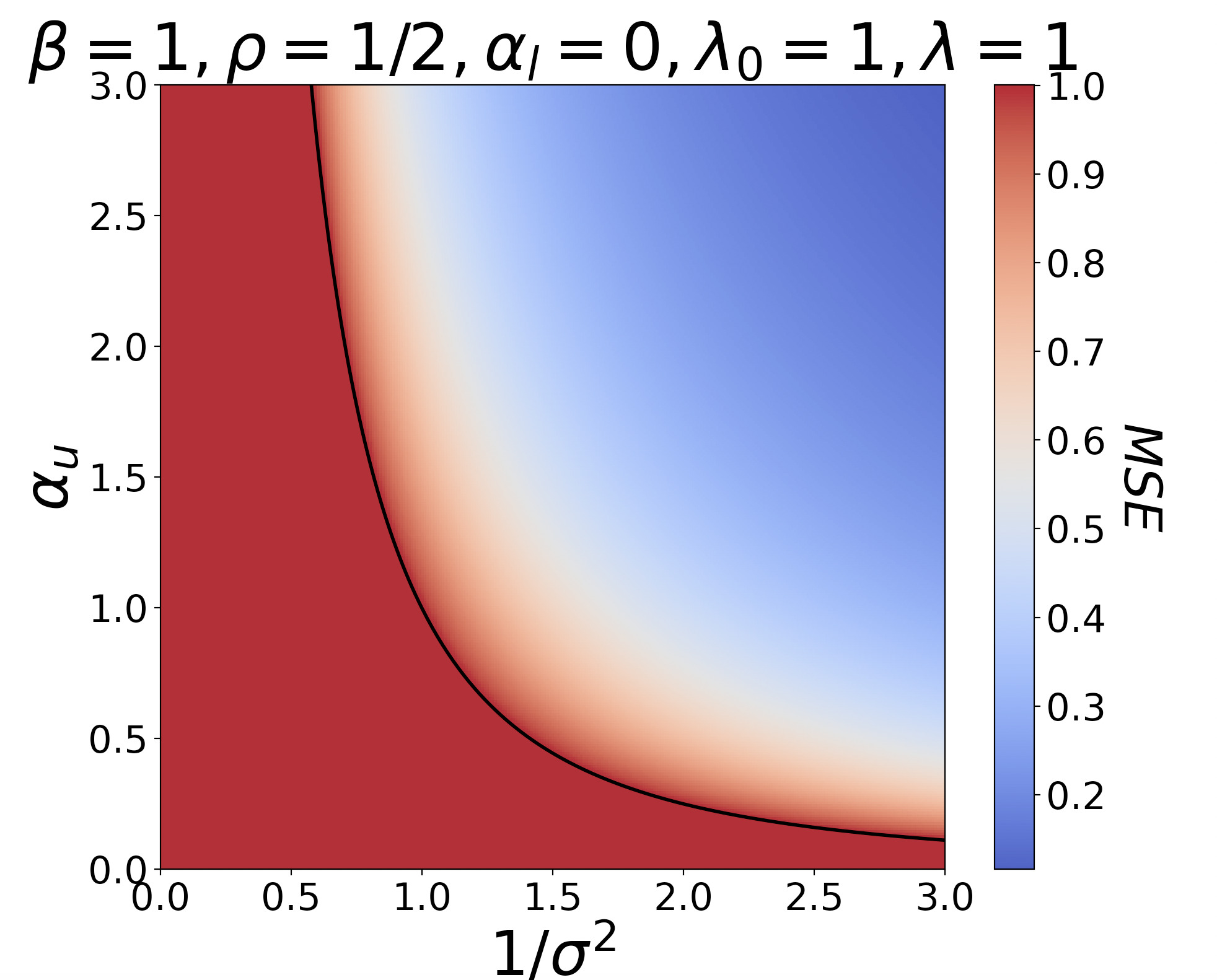}
\includegraphics[width=0.32\textwidth]{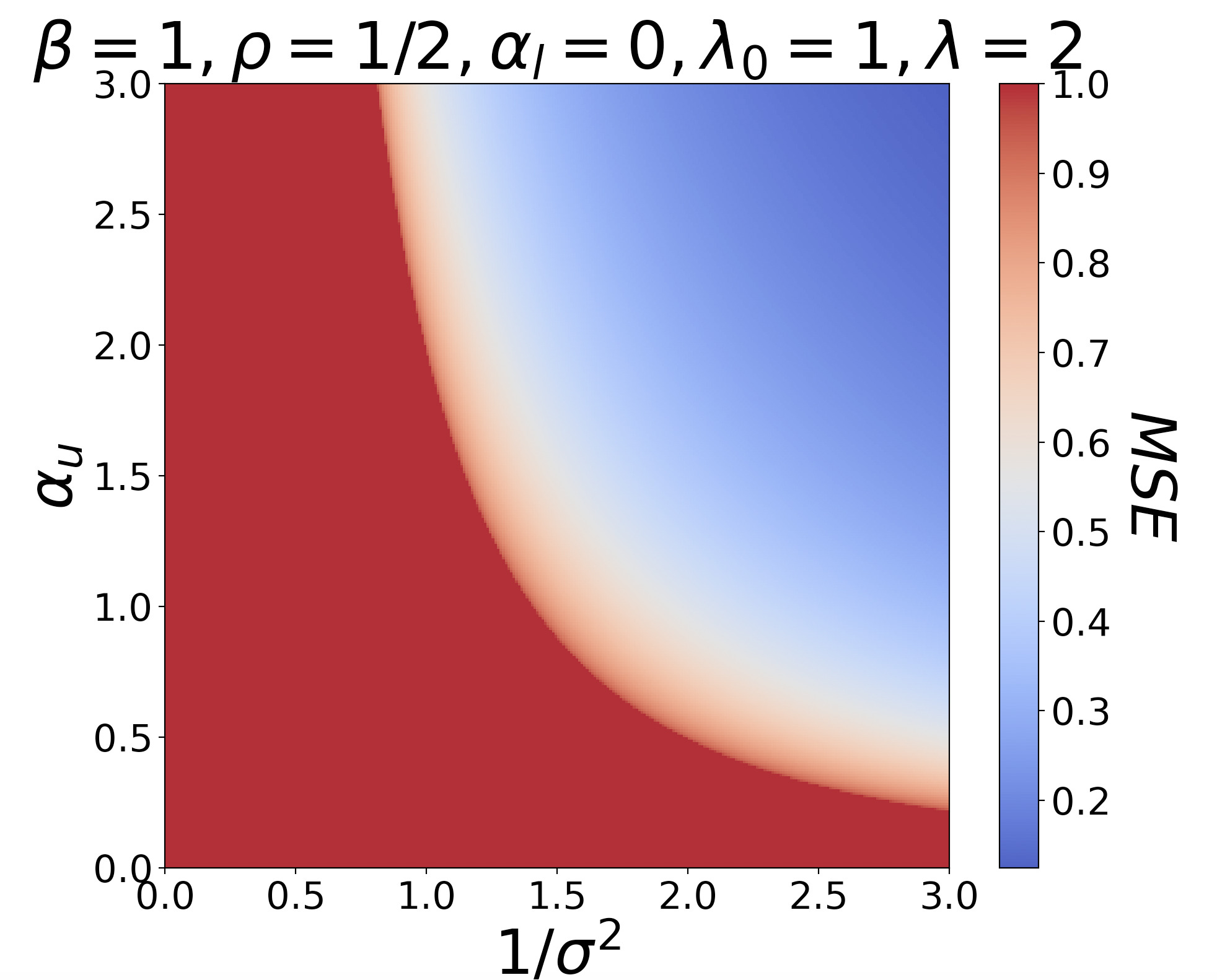}
\includegraphics[width=0.32\textwidth]{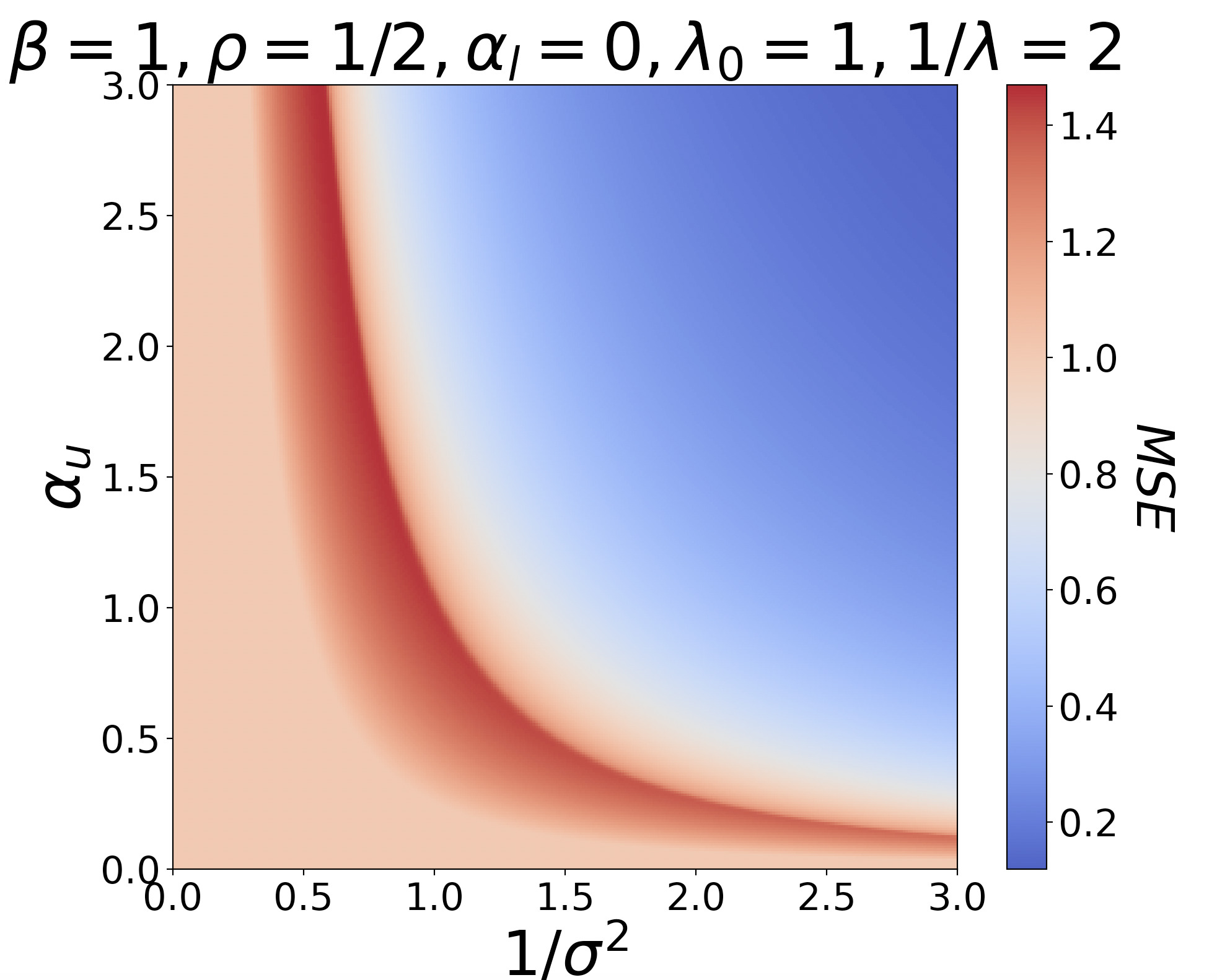}
\caption{\label{fig:fg} The MSE visual representations of the BO ($\lambda=\lambda_0$) case and two model mismatch cases are shown on the left, middle ($\lambda=2$), and right ($\lambda=0.5$) panels, respectively, at $\lambda_0=1, \rho=1/2, \alpha_l=0$. In the BO heatmap, a distinct black curve separates regions where the MSE values are 1 on the left and less than 1 on the right. Conversely, in the case of model mismatch of $\lambda=2$, the heatmap shows a larger region with MSE equal to 1 compared to the BO heatmap. In the other model mismatch $\lambda=0.5$, it exhibits three distinct regions from left to right: MSE $=$ 1, MSE $> 1$, and MSE $< 1$.}
\Lfig{phase transition BO and MM}
\end{figure}

Let us first study the BO case ($\lambda_0=\lambda$) corresponding to the Nishimori line in \Rfig{phase transition BA}. We plot the MSE heatmap in the left panel of \Rfig{phase transition BO and MM}. The parameters are set as $\lambda=\lambda_0=1$, $\rho=1/2$, and $\alpha_l=0$. The red region corresponds to the undetected phase with $\varepsilon_{MSE}=1$. In contrast, the upper-right region transitioning from red to blue is the detected phase, where MSE gradually decreases by increasing SNR or $\alpha_u$. The phase boundary is characterized by the black curve and is given by
\be
    \alpha_u = \Big(\frac{1}{\lambda_0\sigma^2}\Big)^{-2} = \big(\text{SNR}\big)^{-2},\Leq{eq68}
\ee
which is derived from the combination of \Reqs{se1b}{u_d_BO}. Compared to the left panel of \Rfig{phase transition SNR} (RMLE), the BO result shows a smaller red region, showing superiority to RMLE.

For comparison, we also plot the MSE heatmap of the model mismatch with $\lambda\in\{2, 0.5\}$ in the middle and right panels of \Rfig{phase transition BO and MM}, respectively. When $\lambda=2$, the heatmap shows a larger region with $\varepsilon_{MSE}=1$ compared to the BO heatmap. When $\lambda=0.5$, the heatmap is divided into three regions with $\varepsilon_{MSE}=1,>1,<1$ from left to right. The middle red region has $\varepsilon_{MSE}>1$. A portion of the middle red region, where MSE is greater than 1, is unstable due to its overlap with the RSB phase. Another portion of the middle red region also has an MSE greater than 1 but lies within the RS phase.

\begin{figure}[H]
\centering
\includegraphics[width=0.49\textwidth]{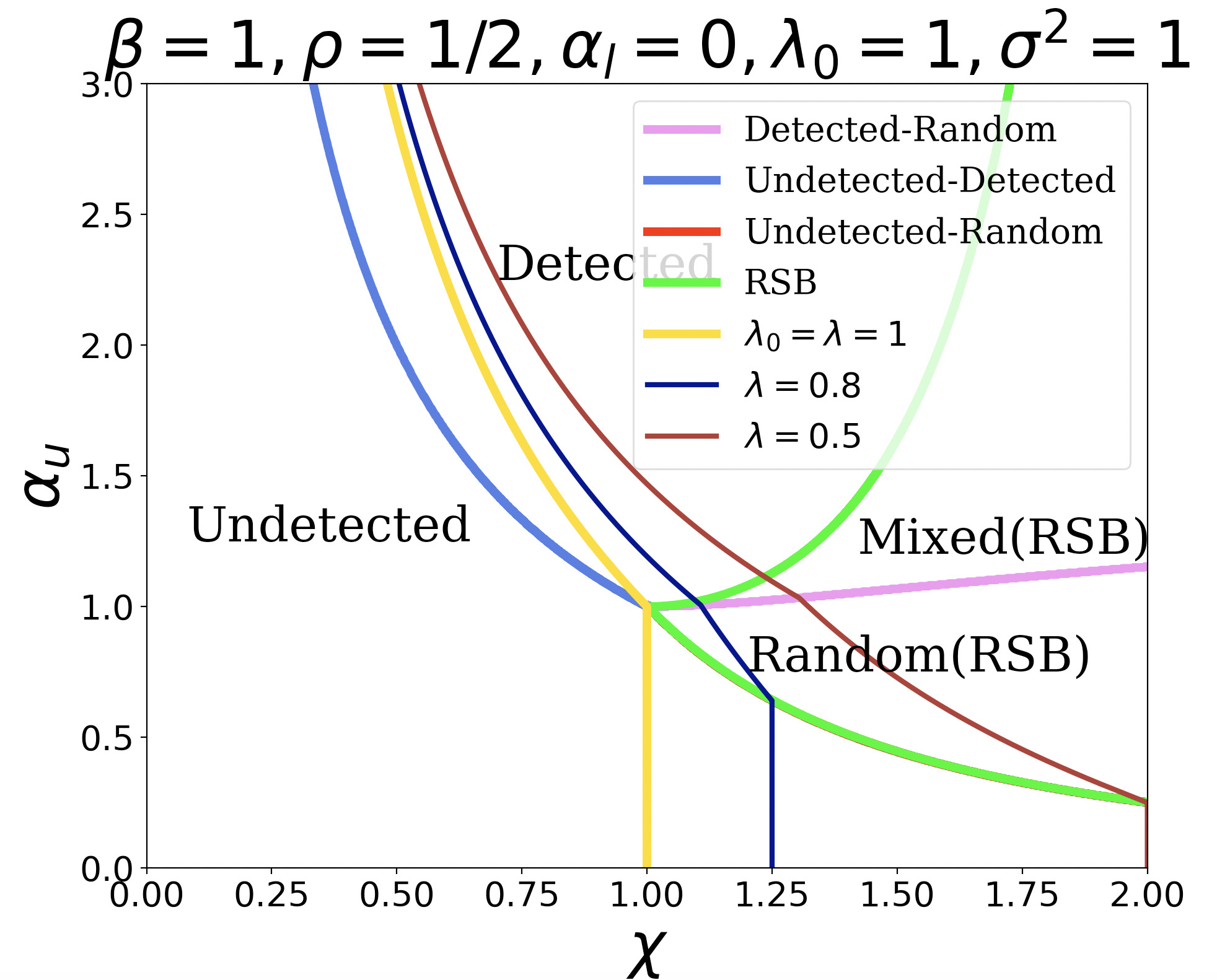}
\includegraphics[width=0.49\textwidth]{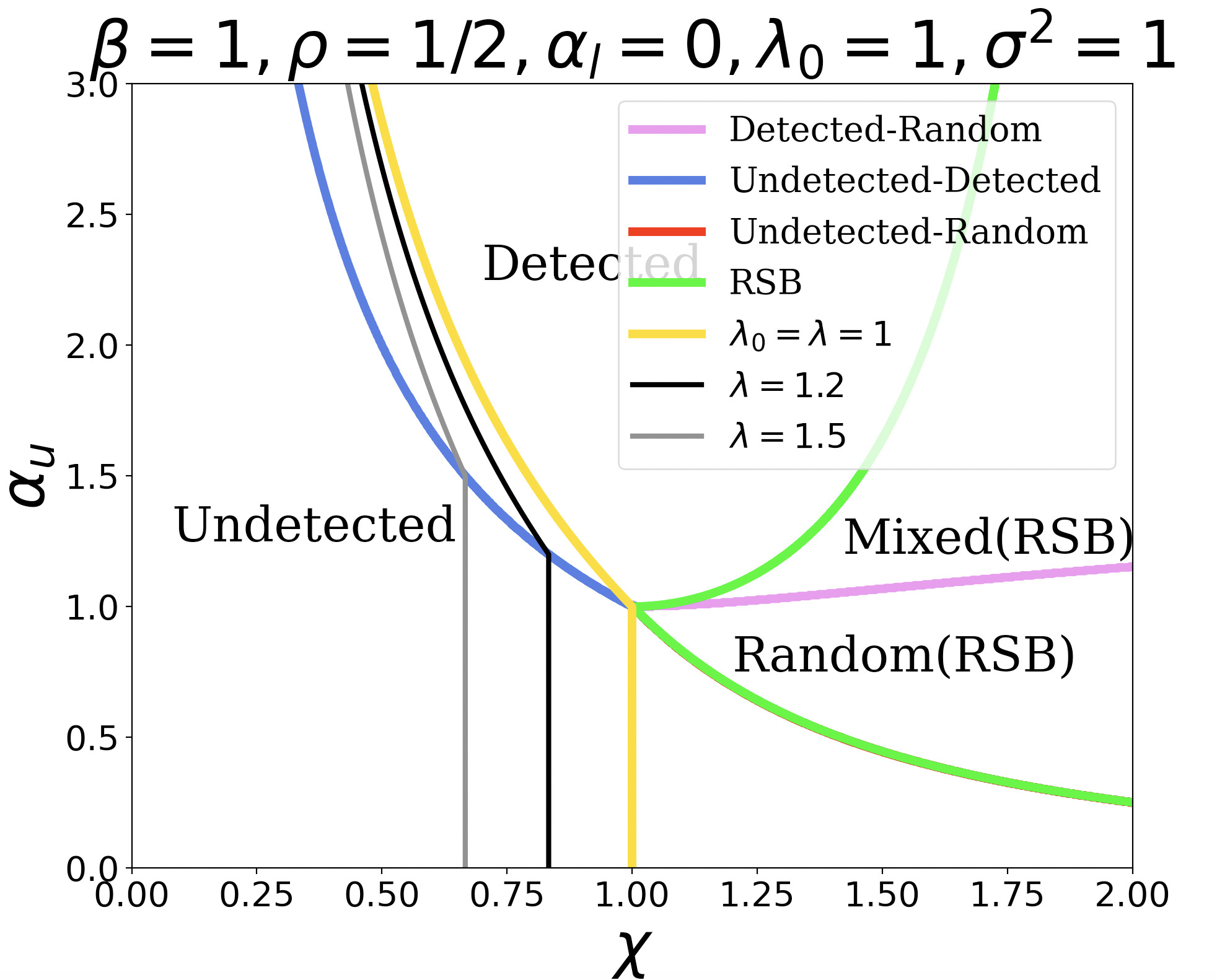}
\\
\includegraphics[width=0.49\textwidth]{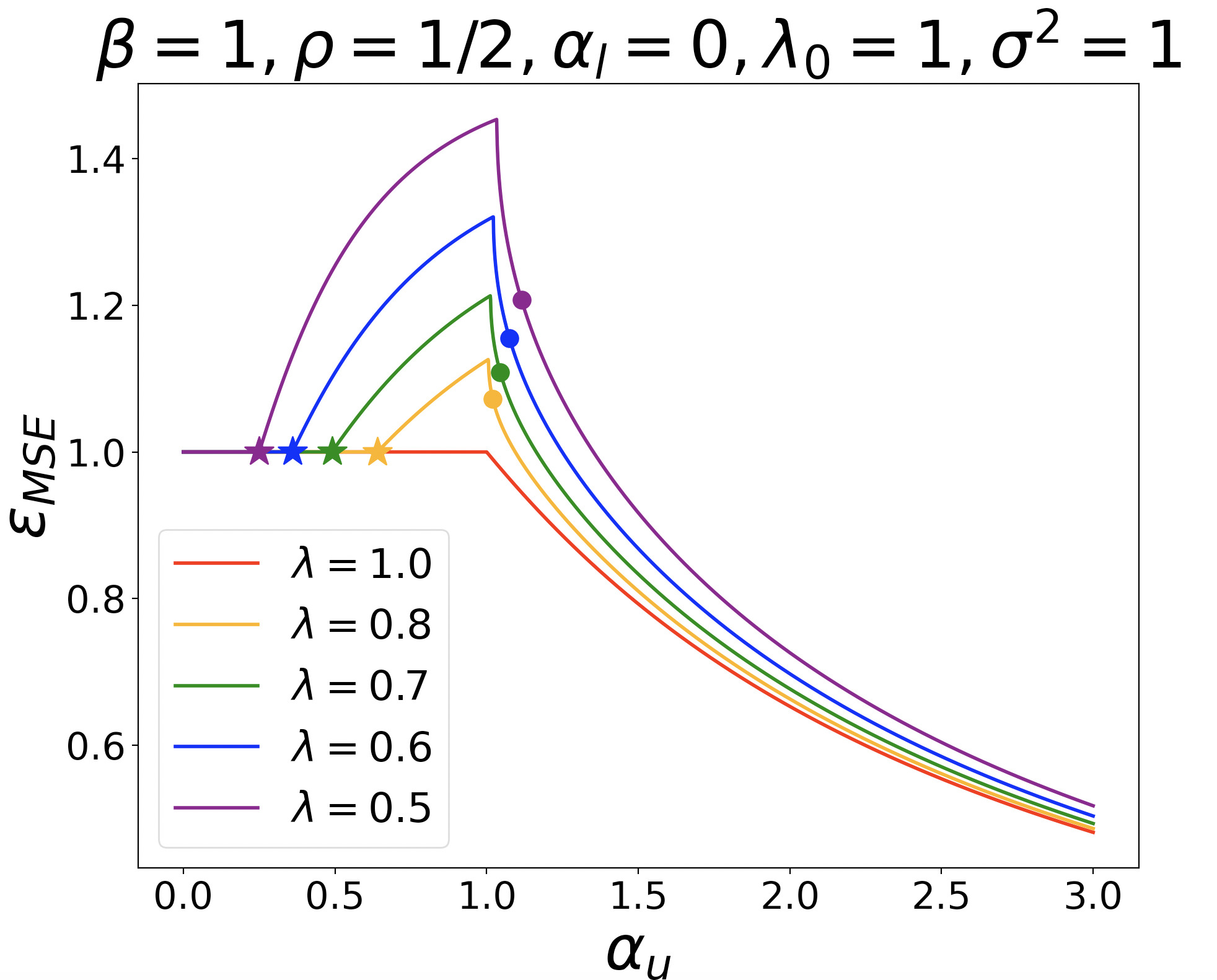}
\includegraphics[width=0.49\textwidth]{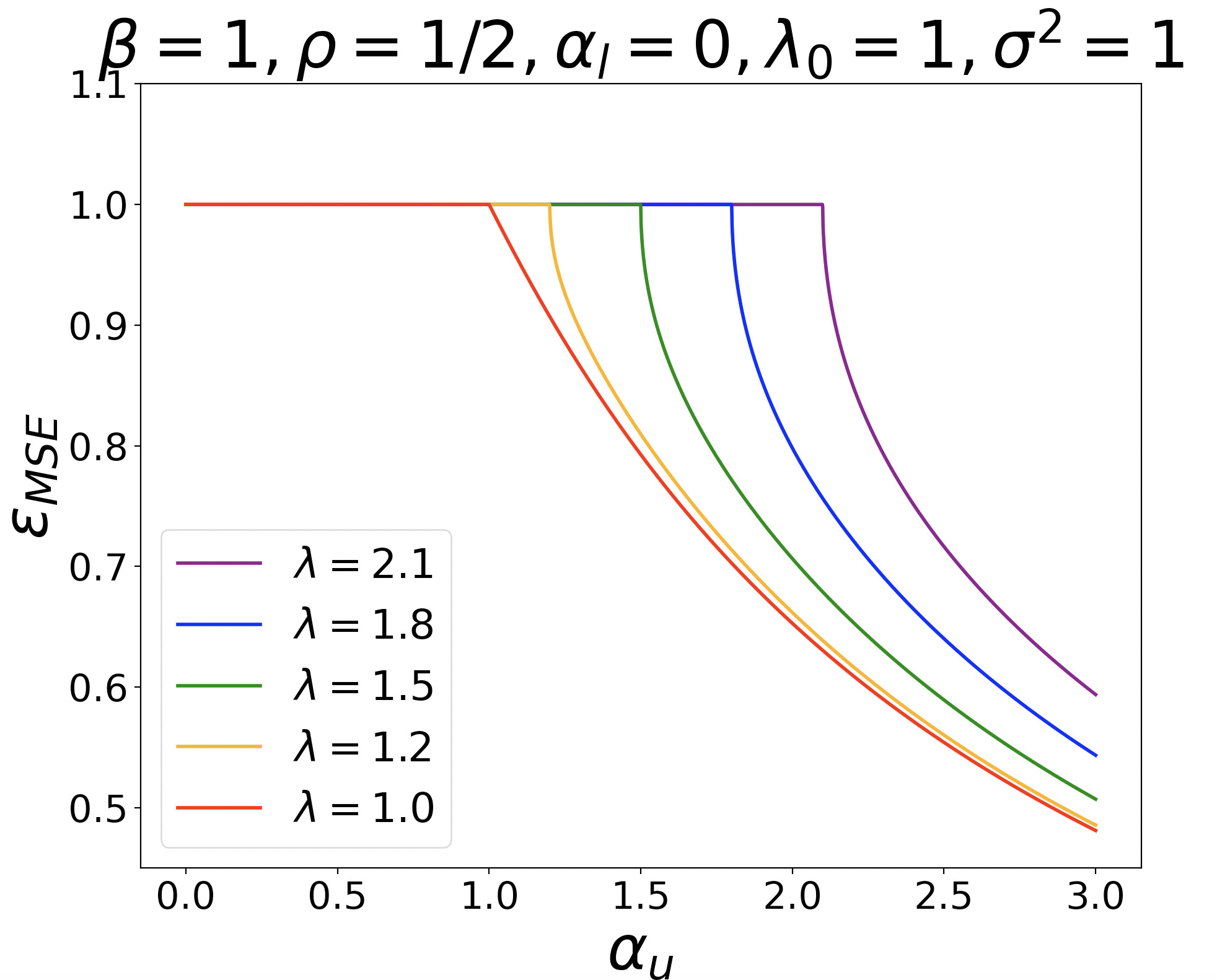}
\caption{\label{fig:fg} Specific $\lambda$ values ($\neq \lambda_0$) for the model mismatch and their performance. The upper panels show the curves for different $\lambda$ values in \Rfig{phase transition BA}. The upper left panel corresponds to $\lambda > \lambda_0$, while the upper right panel corresponds to $\lambda < \lambda_0$. The bottom panels demonstrate the performance of AMP under $\lambda > \lambda_0$ (left) and $\lambda < \lambda_0$ (right).}
\Lfig{RSB BO}
\end{figure}

Next in \Rfig{RSB BO}, we investigate the locations of different $\lambda$ values on the same phase diagram as \Rfig{phase transition BA} and show MSE plotted against $\alpha_u$. In the upper left panel, the curves for $\lambda = 0.5$ and $\lambda = 0.8$ (brown and navy, respectively) are located to the right of the Nishimori line. We observe that both of them intersect the RSB phase at certain values of $\alpha_u$, indicating the AMP instability in this region. In the upper right panel, the curves for $\lambda = 1.2$ and $\lambda = 1.5$ (black and grey, respectively) are located to the left of the Nishimori line. This suggests the absence of RSB. The bottom panels show MSE plotted against $\alpha_u$ for different $\lambda$. In the bottom left panel, we examine the case where $\lambda < \lambda_0$. Our findings show that MSE exceeds 1 at a certain point and then gradually decreases as $\alpha_u$ increases. The RSB critical point, defined as the value of $\alpha_u$ at which \Req{BA_AT_line} exactly holds and marked as ``$\star$'', shifts with different values of $\lambda$. This results in the AMP instability, causing MSE to immediately exceed 1. As $\alpha_u$ is further increased, the AMP algorithm remains unstable until $\alpha_u$ reaches the marked point with``$\bullet$". In the bottom right panel, we examine the case $\lambda > \lambda_0$. We observe that for different values of $\lambda$, MSE initially remains at 1. It then gradually drops at a certain value of $\alpha_u$. Notably, the farther $\lambda$ is from $\lambda_0$, the larger value of $\alpha_u$ becomes necessary for the dropping.

These results imply the AMP algorithm can exhibit stable performance in the non-BO cases when $\lambda > \lambda_0$. However, it may become unstable when $\lambda < \lambda_0$ in certain ranges of $\alpha_u$. When $\lambda_0$ is unknown, we can choose a relatively large value of $\lambda$ to make the AMP algorithm stable.

\subsection{Comparison of RMLE and the BO estimate}\Lsec{sub_CoRB} 
In this section, we compare the performances of RMLE and the BO estimate. To this end, in addition to MSE, we introduce GE. GE is a measure of how accurately the output values of unseen data can be predicted, and it is defined as \Req{ge}.

When we have an estimator of the parameter $\bm{w}_0$, $\hat{\bm{w}}$, it is reasonable to make our prediction according to the maximization of the conditional probability of $y$ given $\hat{\bm{w}}$ and $\bm{x}_{new}$, $p(y \mid \hat{\bm{w}},\bm{x}_{new})$ defined from \Reqs{eq2}{eq3}. The explicit formula is   
\be
\hat{y}_{new} &=& \operatorname*{argmax}_{y \in \{\pm 1\}} p(y \vert \hat{\bm{w}}, \bm{x}_{new})\Leq{ge0},\\
&=&
\left\{ \begin{array}{rcl}
1 & \mbox{if} & \frac{\rho}{1-\rho}\exp\big(\frac{2}{\sigma^2} \frac{\bm{x}_{new}\cdot\hat{\bm{w}}}{\sqrt{N}}\big)>1\\
-1 & \mbox{if} & \frac{\rho}{1-\rho}\exp\big(\frac{2}{\sigma^2} \frac{\bm{x}_{new}\cdot\hat{\bm{w}}}{\sqrt{N}}\big)<1
\end{array}
\right.
= \operatorname{sign}\Big(\frac{\hat{\bm{w}}\cdot \bm{x}_{new}}{\sqrt{N}}+b\Big)\Leq{ge2},\Leq{ge1}
\ee
where $b= \frac{\sigma^2}{2}\log\frac{\rho}{1-\rho}$. Assuming that $\hat{\bm{w}}$ has the overlap with $\bm{w}_0$ and the variance as $k_*$ and $v_*$ respectively, as assumed in Sec. 3.4, we can rewrite \Req{ge} as 
\be
     \textcolor{black}{\mathcal{E}_{GE}} &=& \rho Q\bigg(\frac{k_*/\lambda_0+b}{\sqrt{\sigma^2 (k_*^2/\lambda_0 +v_*)}}\bigg)+(1-\rho)Q\bigg(\frac{k_*/\lambda_0-b}{\sqrt{\sigma^2 (k_*^2/\lambda_0 +v_*)}}\bigg),\Leq{ge3}\\
    Q(x) &=& \int_x^\infty Dz.\Leq{ge4}
\ee
In the Bayesian case, the same formula \NReq{ge3} is also applicable for the evaluation of GE by replacing $k_*$ and $v_*$ with them of $\hat{\bm{w}}_{B}$ in the large dimensional limit. Hence, the SE equations \NReqs{SE_RMLE}{SE_BA} in combination with \Req{ge3} enable a systematic evaluation of GE. 

\begin{figure}[htbp]
\centering
\includegraphics[width=0.49\textwidth]{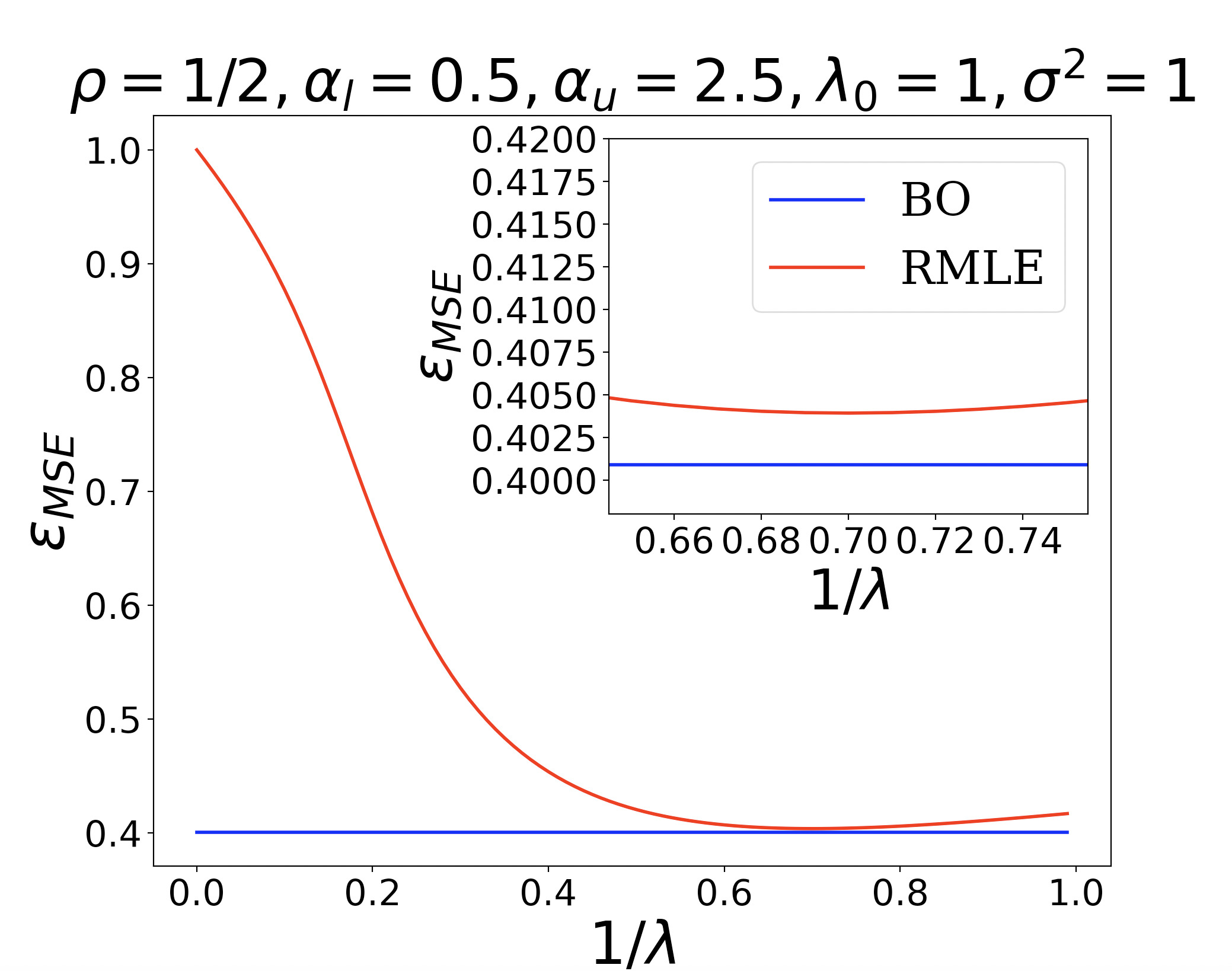}
\includegraphics[width=0.49\textwidth]{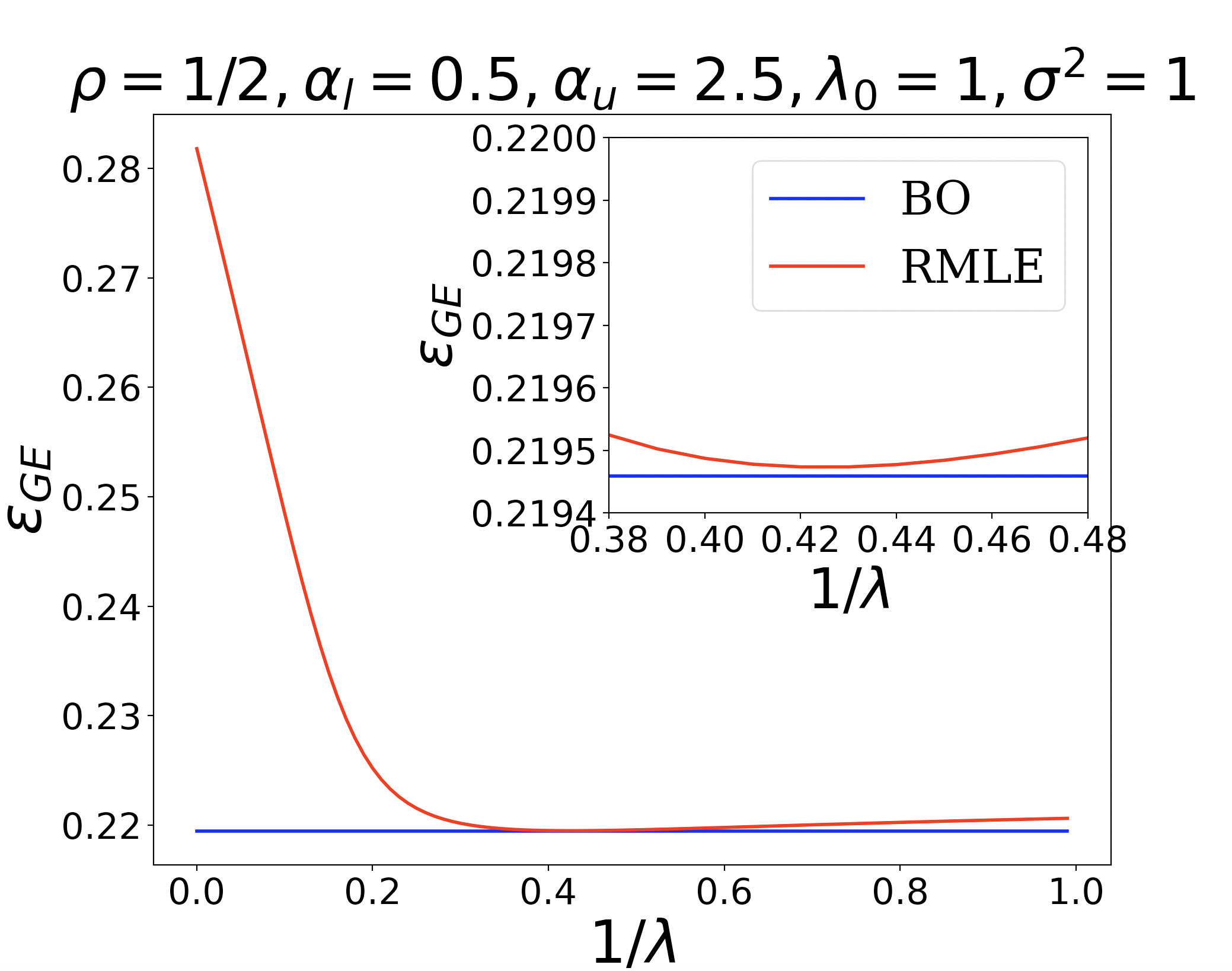}
\caption{\label{fig:rmle_bo_mmse_ge} Comparison of RMLE and the BO estimate in terms of MSE and GE. In the left panel, MSE is plotted against $1/\lambda$. The red curve represents the RMLE result for varying values of $1/\lambda$ from 0 to 1, showing how MSE changes w.r.t. this parameter. The blue straight line represents the BO result at $\lambda=\lambda_0$. Similarly, we plot GE against $1/\lambda$ in the right panel. The optimal values of $1/\lambda^*$ yielding the smallest gap from the BO result, are approximately $0.70$ and $0.42$ for MSE and GE, respectively.}.
\Lfig{MSE_ge}
\end{figure}

In \Rfig{MSE_ge}, we plot MSE (\Req{mse1}) and GE (\Req{ge3}) agianst $1/\lambda$. The left panel shows MSE while the right panel shows GE. The parameters are set as $\alpha_l=0.5$, $\alpha_u=2.5$, and SNR $=1$. The red curve represents the RMLE result while the blue straight line represents the BO result. The $\lambda$ value giving the minimum difference in the respective errors (MSE or GE) of RMLE and the BO estimate is called optimal and is denoted as $\lambda^*$. The values of $1/\lambda^*$ are around 0.70 and 0.42 for MSE and GE, respectively. We call RMLE at $\lambda=\lambda^*$ the optimal RMLE.

The comparison of the optimal RMLE and the BO estimate in terms of MSE and GE with a wide range of SNR is shown in \Rfig{Optimal lambda}. The upper panels show MSE while the bottom panels show GE. In the upper left panel, MSE is plotted against SNR. The red curve represents the optimal RMLE, while the blue curve represents the BO one. Throughout the entire range of SNR, the BO result outperforms the optimal RMLE. However, the ratio of the difference between the optimal RMLE and the BO estimate, which is denoted as $\Delta_{MSE}/\varepsilon_{BO, MSE}$ where $\Delta_{MSE} = \varepsilon_{RMLE, MSE} - \varepsilon_{BO, MSE}$, is small. The largest ratio of the difference is around 0.008 as shown in the inset window of the upper left panel. In the upper middle panel, we plot $\Delta_{MSE}/\varepsilon_{BO, MSE}$ against SNR for different values of $\alpha_u$. As $\alpha_u$ increases, $\Delta_{MSE}/\varepsilon_{BO, MSE}$ gradually decreases within an appropriately large SNR range. In the upper right panel, we examine different $\rho$ values. A larger imbalance in $\rho$ results in a smaller $\Delta_{MSE}/\varepsilon_{BO, MSE}$. The same analysis is conducted for GE in the bottom panels. In the bottom left panel, GE is plotted against SNR. The largest ratio of the difference between the optimal RMLE and the BO estimate, denoted as \textcolor{black}{$\Delta_{GE}/\varepsilon_{BO, GE}$} is around 0.00007. The bottom middle and right panels plot the $\textcolor{black}{\Delta_{GE}/\varepsilon_{BO, GE}}$ against the SNR with different values of $\alpha_u$ and $\rho$, respectively. These $\textcolor{black}{\Delta_{GE}/\varepsilon_{BO, GE}}$ curves exhibit behaviors similar to that of MSE, except for the case of $\rho=0.5$ in the bottom right panel.

\begin{figure}[H]
\centering
\includegraphics[width=0.32\textwidth]{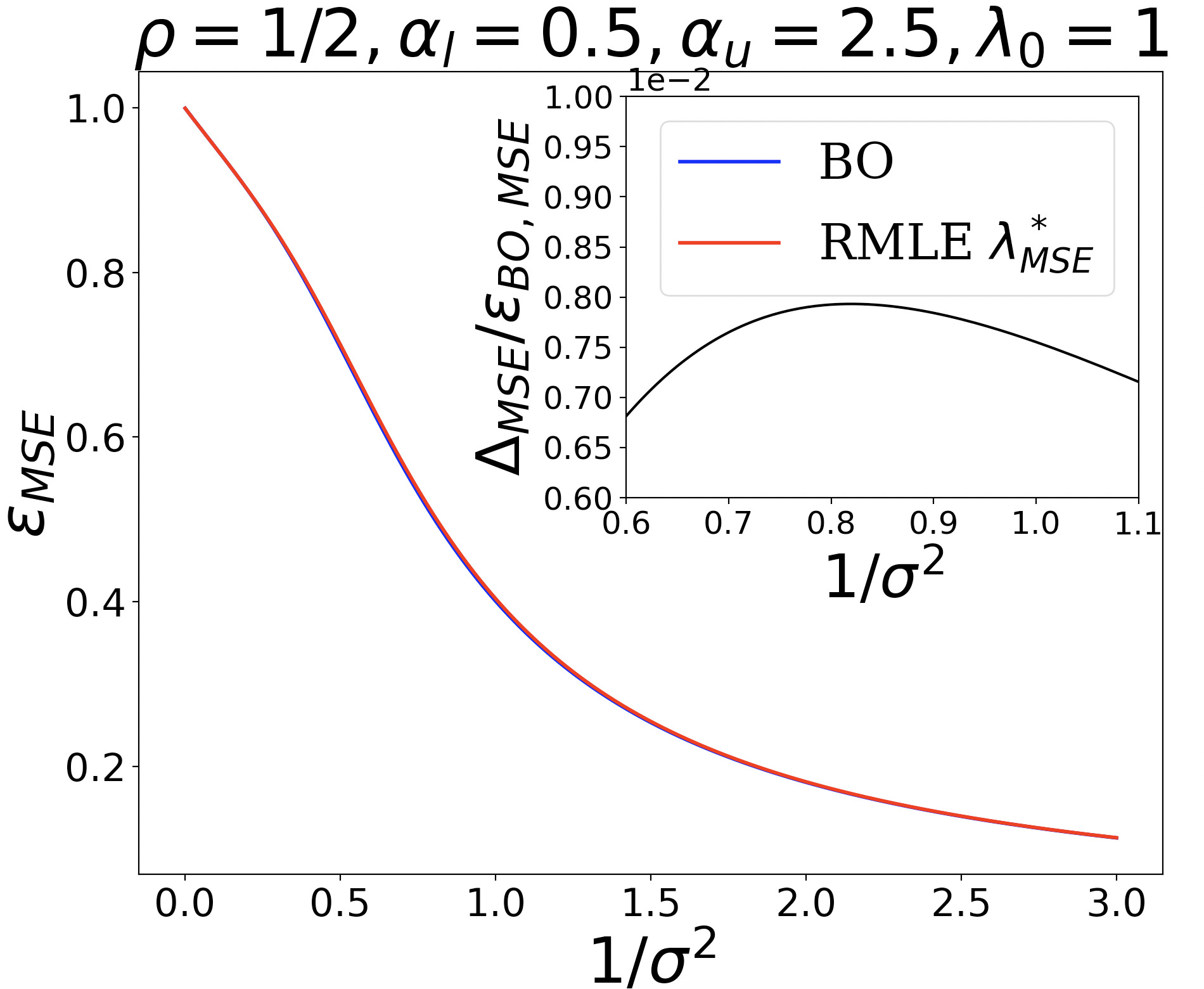}
\includegraphics[width=0.32\textwidth]{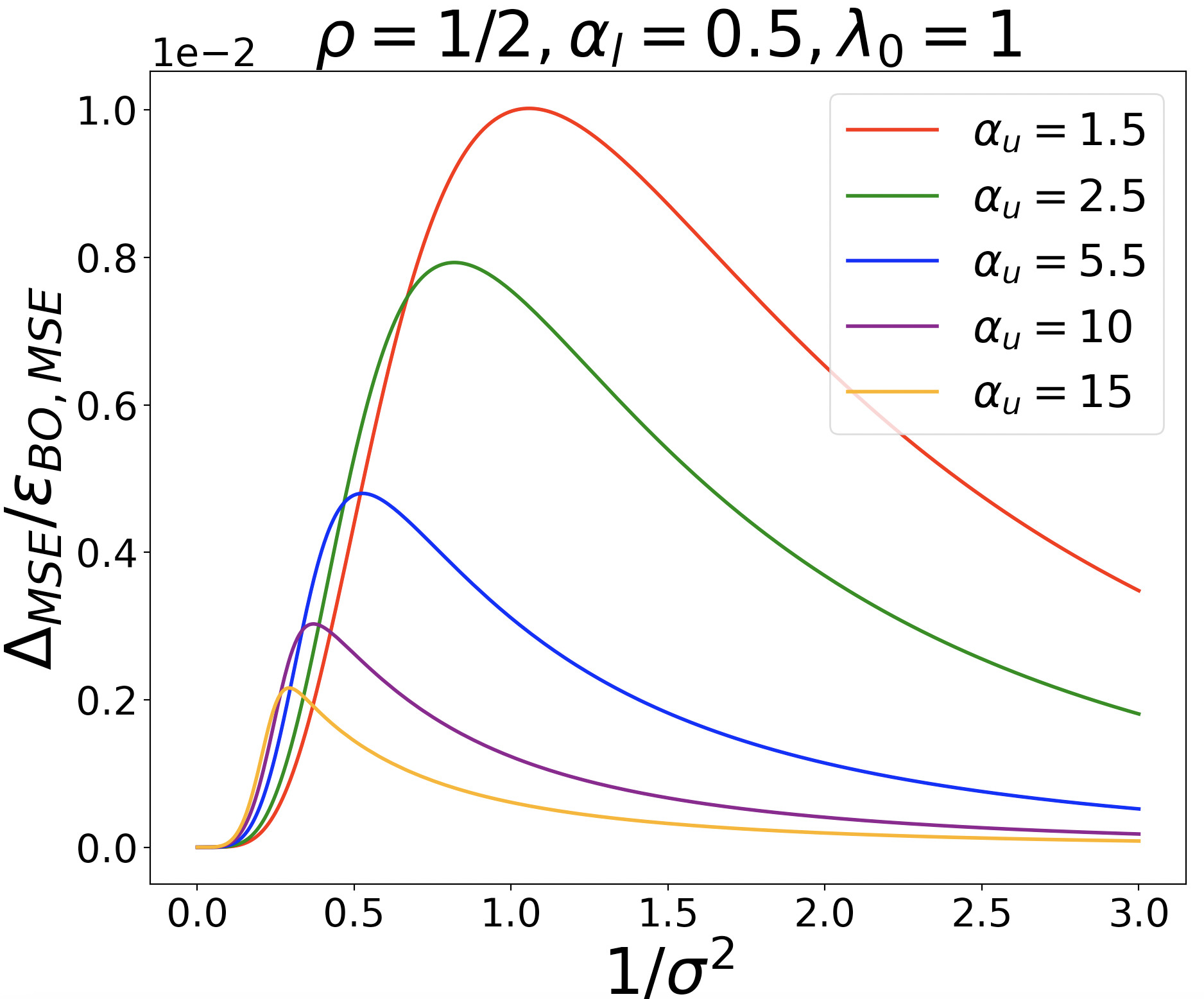}
\includegraphics[width=0.32\textwidth]{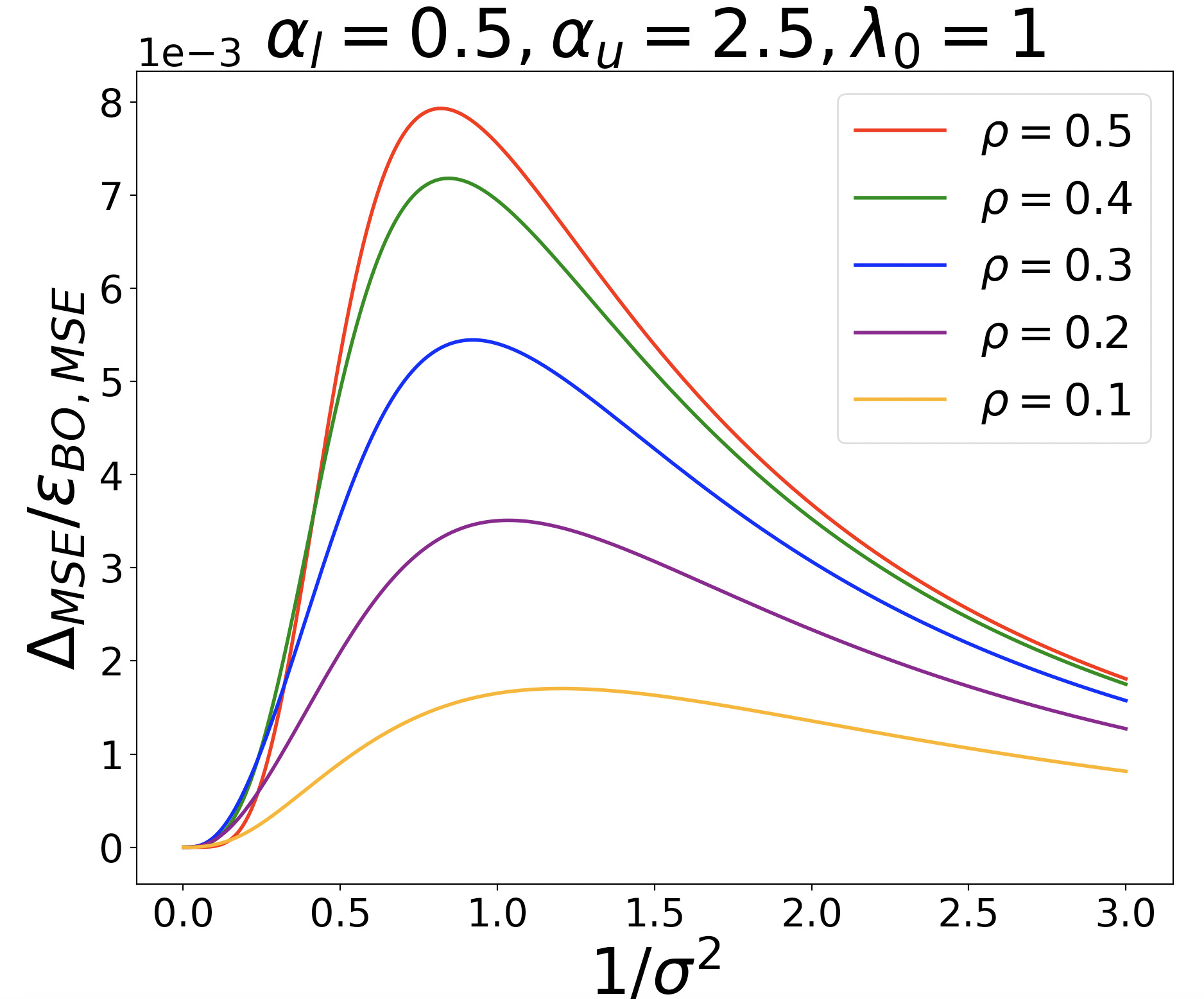}
\\
\includegraphics[width=0.32\textwidth]{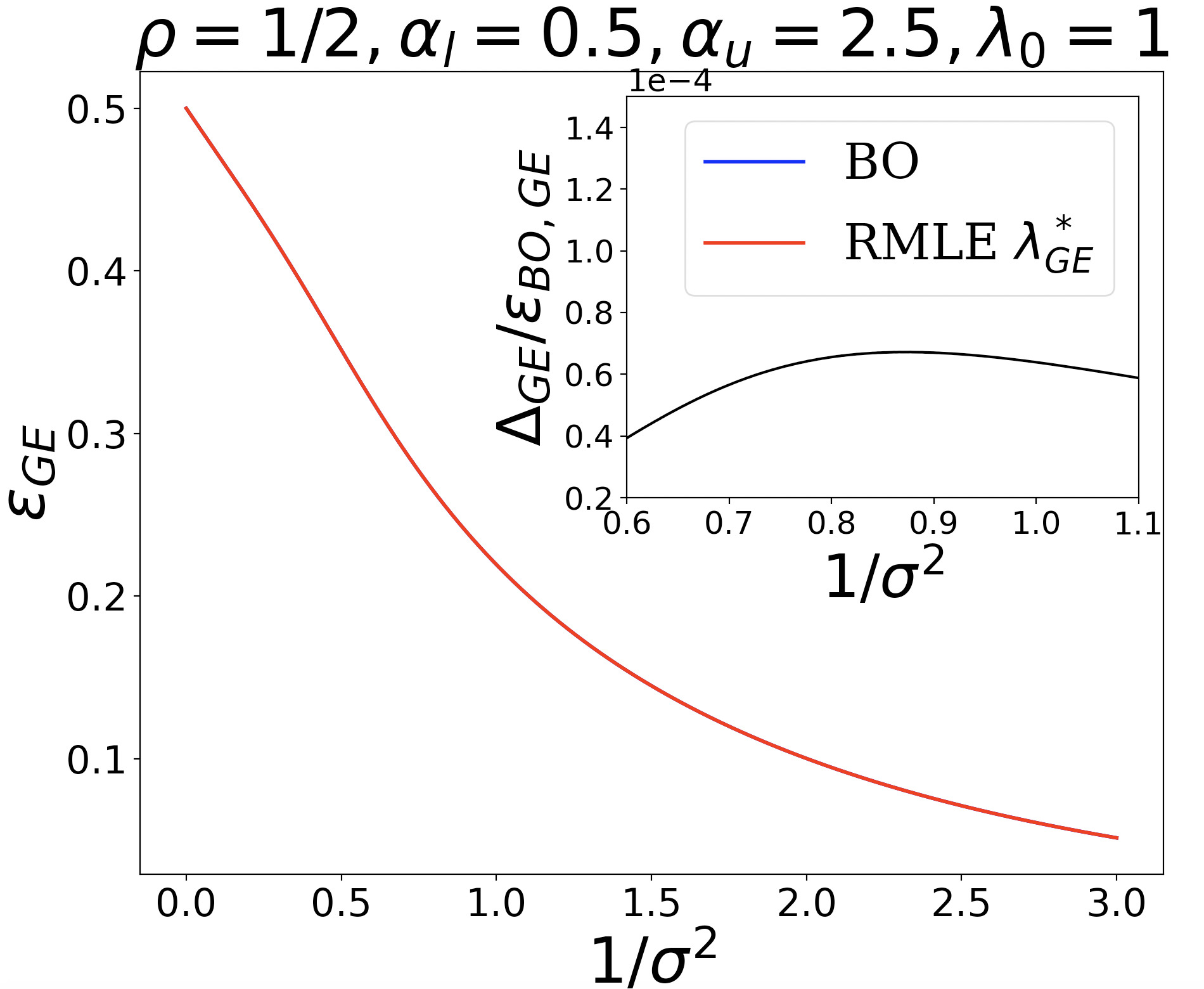}
\includegraphics[width=0.32\textwidth]{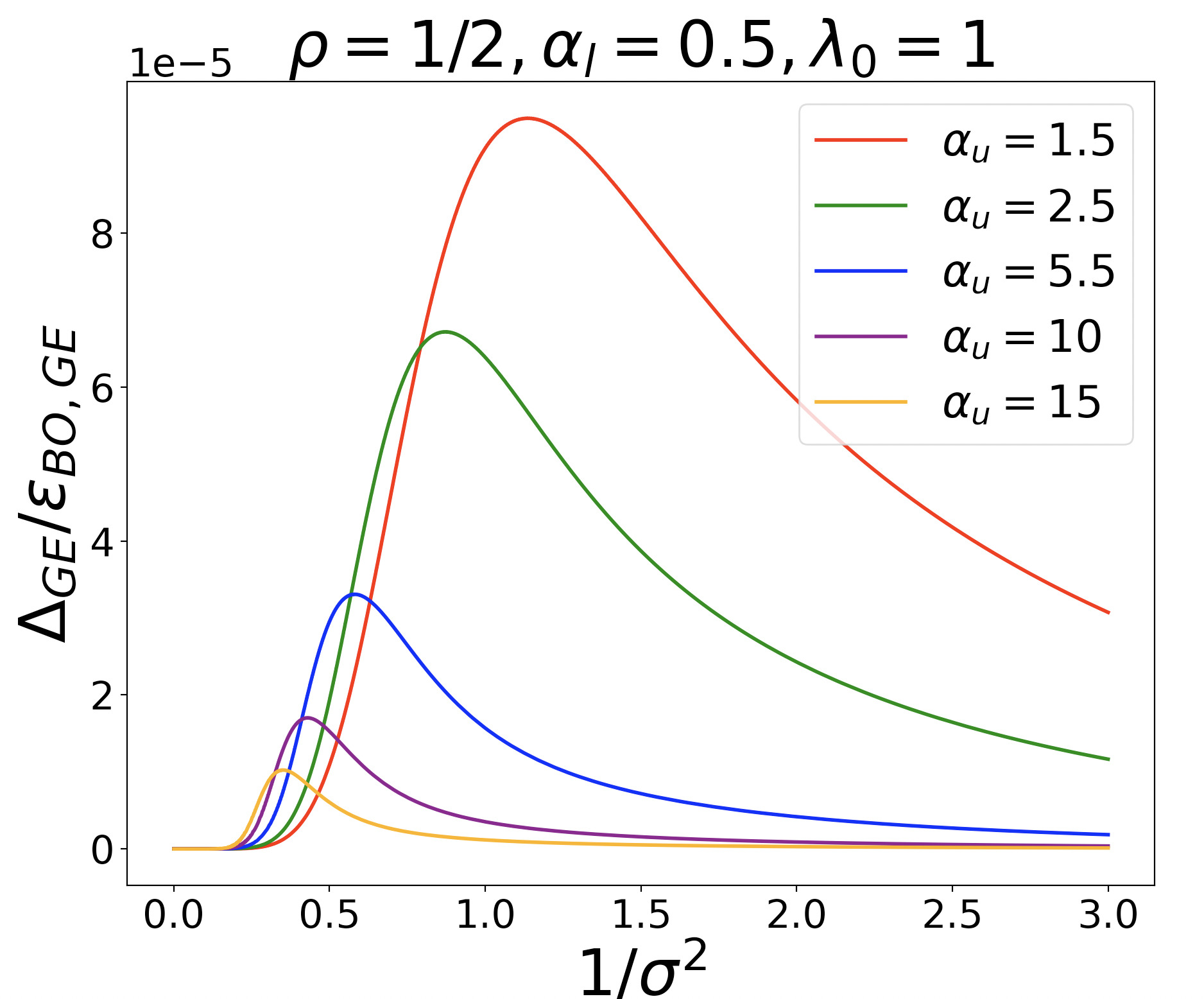}
\includegraphics[width=0.32\textwidth]{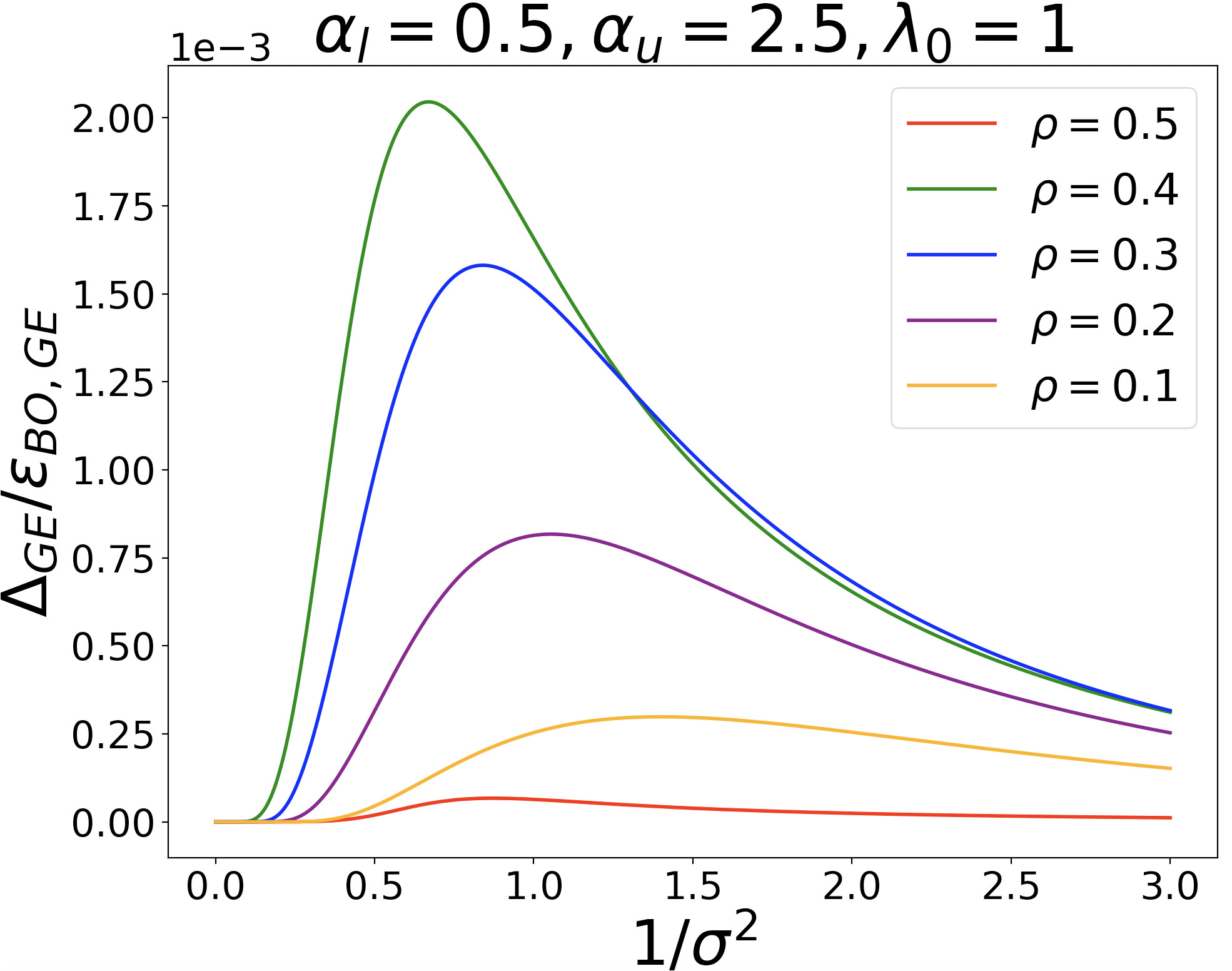}
\caption{\label{fig: optimization lambda} Comparison of the optimal RMLE and the BO estimate in terms of MSE and GE. The upper
panels show MSE while the bottom panels show GE. In the upper left panel, MSE is plotted against SNR. The red curve represents the optimal RMLE, while the blue curve represents the BO result. In the upper middle panel, $\Delta_{MSE}/\varepsilon_{BO, MSE}$ is plotted against SNR for different values of $\alpha_u$. Similarly, In the upper right panel, $\Delta_{MSE}/\varepsilon_{BO, MSE}$ is plotted against SNR for different values of $\rho$. The bottom panels are the counterpart for GE.}
\Lfig{Optimal lambda}
\end{figure}

Overall, whether using MSE or GE as the metric, the ratio of the difference between the optimal RMLE and the BO estimate is quite small, particularly with large $\alpha_u$. Thus, by optimally tuning $\lambda$, we can use RMLE as a good approximation of the BO estimate when handling substantial amounts of unlabeled data.

Next, we investigate how much the optimal $\lambda$ differs between GE and MSE. We show $1/\lambda^*$ plotted against SNR in \Rfig{Optimal lambda rho and alpha_u}. The top panels are for MSE, and the bottom ones are for GE. In the left panels, we show $1/\lambda^*$ with different values of $\alpha_u$. Both optimal values of $\lambda^*$ are finite when $\rho=0.5$, which is different from the supervised case where $\lambda^*$ diverges \cite{mignacco2020role}. Similarly, in the right panels, we show $1/\lambda^*$ with different values of $\rho$. For both MSE and GE, as $\rho$ deviates further from 0.5, the optimal $1/\lambda^*$ approaches 1. In the special case of $\rho=0$, our model transforms into an additive white Gaussian noise model and the MAP ($\beta=\infty, \lambda=\lambda_0$) estimation agrees with the BO result, which is consistent with findings in $\cite{rangan2011generalized}$.

The above analysis reveals that the ratio of the difference between the optimal RMLE and the BO estimate is very small, and the optimal $\lambda^*$ is finite. Although it is highly nontrivial to estimate the optimal $\lambda$ value for MSE, concerning GE we can employ some techniques to estimate GE such as cross-validation. This allows RMLE to be comparable with the BO estimate in terms of GE.

\begin{figure}[H]
\centering
\includegraphics[width=0.49\textwidth]{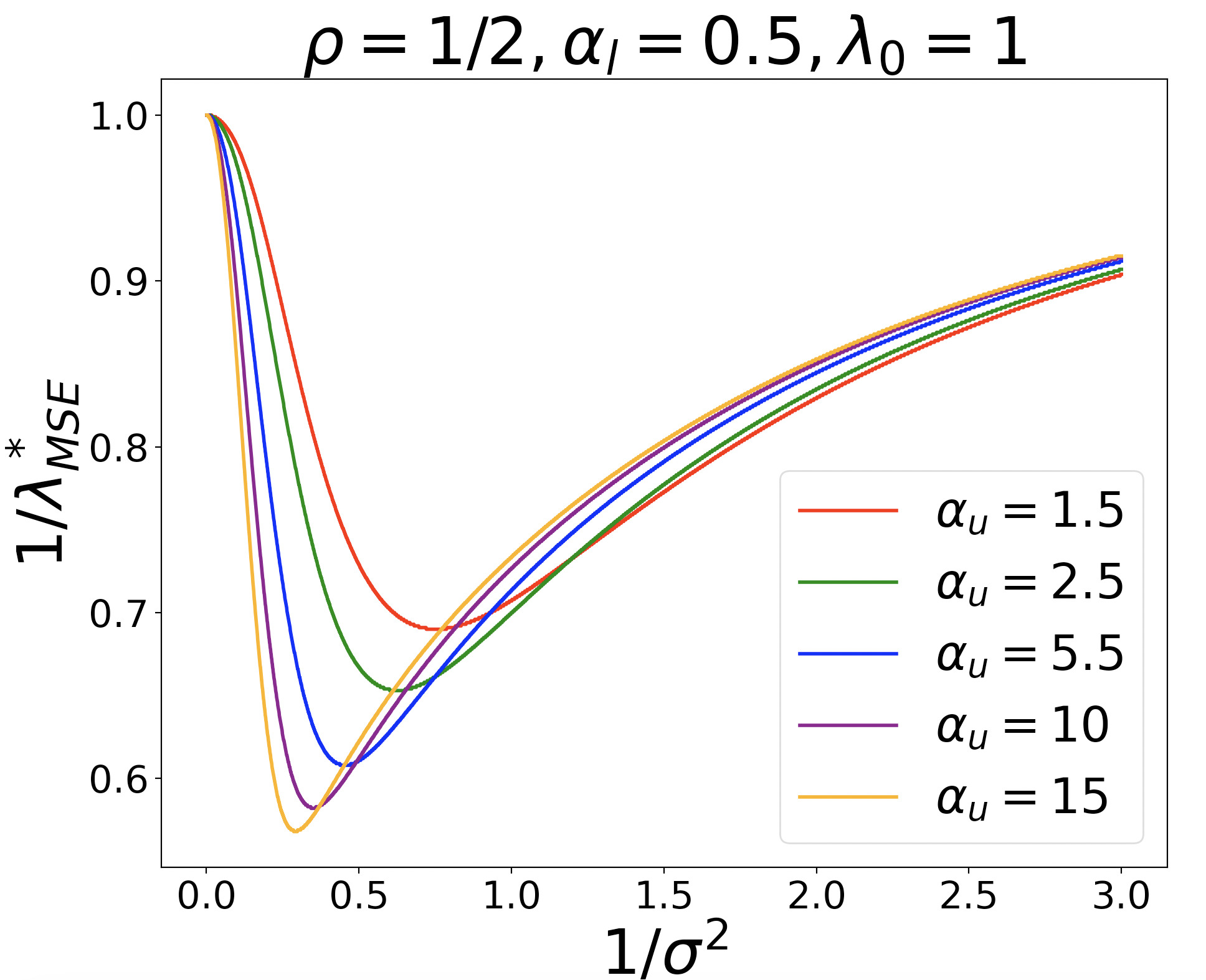}
\includegraphics[width=0.49\textwidth]{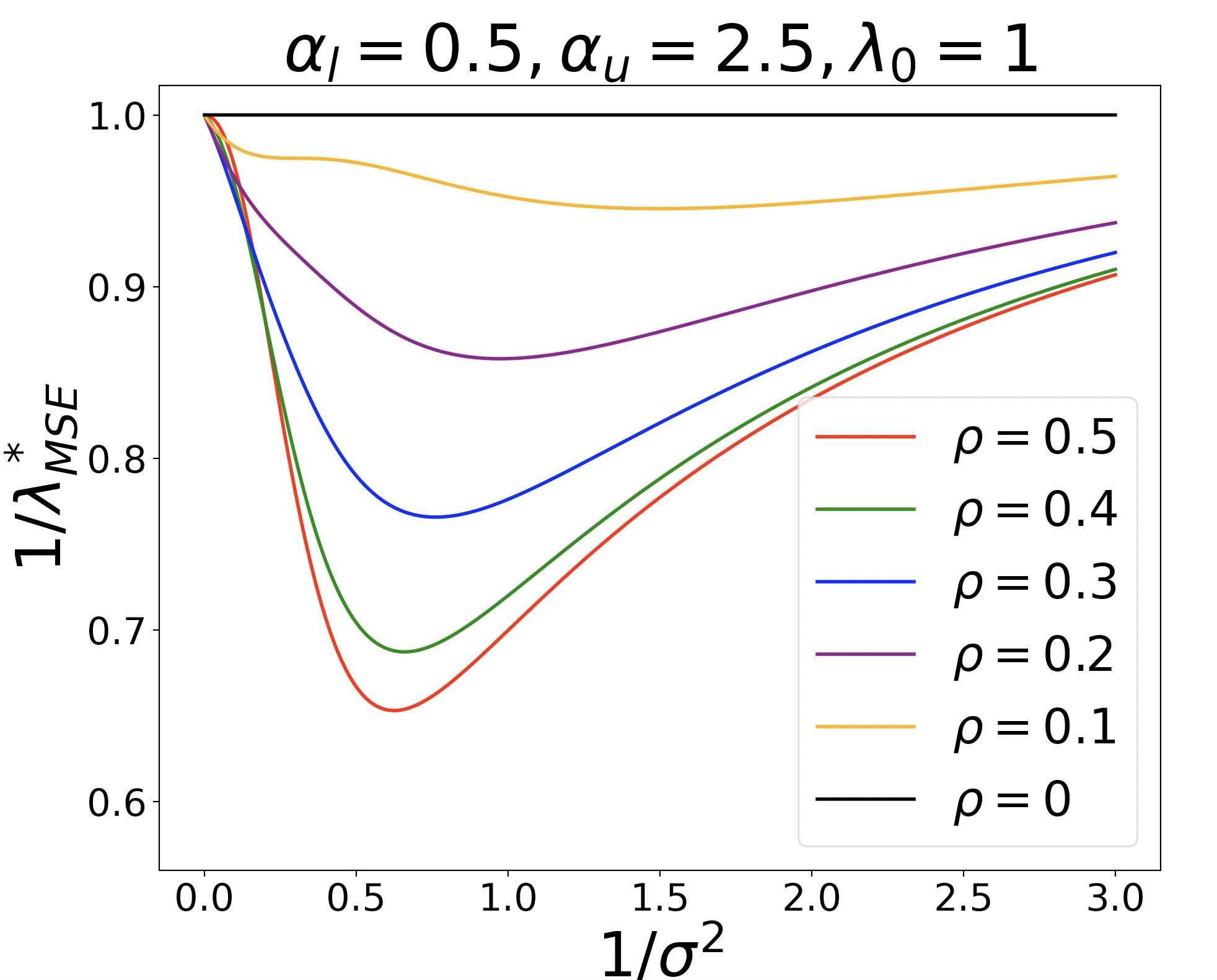}
\\
\includegraphics[width=0.49\textwidth]{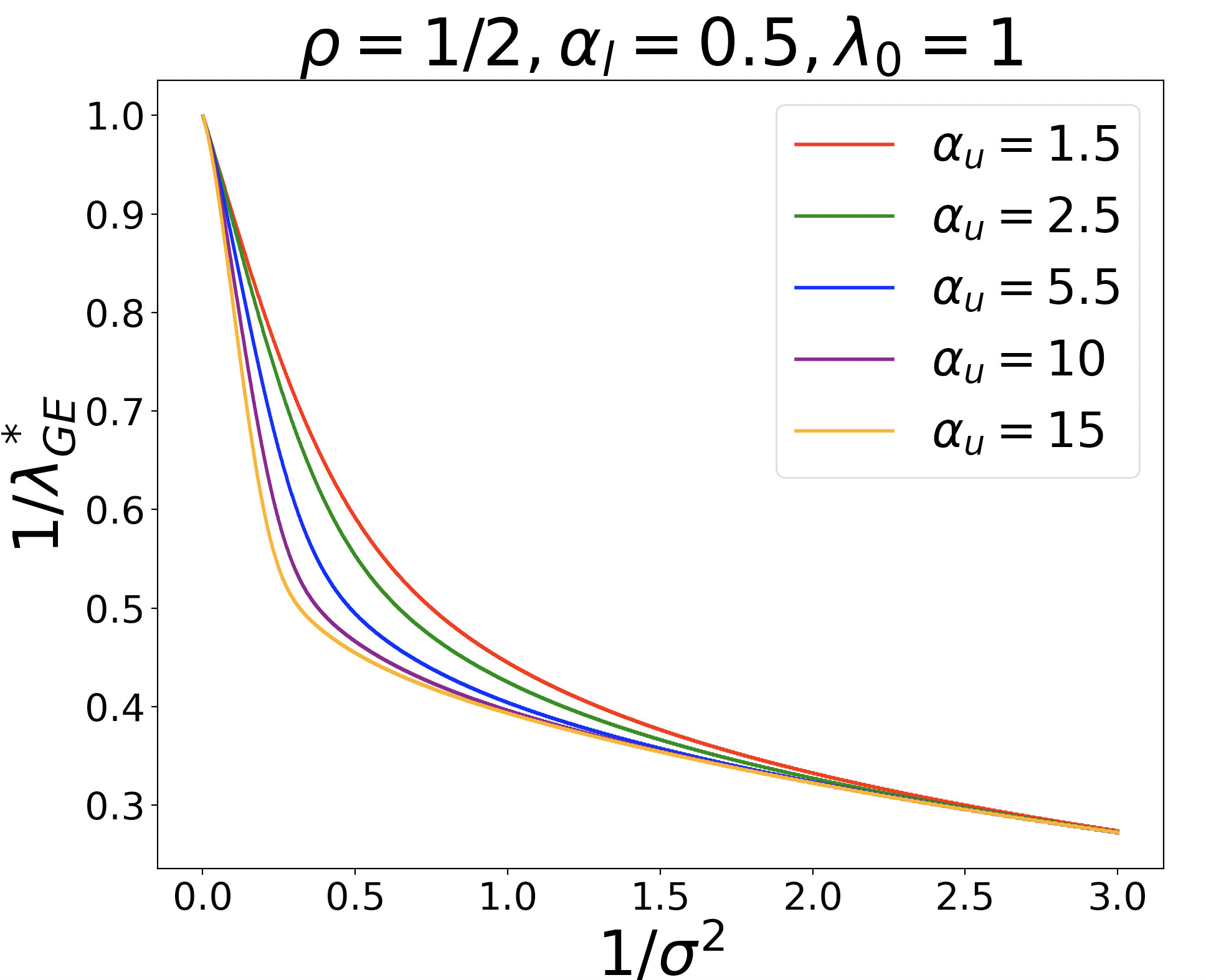}
\includegraphics[width=0.49\textwidth]{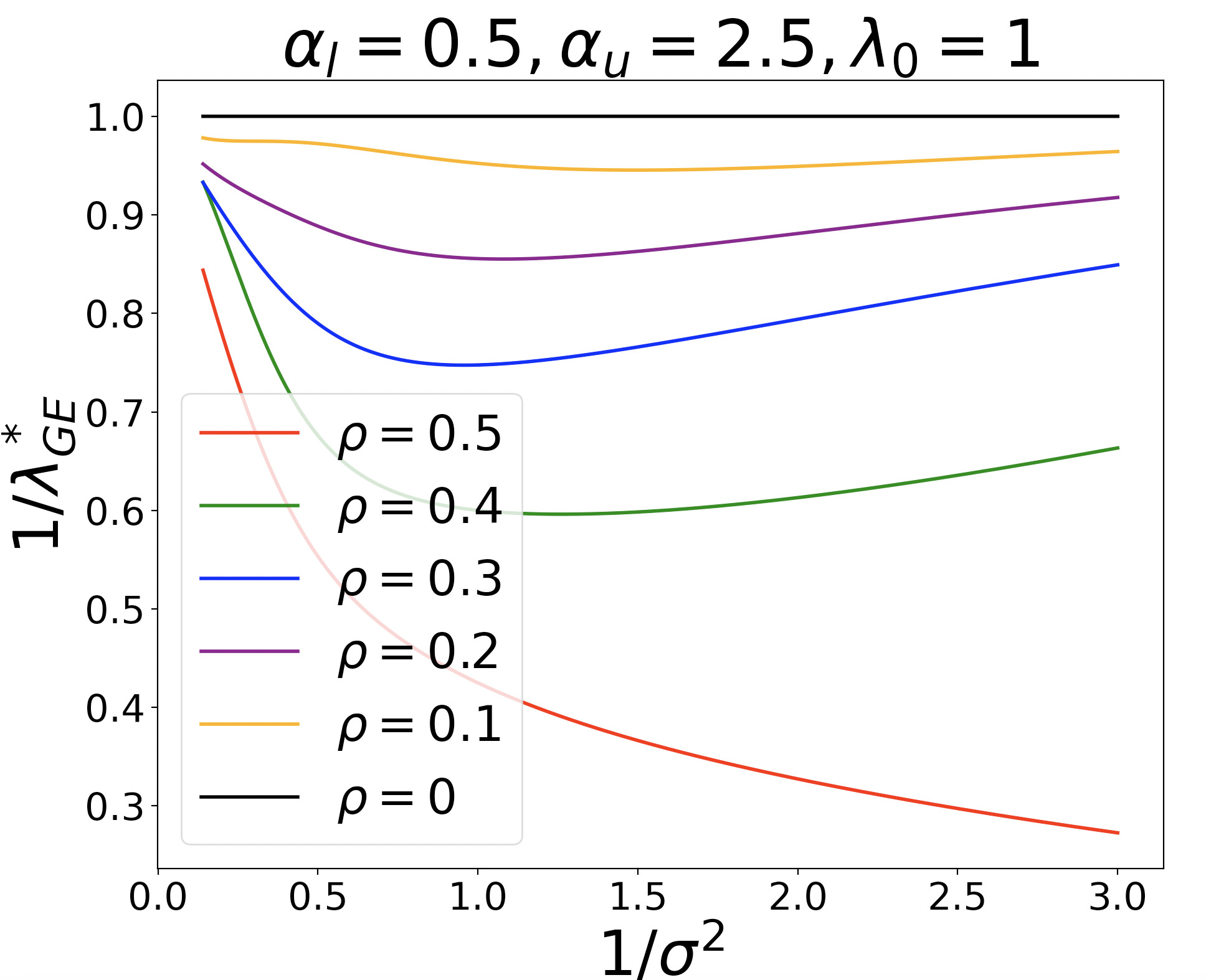}
\caption{\label{fig: optimization lambda rho and alpha_u} The optimal values of $\lambda^*$ under MSE and GE. The upper panels show the relation between $1/\lambda^*$ and SNR when using MSE, while the bottom panels show the counterpart for GE. The left panels plot $1/\lambda^*$ against SNR with various $\alpha_u$ values, while the right panels plot the same with different $\rho$ values.}
\Lfig{Optimal lambda rho and alpha_u}
\end{figure}

\section{Conclusion}
In our study, we investigated the Gaussian mixture problem as a model for high-dimensional labeled-unlabeled classification. We employed message-passing algorithm techniques and developed an efficient algorithm, namely AMP, for both RMLE and the Bayesian approach. This algorithm is an improved version of BP, and its macroscopic behavior was analyzed using the SE technique, showing rapid convergence of AMP. The SE analysis also revealed that there emerge sharp changes in order parameters at specific parameter values, signaling phase transitions. We constructed phase diagrams across a broad parameter range and explored the RSB effect related to multiple modes in the posterior distribution.
\par Our key findings are as follows: we derived phase diagrams for both RMLE ($\beta = \infty$) and Bayesian approach ($\beta = 1$), highlighting a critical phase transition from undetected to detected phases at $\rho=1/2$ and $\alpha_l=0$. The phase diagram is notably influenced by $\rho$, $\alpha_l$, and SNR. We also examined the RSB phenomenon in both RMLE and the Bayesian approach, identifying the RSB region in the parameter space. 
\par In our comparison of RMLE and the BO estimate, RMLE was shown to perform comparably with the BO one when the regularization is optimally tuned and unlabeled data is abundant. This suggests that RMLE can effectively leverage the additional information from unlabeled data to achieve performance on par with or similar to the BO in scenarios where labeled data is limited.
\par However, it is important to note that our study focused solely on binary classification. Practical applications such as text categorization often involve multi-class problems. Furthermore, the AMP algorithm may face challenges in the RSB phase due to instability. Therefore, a more sophisticated message-passing algorithm is desired. Our future work aims to extend our framework to address such cases.

\section{Acknowledgements}
The authors would like to thank Toshiyuki Tanaka for his insightful discussions and comments. This work was partially supported by JST SPRING, Grant Number JPMJSP2110 (XG), as well as JSPS KAKENHI Nos. 22K12179, and 22H05117 (TO). 

\appendix
\section{Appendix}

\subsection{RMLE via BP}\Lapp{subsub_RMLE}
The $\beta$-posterior reduces to RMLE when $\beta=\infty$. In this limit, the saddle-point method can be applied to compute $\hat{\bm{w}}$. Consequently, we obtain the messages from the function node to the variable node $\big\{\tilde{\phi}_{\mu\to i}^l(w_i), \tilde{\phi}_{\nu\to i}^u(w_i)\big\}$ as follows:
\begin{align}
	\tilde{\phi}_{\mu\to i}^l(w_i)&\propto \exp\Bigg\{\beta\bigg(\frac{S_\mu^l}{2} + y_{\mu}\Delta_{\mu i} + y_\mu \tilde{p}_{\mu\to i}^l -\frac{w_i^2}{2\sigma^2 N}  \bigg) \Bigg\},\Leq{Aeq70}\\
	\tilde{\phi}_{\nu\to i}^u(w_i)&\propto \exp\Bigg\{ \beta\bigg(G\big(\tilde{p}_{\nu\to i}^u,S_{\nu}^u\big) + F\big(\tilde{p}_{\nu\to i}^u,S_{\nu}^u\big)\Delta_{\nu i} +\frac{1}{2}T\big(\tilde{p}_{\nu\to i}^u,S_{\nu}^u\big)\Delta^2_{\nu i}-\frac{w_i^2}{2\sigma^2 N}  \bigg)\Bigg\},\Leq{Aeq71}
\end{align}
and the messages from the variable node to function node $\big\{\phi_{i\to\mu}^l(w_i), \phi_{i\to\nu}^u(w_i)\big\}$ as
\begin{align}
	\phi_{i\to\mu}^l(w_i)\propto \exp\Bigg\{\beta\bigg(-\frac{1}{2}\Big(\lambda+\frac{\alpha}{\sigma^2} &-\sum_{\nu = 1}^{M_u}  T\big(\tilde{p}_{\nu\to i}^u,S_\nu^u\big)\frac{\Delta^2_{\nu i}}{w_i^2}\Big)w_i^2 \nonumber \\
	 &+ \sum_{\omega(\neq \mu)}^{M_l}y_\omega \Delta_{\omega i} + \sum_{\nu = 1}^{M_u} F\big(\tilde{p}_{\nu\to i}^u,S_\nu^u\big)\Big)\Delta_{\nu i}\bigg)\Bigg\},\Leq{Aeq72}\\
	\phi_{i\to\nu}^u(w_i)\propto \exp\Bigg\{\beta\bigg(-\frac{1}{2}\Big(\lambda+\frac{\alpha}{\sigma^2} &- \sum_{m(\neq \nu)}^{M_u} T\big(\tilde{p}_{m\to i}^u,S_m^u\big)\frac{\Delta^2_{m i}}{w_i^2}\Big)w_i^2 \nonumber \\
	&+ \sum_{\omega=1}^{M_l}y_\omega \Delta_{\omega i} + \sum_{m(\neq\nu)}^{M_u}F\big(\tilde{p}_{m\to i}^u,S_m^u\big)\Big)\Delta_{m i}\bigg)\Bigg\},\Leq{Aeq73}
\end{align}
where $\Delta_{\mu i}=\frac{1}{\sigma^2 \sqrt{N}}x_{\mu i}w_i$. For the sake of simplicity, we abbreviate $G(y^*|p,t)$ as $G(p,t)$ in \Req{Aeq71}, where $y^*(p,t)$ represents the saddle-point of $G(y|p,t)$. The explicit formula for $y^*(p,t)$ is given by
\be
    y^*(p,t) \equiv\operatorname*{arg\,max}_{y}G(y|p,t),\Leq{Aeq74}
\ee
where
\be
    G(y|p,t) &=& -\frac{y^2}{2} + g(y|p,t),\Leq{Aeq75}\\
    g(y|p,t) &=& \ln\Big(\rho\exp\big\{p+\sqrt{t}y\big\}+(1-\rho)\exp\big\{-\big(p+\sqrt{t}y\big)\big\}\Big).\Leq{Aeq76}
\ee
This $y^*$ satisfies the following equation:
\be
        y^*(p,t) = \frac{\partial g(y^*|p,t)}{\partial y^*}=\frac{\sqrt{t}\Big(\rho \exp\big\{p+\sqrt{t}y^*\big\}-(1-\rho)\exp\big\{-\big(p+\sqrt{t}y^*\big) \big\}\Big)}{\rho \exp\big\{p+\sqrt{t}y^*\big\}+(1-\rho)\exp\big\{-\big(p+\sqrt{t}y^*\big) \big\}}.\Leq{Aeq77}
\ee

Furthermore, we abbreviate $F(y^*|p,t)$ and $T(y^*|p,t)$ as $F(p,t)$ and $T(p,t)$ in \Reqss{Aeq71}{Aeq73}, respectively. These quantities are defined as follows:
\be
    F(y^*|p,t) &=& \frac{\partial G(y^*|p,t)}{\partial p} =
    \frac{\rho \exp\Big\{p+\sqrt{t}y^*\Big\}-(1-\rho)\exp\Big\{-\big(p+\sqrt{t}y^*\big) \Big\}}{\rho \exp\Big\{p+\sqrt{t}y^*\Big\}+(1-\rho)\exp\Big\{-\big(p+\sqrt{t}y^*\big) \Big\}},\Leq{Aeq78}\\
    T(y^*|p,t) &=& \frac{\partial^2 G(y^*|p,t)}{\partial p^2} = \frac{1-F^2(y^*|p,t)}{1-t\big(1-F^2(y^*|p,t)\big)}.\Leq{Aeq79}
\ee
The messages $\big\{\phi_{i\to\mu}^l(w_i), \phi_{i\to\nu}^u(w_i)\big\}$ represent the marginal distributions of the variable node to the function node for the labeled and unlabeled cases, respectively, in the limit of convergence of the BP iteration. In the case of large systems, the discrepancy between \Req{Aeq72} and \Req{Aeq73} becomes negligible, allowing us to omit the superscripts $l$ and $u$ in the subsequent discussion. Furthermore, based on the observation, these messages follow a simple Gaussian form:
\be
	\phi_{i\to\nu}(w_i) \approx \exp\bigg\{-\frac{1}{2\chi_{i\to\nu}/\beta}\big(w_i - \hat{w}_{i\to\nu}\big)^2\bigg\},\Leq{Aeq80}
\ee
where
\be
    \chi_{i\to \nu} &=& \Bigg(\lambda+\frac{\alpha}{\sigma^2} - \frac{1}{\sigma^4 N}\sum_{m(\neq\nu)}^{M_u} x_{m i}^2  T\big(\tilde{p}_{m\to i},S_m\big)\Bigg)^{-1},\Leq{Aeq81}\\
    \hat{w}_{i\to\nu} &=& \frac{\chi_{i\to \nu}}{\sigma^2 \sqrt{N}}\Bigg(\sum_{\mu=1}^{M_l}y_\mu x_{\mu i} + \sum_{m(\neq\nu)}^{M_u}x_{m i}F\big(\tilde{p}_{m\to i},S_m\big)\Bigg).\Leq{Aeq82}
\ee
In \Req{Aeq80}, the message $\phi_{i\to\nu}(w_i)$ can be approximated as a Gaussian distribution for $w_i$, with mean $\hat{w}_{i\to\nu}$ (\Req{Aeq82}) and variance $\chi_{i\to\nu}/\beta$ (\Req{Aeq81} scaled by $1/\beta$). As $\beta$ approaches infinity, the variance $\chi_{i\to\nu}/\beta$ tends to zero, resulting in a point estimator for $w_i$. Finally, we derive the BP iteration equations for RMLE as follows:
\subbe
\be
    \tilde{p}_{\nu\to i}^{(t)} &=& \frac{1}{\sigma^2\sqrt{N}}\sum_{j(\neq i)}^N x_{\nu j}\hat{w}_{j\to\nu}^{(t)}, \Leq{Abp_rmle1}\\
    S_\nu^{(t)} &=& \frac{1}{\sigma^4 N}\sum_{i}^N x_{\nu i}^2 \chi_{i\to \nu}^{(t)},\Leq{Abp_rmle2}\\
    \chi_{i\to \nu}^{(t+1)} &=& \Bigg(\lambda+\frac{\alpha}{\sigma^2} - \frac{1}{\sigma^4 N}\sum_{m(\neq\nu)}^{M_u} x_{m i}^2  T\Big(\tilde{p}_{m\to i}^{(t)},S_m^{(t)}\Big)\Bigg)^{-1},\Leq{Abp_rmle3}\\
    \hat{w}_{i\to\nu}^{(t+1)} &=& \frac{\chi_{i\to \nu}^{(t+1)}}{\sigma^2 \sqrt{N}}\Bigg(\sum_{\mu=1}^{M_l}y_\mu x_{\mu i} + \sum_{m(\neq\nu)}^{M_u}x_{m i}F\Big(\tilde{p}_{m\to i}^{(t)},S_m^{(t)}\Big)\Bigg).\Leq{Abp_rmle4}
\ee
\Leq{A_BP_RMLE}
\subee

\subsection{The Bayesian approach via BP}\Lapp{subsub_BO}
The $\beta$-posterior reduces to the Bayesian approach when $\beta=1$. Consequently, we obtain the messages from the function node to the variable node $\big\{\tilde{\phi}_{\mu\to i, B}^l(w_i), \tilde{\phi}_{\nu\to i, B}^u(w_i)\big\}$ as follows:
\be
    \tilde{\phi}_{\mu\to i,B}^l(w_i)&=& \exp\Bigg\{\bigg(\frac{S_{\mu,B}^l}{2} + y_\mu \Delta_{\mu i}+ y_\mu \tilde{p}_{\mu\to i,B}^l -\frac{w_i^2}{2\sigma^2 N}  \bigg) \Bigg\},\Leq{Aeq84}\\
      \tilde{\phi}_{\nu\to i,B}^u(w_i) &\propto& \exp\bigg\{ \tilde{F}\big(\tilde{p}_{\nu\to i,B}^u\big)\Delta_{\nu i}+ \frac{1}{2}\tilde{T}\big(\tilde{p}_{\nu\to i,B}^u\big)\Delta^2_{\nu i} - \frac{w_i^2}{2\sigma^2 N}  \bigg\}, \Leq{Aeq85}
\ee
and the messages from the variable node to function node $\big\{\phi_{i\to\mu,B}^l(w_i), \phi_{i\to\nu,B}^u(w_i)\big\}$ as
\begin{align}
    \phi_{i\to\mu,B}^l(w_i) \propto \exp\Bigg\{\bigg(-\frac{1}{2}\Big(\lambda+\frac{\alpha}{\sigma^2} &- \sum_{\nu=1}^{M_u}\tilde{T}\big(\tilde{p}_{\nu\to i,B}^u\big)\frac{\Delta^2_{\nu i}}{w_i^2}\Big)w_i^2 \nonumber \\ 
    &+ \sum_{\omega(\neq \mu)}^{M_l}y_\omega\Delta_{\omega i} + \sum_{\nu = 1}^{M_u}\tilde{F}\big(\tilde{p}_{\nu\to i,B}^u\big)\Delta_{\nu i}\bigg)\Bigg\}, \Leq{Aeq86}\\
    \phi_{i\to\nu,B}^u(w_i) \propto \exp\Bigg\{\bigg(-\frac{1}{2}\Big(\lambda+\frac{\alpha}{\sigma^2} &- \sum_{m(\neq\nu)}^{M_u}\tilde{T}\big(\tilde{p}_{m\to i,B}^u\big)\frac{\Delta^2_{m i}}{w_i^2}\Big)w_i^2 \nonumber \\ 
    &+ \sum_{\mu=1}^{M_l} y_\mu\Delta_{\mu i} + \sum_{m(\neq\nu)}^{M_u}\tilde{F}\big(\tilde{p}_{m\to i,B}^u\big)\Delta_{m i}\bigg)\Bigg\},\Leq{Aeq87}
\end{align}
where
\be
	\tilde{F}(p)&=&\frac{\rho \exp\{p\}-(1-\rho)\exp\{-p\}}{\rho\exp\{p\}+(1-\rho)\exp\{-p\}},\Leq{Aeq88}\\
	\tilde{T}(p)&=&\frac{\partial \tilde{F}}{\partial p}(p) = \frac{4\rho(1-\rho)}{\big(\rho\exp\{p\}+(1-\rho)\exp\{-p\}\big)^2}.\Leq{Aeq89}
\ee
Similar to the reason in \Rapp{subsub_RMLE}, we omit the superscripts $l$ and $u$, and the messages can be approximated as Gaussian distributions as
\be
	\phi_{i\to\nu,B}(w_i)\approx \exp\bigg\{-\frac{1}{2\chi_{i\to\nu,B}}\big(w_i - \hat{w}_{i\to\nu,B}\big)^2\bigg\},\Leq{Aeq90}
\ee
where
\be
    \chi_{i\to \nu,B} &=& \Bigg(\lambda+\frac{\alpha}{\sigma^2} - \frac{1}{\sigma^4 N}\sum_{m(\neq\nu)}^{M_u} x_{m i}^2  \tilde{T}\big(\tilde{p}_{m\to i,B}\big)\Bigg)^{-1},\Leq{Aeq91}\\
    \hat{w}_{i\to\nu,B} &=& \frac{\chi_{i\to \nu,B}}{\sigma^2 \sqrt{N}}\Bigg(\sum_{\mu=1}^{M_l}y_\mu x_{\mu i} + \sum_{m(\neq\nu)}^{M_u}x_{m i}\tilde{F}\big(\tilde{p}_{m\to i,B}\big)\Bigg).\Leq{Aeq92}
\ee
The distinction between $\beta=\infty$ and $\beta=1$ lies in the output of the estimation. The latter yields the posterior mean through the message-passing algorithm by \Req{Aeq90}, while the former provides a point estimator by \Req{Aeq80}. Finally, we derive the BP iteration equations for the Bayesian approach as follows:
\subbe
\be
\tilde{p}_{\nu\to i,B}^{(t)} &=& \frac{1}{\sigma^2\sqrt{N}}\sum_{j(\neq i)}^N x_{\nu j}\hat{w}_{j\to\nu,B}^{(t)},\Leq{Abp_bo1}\\
\chi^{(t+1)}_{i\to\nu,B} &=& \Bigg(\lambda+\frac{\alpha}{\sigma^2}-\frac{1}{\sigma^4 N}\sum_{m\neq\nu}^{M_\mu}x_{m i}^2\tilde{T}\Big(\tilde{p}_{m\to i,B}^{(t)}\Big)\Bigg)^{-1},\Leq{Abp_bo2}\\
\hat{w}_{i\to\nu,B}^{(t+1)} &=& \frac{\chi^{(t+1)}_{i\to\nu,B}}{\sigma^2 \sqrt{N}}\Bigg(\sum_{\mu=1}^{M_l}y_\mu x_{\mu i} + \sum_{m(\neq\nu)}^{M_u}x_{m i}\tilde{F}\Big(\tilde{p}_{m\to i,B}^{(t)}\Big)\Bigg).\Leq{Abp_bo3}
\ee
\Leq{A_BP_BO}
\subee

\subsection{AMP for RMLE and the Bayesian approach}\Lapp{AMP} 
The key idea underlying AMP, which drew inspiration from the Thouless-Anderson-Palmer \cite{Thouless01031977} (TAP) mean-field approach, is to connect the “cavity" solution $\big\{ \hat{w}_{i\to\nu}^{(t+1)} \big\}_\nu$ to the full solution defined by
\be
    \hat{w}_i =\frac{\chi_{i \to \nu}}{\sigma^2 \sqrt{N}}\bigg(\sum_{\mu = 1}^{M_l}y_\mu x_{\mu i} + \sum_{m = 1}^{M_u}x_{m i}F\big(\tilde{p}_{m\to i},S_m\big)\bigg).\Leq{Aeq106}
\ee
For simplicity, we provide the AMP derivation for RMLE only. Further, we introduce the notation as
\be
    \tilde{p}_{\nu} = \frac{1}{\sigma^2 \sqrt{N}}\sum_j^N x_{\nu j}\hat{w}_{j\to\nu} = \tilde{p}_{\nu\to i} + \frac{1}{\sigma^2 \sqrt{N}}x_{\nu i}\hat{w}_{i\to\nu}.\Leq{Aeq107}
\ee
The difference between full and cavity solutions can be written as
\begin{align}
    \hat{w}_{i\to\nu} &= \hat{w}_i - \frac{\chi_{i\to\nu} }{\sigma^2\sqrt{N}}x_{\nu i}F(\tilde{p}_{\nu\to i},S_\nu)\no \\
    &\approx \hat{w}_i - \frac{\chi }{\sigma^2\sqrt{N}}x_{\nu i}F\big(\tilde{p}_\nu,S_\nu\big) + \mathcal{O}(N^{-1}).\Leq{Aeq108}
\end{align}
We omit the subscript of $\chi$ for the same reason as \Req{Aeq100}. Additionally, when $N$ is sufficiently large, terms of $\mathcal{O}(N^{-1})$ can be neglected. As a result, $\tilde{p}_\nu$ can be approximated as follows:
\be
    \tilde{p}_\nu
    \approx\frac{1}{\sigma^2\sqrt{N}}\sum_j^N x_{\nu j}\hat{w}_j - \frac{\chi}{\sigma^4 N}\sum_{j}^N x_{\nu j}^2 F\Big(\tilde{p}_\nu,\frac{\chi}{\sigma^2}\Big).\Leq{Aeq109}
\ee
In the same approximation, we have
\be
   \hat{w}_i
   \approx \frac{\chi}{\sigma^2 \sqrt{N}}\Bigg(\sum_{\mu = 1}^{M_l}y_\mu x_{\mu i} + \sum_{m = 1}^{M_u}x_{m i}\bigg(F\Big(\tilde{p}_{m},\frac{\chi}{\sigma^2}\Big) - \frac{\hat{w}_{i}}{\sigma^2\sqrt{N}}x_{m i} T\Big(\tilde{p}_m,\frac{\chi}{\sigma^2}\Big)\bigg)\Bigg).\Leq{Aeq110}
\ee
Therefore, we obtain the AMP iteration of RMLE as
\subbe
\begin{align}
    \tilde{p}_\nu^{(t)} &= \frac{1}{\sigma^2\sqrt{N}}\sum_j^N x_{\nu j}\hat{w}_j^{(t)} - \frac{\chi^{(t)}}{\sigma^4 N}\sum_{j}^N x_{\nu j}^2 F\Big(\tilde{p}_\nu^{(t-1)},\frac{\chi^{(t-1)}}{\sigma^2}\Big),\Leq{Aamp_rmle1}\\
    \chi^{(t+1)} &= \Bigg(\lambda+\frac{\alpha}{\sigma^2} - \frac{1}{\sigma^4 N}\sum_{\nu=1}^{M_u}  x_{\nu i}^2 T\Big(\tilde{p}_{\nu}^{(t)},\frac{\chi^{(t)}}{\sigma^2}\Big)\Bigg)^{-1},\Leq{Aamp_rmle2}\\
    \hat{w}_i^{(t+1)} &= \frac{\chi^{(t+1)}}{\sigma^2 \sqrt{N}}\Bigg(\sum_{\mu = 1}^{M_l}y_\mu x_{\mu i} + \sum_{m = 1}^{M_u}x_{m i} F\Big(\tilde{p}_{m}^{(t)},\frac{\chi^{(t)}}{\sigma^2}\Big) - \frac{\hat{w}_{i}^{(t)}}{\sigma^2\sqrt{N}}\sum_{m = 1}^{M_u} x_{m i}^2 T\Big(\tilde{p}_m^{(t)},\frac{\chi^{(t)}}{\sigma^2}\Big)\Bigg).\Leq{Aamp_rmle3}
\end{align}
\Leq{Aamp_rmle}
\subee
Similarly, with the same techniques, we derive the AMP iteration for the Bayesian approach as
\subbe
\begin{align}
    \tilde{p}_{\nu,B}^{(t)} &= \frac{1}{\sigma^2\sqrt{N}}\sum_j^N x_{\nu j}\hat{w}_{j,B}^{(t)} - \frac{\chi_{B}^{(t)}}{\sigma^4 N}\sum_j^N x_{\nu j}^2 \tilde{F}{\Big(\tilde{p}_{\nu,B}^{(t-1)}\Big)},\Leq{Aamp_bo1}\\
    \chi^{(t+1)}_{B} &= \Bigg(\lambda + \frac{\alpha}{\sigma^2} - \frac{1}{\sigma^4 N}\sum_{\nu=1}^{M_u}x_{\nu i}^2\tilde{T}\Big(\tilde{p}_{\nu,B}^{(t)}\Big)\Bigg)^{-1},\Leq{Aamp_bo2}\\
    \hat{w}_{i,B}^{(t+1)} &= \frac{\chi^{(t+1)}_{B}}{\sigma^2\sqrt{N}}\Bigg(\sum_{\mu=1}^{M_l}y_{\mu}x_{\mu i}+\sum_{m=1}^{M_u} x_{m i}\tilde{F}{\Big(\tilde{p}_{m,B}^{(t)}\Big)} - \frac{\hat{w}_i^{(t)}}{\sigma^2\sqrt{N}}\sum_{m=1}^{M_u}x_{m i}^2\tilde{T}\Big(\tilde{p}_{m,B}^{(t)}\Big)\Bigg).\Leq{Aamp_bo3}
\end{align}
\Leq{Aamp_bo}
\subee


\subsection{SE for RMLE and the Bayesian approach}\Lapp{state_evolution}
Let us first analyze the macroscopic behavior of BP iteration \Req{Abp_rmle4}. We assume the inferred solution $\big\{\hat{w}_{i\to\nu}^{(t)}\big\}_{i,\nu}$ obeys the following Gaussian distribution:
\be
    \hat{w}_{i\to\nu}^{(t)} \sim \mathcal{N}\Big(k_t w_{0i},v_t\Big).\Leq{Aeq94}
\ee
This assumption allows us to express $\hat{w}_{i\to\nu}^{(t)}$ as
\be
    \hat{w}_{i\to\nu}^{(t)} = k_t w_{0i} + \sqrt{v_t} z_{i\nu}, \ z_{i\nu}\sim \mathcal{N}\big(0,1\big).\Leq{Aeq95}
\ee
Then,
\begin{align}
    \tilde{p}_{\nu\to i}^{(t)} = \frac{1}{\sigma^2\sqrt{N}}\sum_{j(\neq i)}^N x_{\nu j}\hat{w}_{j\to\nu}  &= \frac{1}{\sigma^2\sqrt{N}}\sum_{j(\neq i)}^N \bigg(y_\nu\frac{w_{0j}}{\sqrt{N}} +\sigma \xi_{\nu j}\bigg)\Big(k_t w_{0j} + v_t z_{i\nu}\Big)\no \\
    &\approx y_\nu k_t v_s + \sqrt{\tilde{v}_t} z,\Leq{Aeq96}
\end{align}
where $\xi_{\nu j}\sim \mathcal{N}\big(0,1\big)$, $z\sim\mathcal{N}\big(0,1\big)$, and
\begin{align}
    v_s &= \frac{1}{N}\sum_{j(\neq i)}^N w_{0 j}^2,\Leq{Aeq97}\\
    \tilde{v}_t &= k_t^2 v_s + v_t.\Leq{Aeq98}
\end{align}
In large $N$ limit, $v_s$ converges to the true signal variance $1/\lambda_0$, and hence we assume $v_s = 1/\lambda_0$ holds throughout this paper. Thanks to the \textcolor{black}{law of large numbers}, the third part in \Req{Abp_rmle3} can be approximated as follows:
\be
    \frac{1}{\sigma^4 N}\sum_{m(\neq\nu)}^{M_u} x_{m i}^2  T\big(\tilde{p}_{m\to i}^{(t)},S_m^{(t)}\big)
    \approx \frac{\alpha_u}{\sigma^2} \int Dz\  \bigg(\rho T\Big(\mathcal{P},S_m^{(t)}\Big)+(1-\rho) T\Big(\mathcal{Q},S_m^{(t)}\Big) \bigg),\Leq{Aeq99}
\ee
where $Dz = e^{-z^2/2}/\sqrt{2\pi}\ dz$, $\alpha_u = M_u/N$ ($\alpha=\alpha_l+\alpha_u$), $\mathcal{P} = k_t/(\lambda_0\sigma^2) + \sqrt{\tilde{v}_t/\sigma^2} z$, and $\mathcal{Q} = -k_t/(\lambda_0\sigma^2) + \sqrt{\tilde{v}_t/\sigma^2} z$. Hence, \Req{Abp_rmle2}, \Req{Abp_rmle3}, and \Req{Aeq99} imply that the subscripts of $\chi$ and $S$ are negligible as
\begin{align}
    \chi_{i\to \nu}^{(t+1)}&\approx \chi_{t+1}\no \\
    &= \Bigg(\lambda+\frac{\alpha}{\sigma^2} - \frac{\alpha_u}{\sigma^2} \int Dz\ \bigg(\rho T\Big(\mathcal{P}, S^{(t)}_m\Big)+(1-\rho) T\Big(\mathcal{Q}, S^{(t)}_m\Big) \bigg)\Bigg)^{-1},\Leq{Aeq100}\\
    S_m^{(t)}&\approx S^{(t)} = \frac{\chi_{t}}{\sigma^4 N}\sum_i^N x_{mi}^2 = \frac{\chi_{t}}{\sigma^2}.\Leq{Aeq101}
\end{align}
Similarly, we evaluate the mean and variance of $\frac{1}{\sigma^2 \sqrt{N}}\sum_{m(\neq\nu)}^{M_u}x_{m i}F\Big(\tilde{p}_{m\to i}^{(t)},S_m^{(t)}\Big)$ in \Req{Abp_rmle4} as
\begin{align}
    \mathbb{E}\Bigg[\frac{1}{\sigma^2 \sqrt{N}}\sum_{m(\neq\nu)}^{M_u}x_{m i}F\Big(\tilde{p}_{m\to i}^{(t)},S_m^{(t)}\Big)\Bigg]
    \approx \frac{\alpha_u w_{0i}}{\sigma^2} \int Dz\ \bigg(\rho F\Big(\mathcal{P},S^{(t)}\Big)-(1-\rho) F\Big(\mathcal{Q},S^{(t)}\Big)\bigg),\Leq{Aeq102}\\
    \mathbb{V}\Bigg(\frac{1}{\sigma^2 \sqrt{N}}\sum_{m(\neq\nu)}^{M_u}x_{m i}F\Big(\tilde{p}_{m\to i}^{(t)},S_m^{(t)}\Big)\Bigg)
    \approx \frac{\alpha_u}{\sigma^2} \int Dz\ \bigg(\rho F^2\Big(\mathcal{P},S^{(t)}\Big) +(1-\rho) F^2\Big(\mathcal{Q} ,S^{(t)}\Big)\bigg).\Leq{Aeq103}
\end{align}
Therefore, SE iteration for RMLE is obtained as
\subbe
\be
    \chi_{t+1} &=& \Bigg(\lambda+\frac{\alpha}{\sigma^2} - \frac{\alpha_u}{\sigma^2} \int Dz\ \bigg(\rho T\Big(\mathcal{P},\frac{\chi_t}{\sigma^2}\Big)+(1-\rho)T\Big(\mathcal{Q} ,\frac{\chi_t}{\sigma^2}\Big) \bigg)\Bigg)^{-1},\Leq{Ase_rmle1}\\
    k_{t+1} &=& \chi_{t+1}\Bigg(\frac{\alpha_l}{\sigma^2}+\frac{\alpha_u}{\sigma^2}\int Dz\ \bigg(\rho F\Big(\mathcal{P} ,\frac{\chi_t}{\sigma^2}\Big)-(1-\rho) F\Big(\mathcal{Q} ,\frac{\chi_t}{\sigma^2}\Big)\bigg)\Bigg),\Leq{Ase_rmle2}\\
    v_{t+1} &=& \chi_{t+1}^2\Bigg(\frac{\alpha_l}{\sigma^2} + \frac{\alpha_u}{\sigma^2} \int Dz\ \bigg(\rho F^2\Big(\mathcal{P}, \frac{\chi_t}{\sigma^2}\big) +(1-\rho) F^2\Big(\mathcal{Q}, \frac{\chi_t}{\sigma^2}\Big)\bigg)\Bigg).\Leq{Ase_rmle3}
\ee
\Leq{Ase_rmle}
\subee
Similarly, by employing the same techniques, we derive the SE iteration for the Bayesian approach as
\subbe
\be
    \chi_{t+1,B} &=& \Bigg(\lambda+\frac{\alpha}{\sigma^2} - \frac{\alpha_u}{\sigma^2} \int Dz\ \bigg(\rho \tilde{T}\big(\mathcal{P}_{B}\big)+(1-\rho)\tilde{T}\big(\mathcal{Q}_{B}\big) \bigg)\Bigg)^{-1},\Leq{Ase_bo1}\\
    k_{t+1,B} &=& \chi_{t+1,B}\Bigg(\frac{\alpha_l}{\sigma^2}+\frac{\alpha_u}{\sigma^2}\int Dz\ \bigg(\rho \tilde{F}\big(\mathcal{P}_{B}\big)-(1-\rho) \tilde{F}\big(\mathcal{Q}_{B}\big)\bigg)\Bigg),\Leq{Ase_bo2}\\
    v_{t+1,B} &=& \chi_{t+1,B}^2\Bigg(\frac{\alpha_l}{\sigma^2} + \frac{\alpha_u}{\sigma^2} \int Dz\ \bigg(\rho \tilde{F}^2\big(\mathcal{P}_{B}\big)+(1-\rho) \tilde{F}^2\big(\mathcal{Q}_{B}\big)\bigg)\Bigg).\Leq{Ase_bo3}
\ee
\Leq{Ase_bo}
\subee

\subsection{Microscopic dynamical instability or RSB} \Lapp{sub_Micr}
\textcolor{black}{Although the SE iterations \NReq{SE_RMLE} converge to a stationary state, there is no guarantee that the microscopic variables updated by the AMP/BP algorithm, such as $\hat{w}_{i\to \nu}^{(t)}$, will also converge.} The \textcolor{black}{microscopic} dynamical instability of AMP/BP is known to be connected to RSB \cite{kabashima2003cdma}. RSB is a concept in physics that is connected to the emergence of the massive number of (exponentially many w.r.t. $N$) modes in the posterior distribution. Thus, RSB implies the computational difficulty (or even impossibility) in handling the posterior distribution. We clarify the parameter region where RSB occurs in the present problem. Let us discuss the local stability of the fixed point of \Req{ABP_RMLE}. \textcolor{black}{Similar analyses can be found in \cite{kabashima2003cdma,sakata2018approximate,takahashi2022macroscopic}.}

In the following discussion, we neglect the subscript dependence of $S$ and $\chi$ for the same reason we discussed in \Rsec{state_evolution}, yielding $S^{(t)} = \chi^{(t)}/\sigma^2$. In addition, we denote
\be
    \hat{w}_{i\to \nu}^{(t)} = \hat{w}_{i\to\nu}^*+ \Delta w_{i\to\nu}^{(t)},\Leq{eq57}
\ee
where $\hat{w}_{i\to\nu}^*$ is the fixed point solution of ABP interations \NReq{ABP_RMLE}, and $\Delta w_{i\to\mu}^{(t)}$ is a small perturbation. We pursue how $\Delta w$ evolves through \Req{ABP_RMLE}. Assuming the perturbation is small and taking the first-order corrections only, we have
\begin{align}
    \Delta \tilde{p}_{\nu\to i}^{(t)} &\approx \frac{1}{\sigma^2 \sqrt{N}}\sum_{j(\neq i)}^N x_{\nu j} \Delta w_{j\to\nu}^{(t)},\Leq{eq58}\\
    \Delta F_{\nu\to i} &\approx \frac{\partial F\big(p,\chi/\sigma^2\big)}{\partial p}\bigg|_{p=\tilde{p}_{\nu\to i}^*} \Delta \tilde{p}_{\nu\to i}^{(t)} = T\Big(\tilde{p}_{\nu\to i}^*,\frac{\chi}{\sigma^2}\Big)\Delta \tilde{p}_{\nu\to i}^{(t)},\Leq{eq59}\\
    \Delta w_{i\to\nu}^{(t+1)} &\approx \frac{\chi}{\sigma^2\sqrt{N}}\sum_{m(\neq \nu)}^{M_u} x_{m i} \Delta F_{m\to i} \approx \frac{\chi}{\sigma^4 N}\sum_{m(\neq \nu)}^{M_u} \sum_{j(\neq i)}^N x_{m i}x_{m j}T\Big(\tilde{p}_{m\to i}^*,\frac{\chi}{\sigma^2}\Big)\Delta w_{j\to m}^{(t)}, \Leq{eq60}
\end{align}
\textcolor{black}{where $\tilde{p}^*$ also denotes the fixed point solution of \Req{ABP_RMLE}. Next, we conduct a linear stability analysis of \Req{eq60}. Since the random variables $\bm{x}_m$ are i.i.d., CLT ensures that the right-hand side of \Req{eq60} is also a Gaussian random variable. Additionally, the dependence of $\Delta w_{i\to\nu}^{(t+1)}$ on the indices $i$ and $\nu$ are negligible, since the right-hand side of \Req{eq60} is expected to be asymptotically independent of $i$ and $\nu$. These observations enable us to assess the fixed point's stability by tracking the evolution of the first and second moments of $\Delta w_{i\to\nu}^{(t+1)}$ at each update.}

\textcolor{black}{The first moment of $\Delta w_{i\to\nu}^{(t+1)}$ becomes negligible because the mean of the random variables $\bm{x}_m$ are zero. The second moment can be expressed as the average of the squares of $\Delta w_{i\to\nu}^{(t+1)}$ as}
\textcolor{black}{
\begin{align}
\underbrace{\frac{1}{N M_u}\sum_{\nu}^{M_u}\sum_{i}^N\Big(\Delta w_{i\to\nu}^{(t+1)}\Big)^2}_{v_{t+1}} 
&\overset{(a)}{\approx} \frac{1}{N M_u}\sum_{\nu}^{M_u}\sum_{i}^N \frac{\chi^2}{\sigma^8 N^2}\Bigg(\sum_{m(\neq \nu)}^{M_u} \sum_{j(\neq i)}^N x_{m i}x_{m j}T\Big(\tilde{p}_{m\to i}^*,\frac{\chi}{\sigma^2}\Big)\Delta w_{j\to m}^{(t)}\Bigg)^2\nonumber\\
&\overset{(b)}{\approx} \frac{\chi^2}{\sigma^8 N^2}\overline{\Bigg(\sum_{m(\neq \nu)}^{M_u} \sum_{j(\neq i)}^N x_{m i}x_{m j}T\Big(\tilde{p}_{m\to i}^*,\frac{\chi}{\sigma^2}\Big)\Delta w_{j\to m}^{(t)}\Bigg)^2}\nonumber\\
 &\overset{(c)}{\approx} \underbrace{\bigg(\frac{\alpha_u \chi^2}{\sigma^4}\overline{T^2\Big(\tilde{p}_{m\to i}^*,\frac{\chi}{\sigma^2}\Big)}\bigg)}_{J}\underbrace{\overline{\Big(\Delta w_{j\to m}^{(t)}\Big)^2}}_{v_t},\Leq{mdi}
\end{align}
where $\overline{(\cdots)}$ denotes average over all the configurations. The step (a) is derived from \Req{eq60} by squaring both sides and taking the average. The step from (a) to (b) follows from the law of large numbers, which allows each term to be replaced by its average, denoted by $\overline{(\cdots)}$. Finally, the step from (b) to (c), the factors $x_{mi}$, $x_{mj}$, $T(\cdot)$, and $\Delta w_{j\to m}^{(t)}$ are replaced by their averages. This is valid because the correlation between $T(\tilde{p}_{m\to i}^*,\chi/\sigma^2)$ and $\Delta w_{j\to m}$  is asymptotically negligible in the present limit. The left-hand side of \Req{mdi} represents the variance of $\Delta w_{i\to\nu}^{(t+1)}$, which we denote by $v_{t+1}$, similar to the second term on the right-hand side. The first term on the right-hand side of \Req{mdi} is denoted by $J$. Therefore, the iteration can be represented as $v_{t+1} = J v_t$, and the fixed point $v_t = 0$ is stable if $J < 1$. Furthermore, the macroscopic variable $\overline{T^2\Big(\tilde{p}_{m\to i}^*,\chi/\sigma^2\Big)}$ can be expressed as}
\textcolor{black}{\begin{align}
\overline{T^2\Big(\tilde{p}_{m\to i}^*,\frac{\chi}{\sigma^2}\Big)} &\approx
      \rho \int Dz\ T^2\bigg(\frac{k_*}{\lambda_0\sigma^2} + 
      \sqrt{\frac{k_*^2/\lambda_0+v_*}{\sigma^2}} z,\frac{\chi}{\sigma^2}\bigg) \nonumber\\
      &\quad\quad\quad\quad\  + (1-\rho) \int Dz\ T^2\bigg(-\frac{k_*}{\lambda_0\sigma^2} + 
      \sqrt{\frac{k_*^2/\lambda_0+v_*}{\sigma^2}} z,\frac{\chi}{\sigma^2}\bigg),
\end{align}
where $k_*,~v_*$ are the fixed point solution of SE. Hence, the fixed point solution $v_t=0$ becomes unstable if
\begin{align}
1 < \frac{\alpha_u \chi^2}{\sigma^4} \Bigg(
      &\rho \int Dz\ T^2\bigg(\frac{k_*}{\lambda_0\sigma^2} + 
      \sqrt{\frac{k_*^2/\lambda_0+v_*}{\sigma^2}} z,\frac{\chi}{\sigma^2}\bigg) \nonumber\\
      &\quad\quad\quad\quad\  + (1-\rho) \int Dz\ T^2\bigg(-\frac{k_*}{\lambda_0\sigma^2} + 
      \sqrt{\frac{k_*^2/\lambda_0+v_*}{\sigma^2}} z,\frac{\chi}{\sigma^2}\bigg) \Bigg).
\Leq{gAT-line}
\end{align}}
\BReq{at-line} represents a special case of \Req{gAT-line} under $\rho=1/2$ and $\alpha_l=0$.

\subsection{\textcolor{black}{Numerical validation}\Lapp{numerical_validation}}
\subsubsection{Consistency check between AMP and SE}
We first check the consistency between AMP and SE of RMLE. The AMP iterations \NReq{AMP_RMLE} are terminated when $ \lVert \hat{\bm{w}}_{AMP}^{(t+1)}-\hat{\bm{w}}_{AMP}^{(t)}\rVert_2/\lVert\hat{\bm{w}}_{AMP}^{(t+1)}\rVert_2$ fell below a threshold value of $\varepsilon_{AMP}=10^{-8}$. We apply the same threshold value for terminating SE iterations \NReq{SE_RMLE}.

\Rfig{AMP_SE_v_k} shows the plots of the order parameters $k$ and $v$ against the iteration step. In each panel of the figure, both the SE (joined markers $``\times,+"$) and AMP (joined markers $``\bullet,\scriptstyle\blacktriangle"$ with standard deviation depicted by shaded area) results are simultaneously shown. The agreement between these two curves is excellent (the SE curve is hard to see due to the overlap with the AMP curve), demonstrating the consistency between these two computations in the RMLE case.

\begin{figure}[H]
\centering
\includegraphics[width=0.49\textwidth]{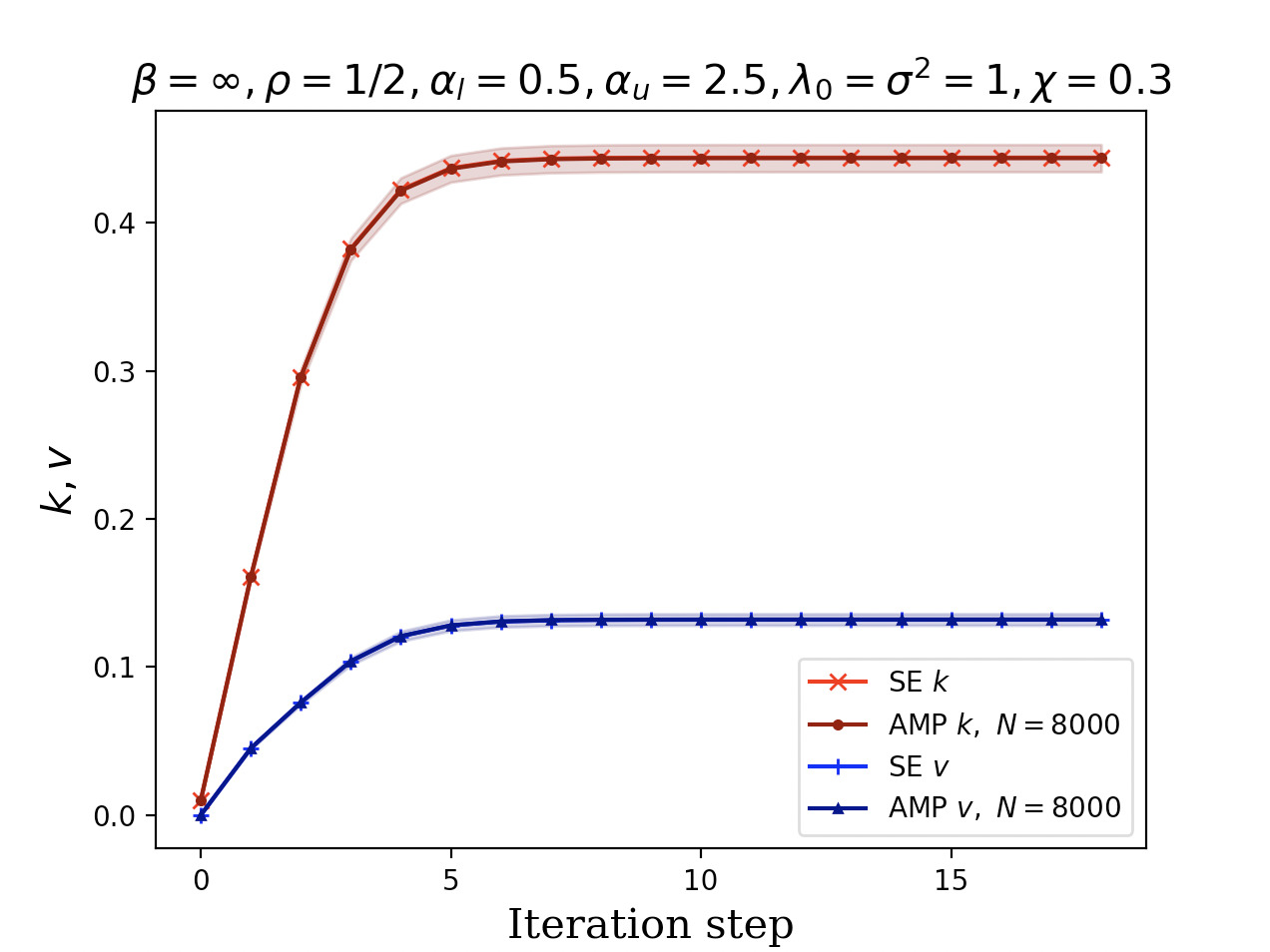}
\includegraphics[width=0.49\textwidth]{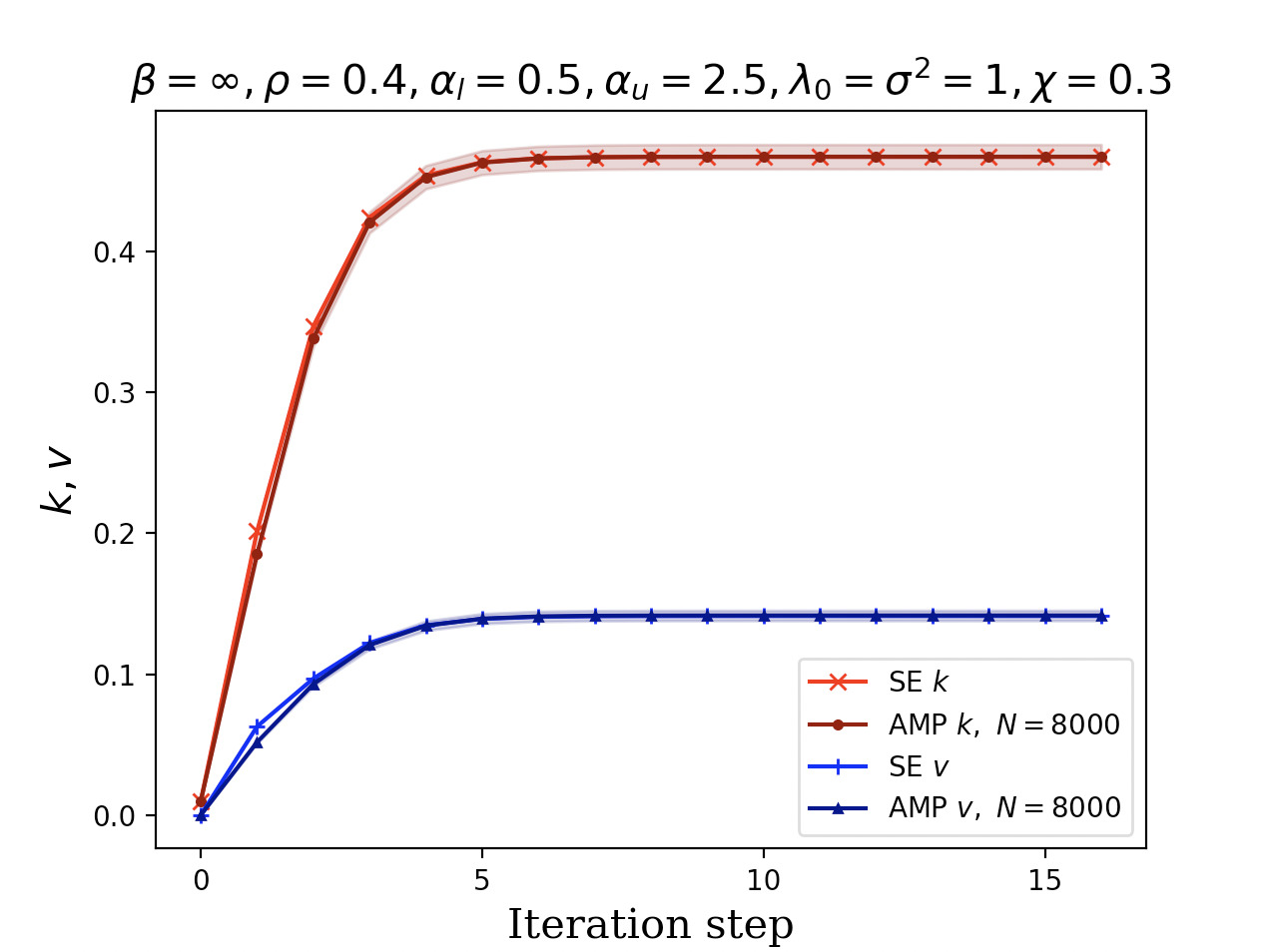}
\\
\includegraphics[width=0.49\textwidth]{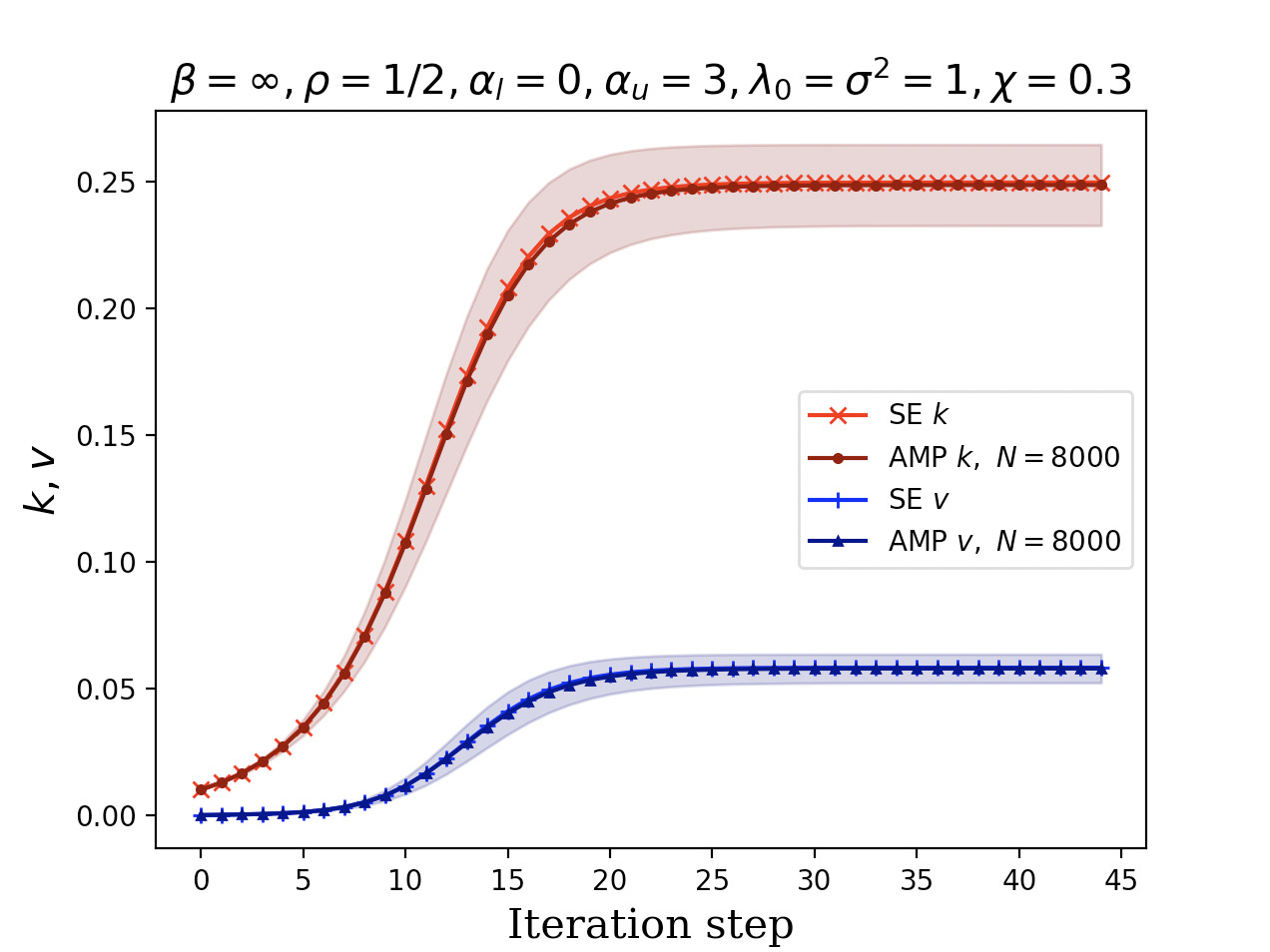}
\includegraphics[width=0.49\textwidth]{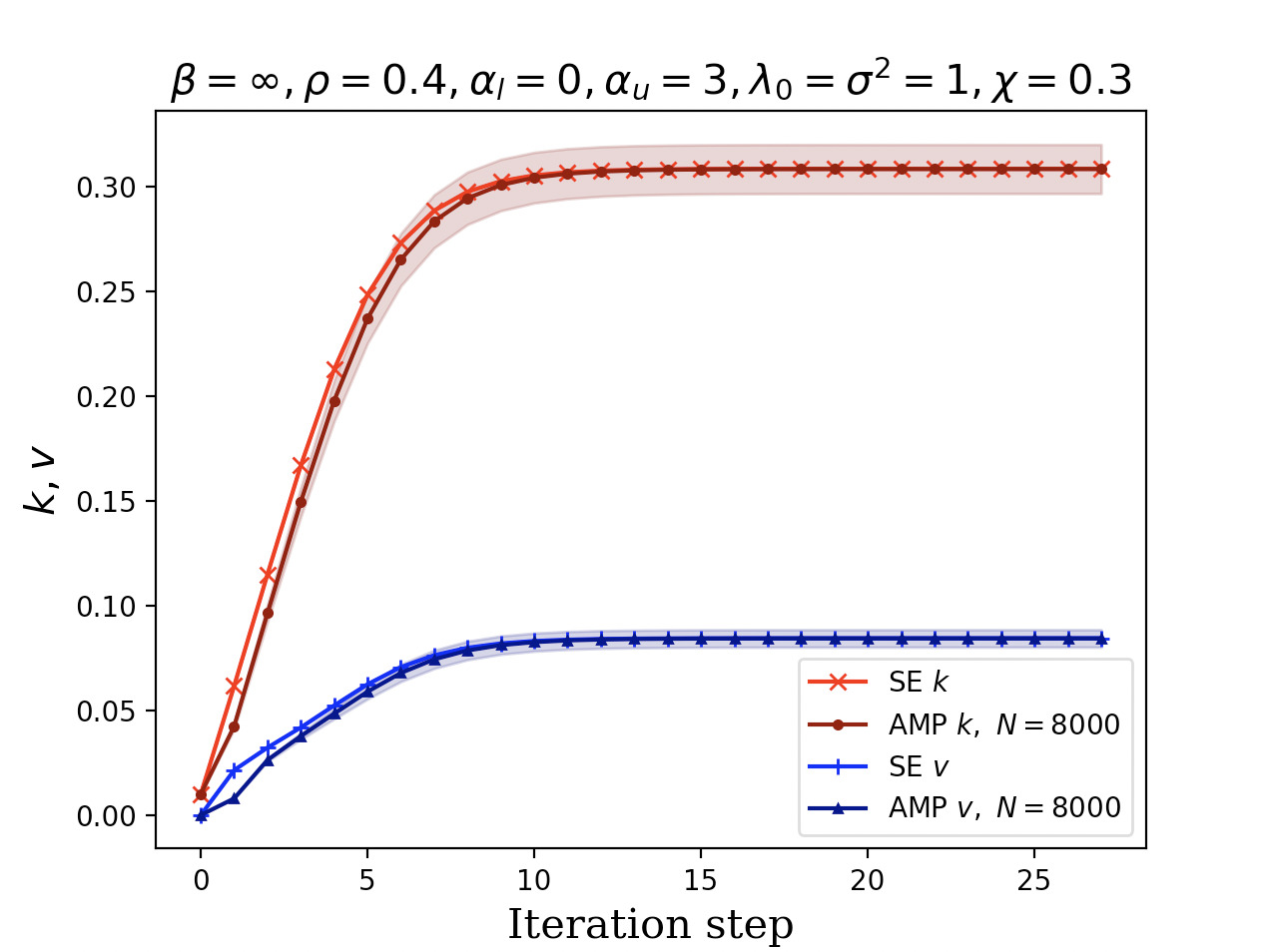}
\caption{Consistency check between AMP and SE in the RMLE case. The upper and bottom panels are for $(\alpha_l,\alpha_u)=(0.5,2.5)$ and $(\alpha_l,\alpha_u)=(0,3)$, respectively, making a comparison between the semi-supervised and unsupervised cases. The left and right panels are for $\rho=0.5$ and $\rho=0.4$, respectively, making a comparison between the balanced and imbalanced cases. Other parameters are fixed as $\lambda_0=\sigma^2=1$ and $\chi=0.3$. The joined markers $``\times,+"$ represent the SE results obtained by \Req{SE_RMLE}, while the joined markers $``\bullet,\scriptstyle\blacktriangle"$ denote the AMP result by \Req{AMP_RMLE} at $N=8000$: for AMP, the 100 independent runs are conducted to obtain the markers and the shaded area describing the standard deviation.
}
\Lfig{AMP_SE_v_k}
\end{figure}

Next, we conduct the same consistency check for the Bayesian approach. \Rfig{AMP_SE_v_k_BA} is the Bayesian case counterpart of \Rfig{AMP_SE_v_k}. This exhibits the nice agreement between the SE and AMP results again, showing the consistency in the Bayesian case.  
\begin{figure}[H]
\centering
\includegraphics[width=0.49\textwidth]{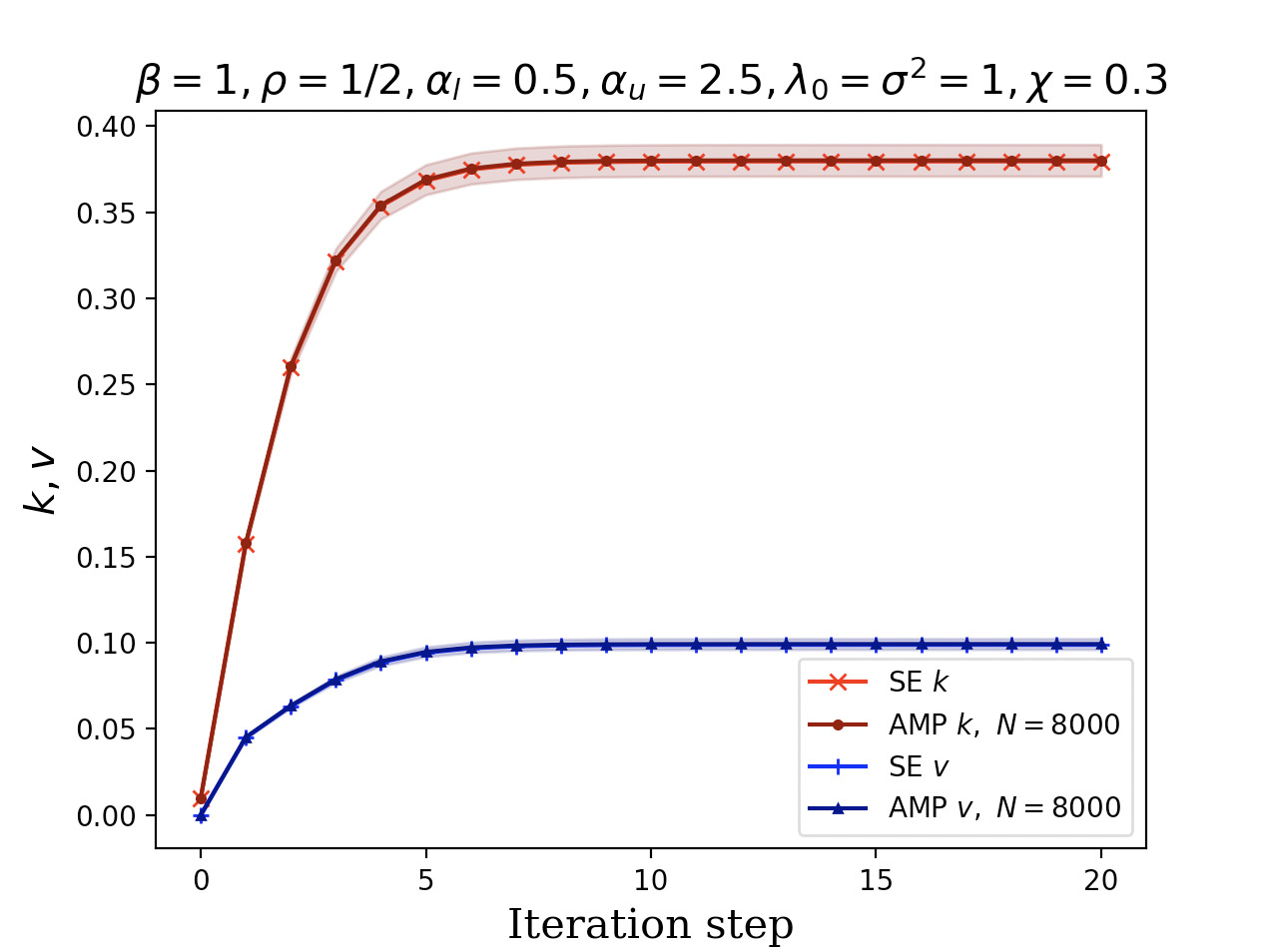}
\includegraphics[width=0.49\textwidth]{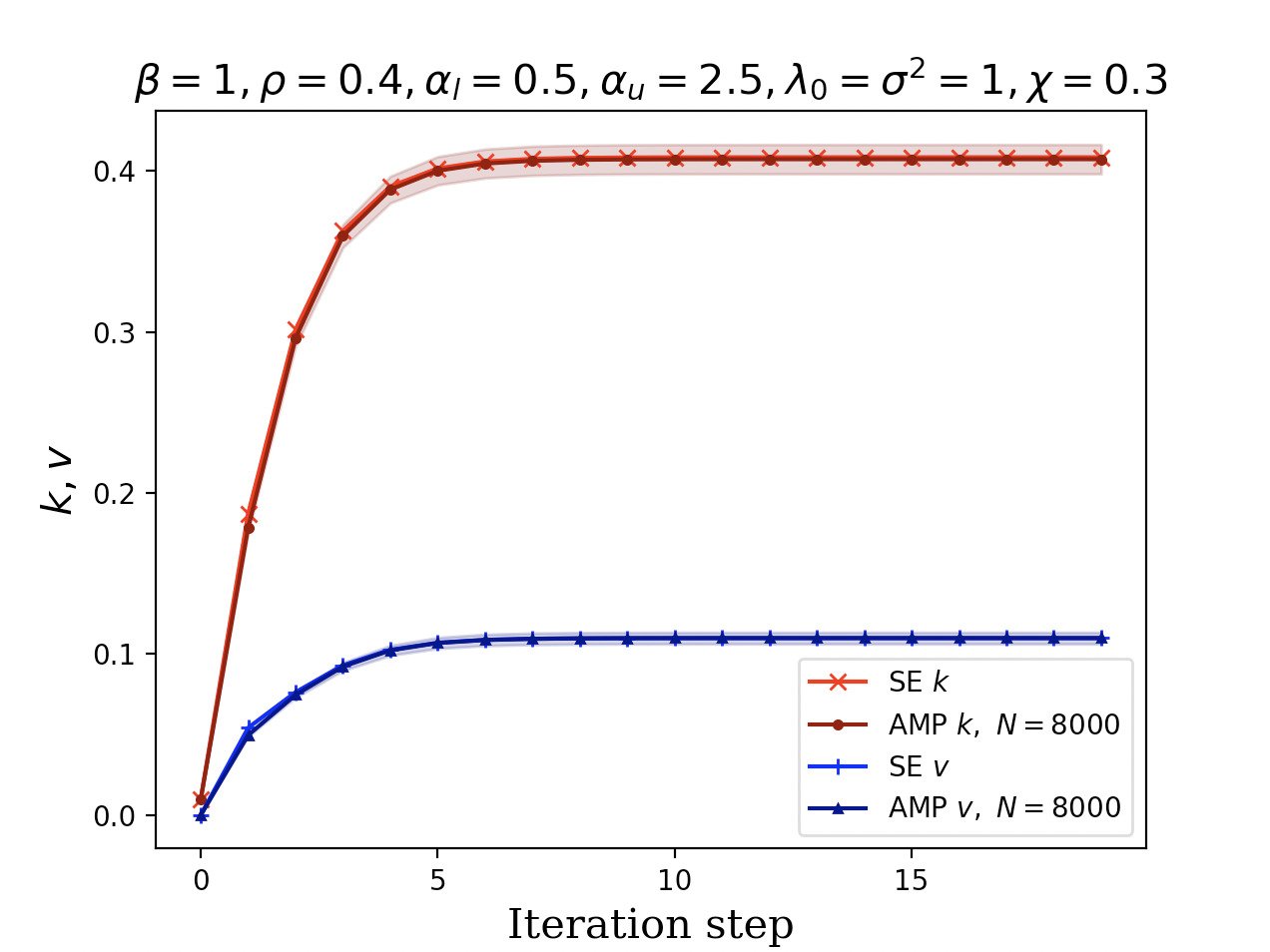}
\\
\includegraphics[width=0.49\textwidth]{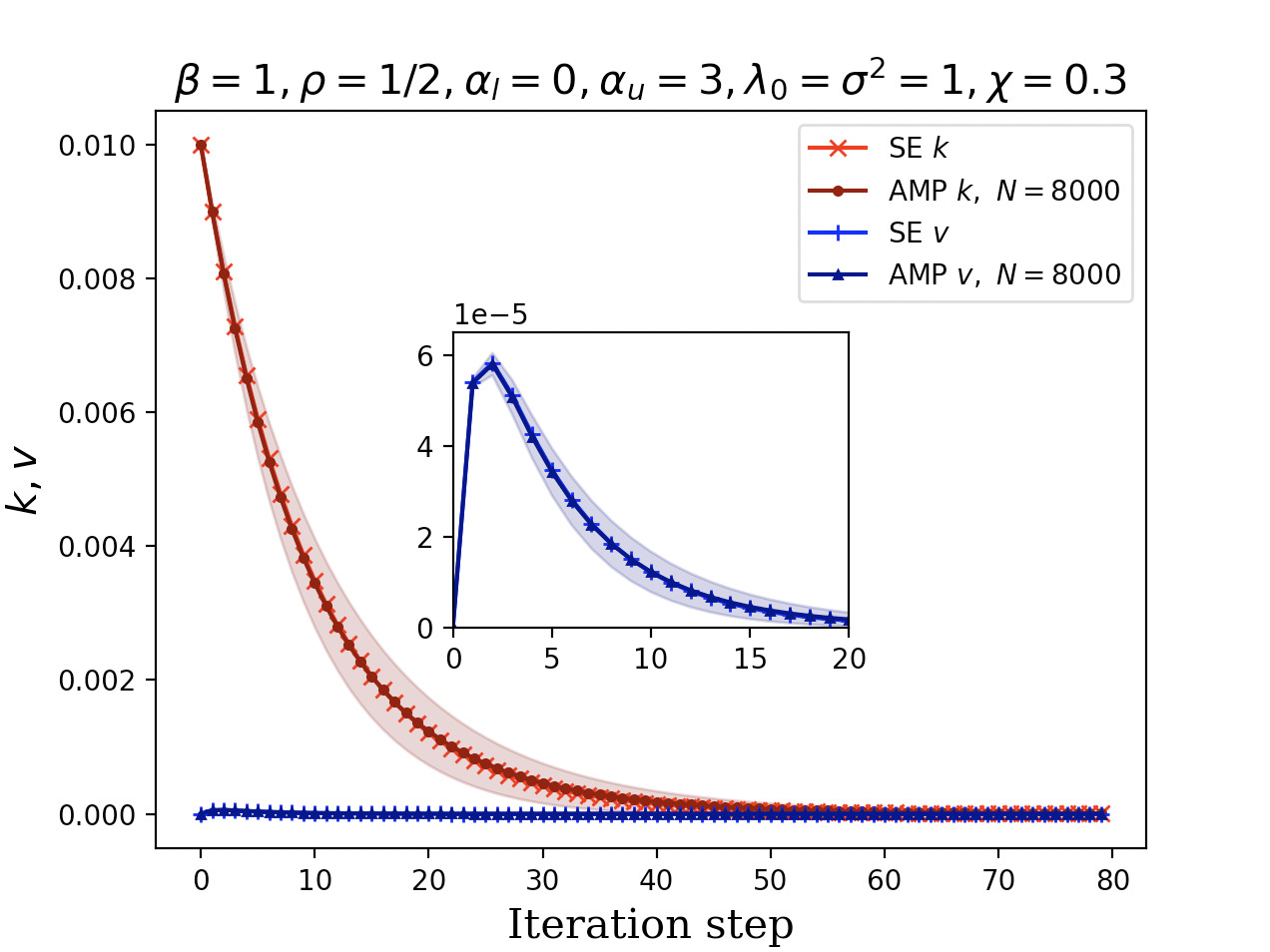}
\includegraphics[width=0.49\textwidth]{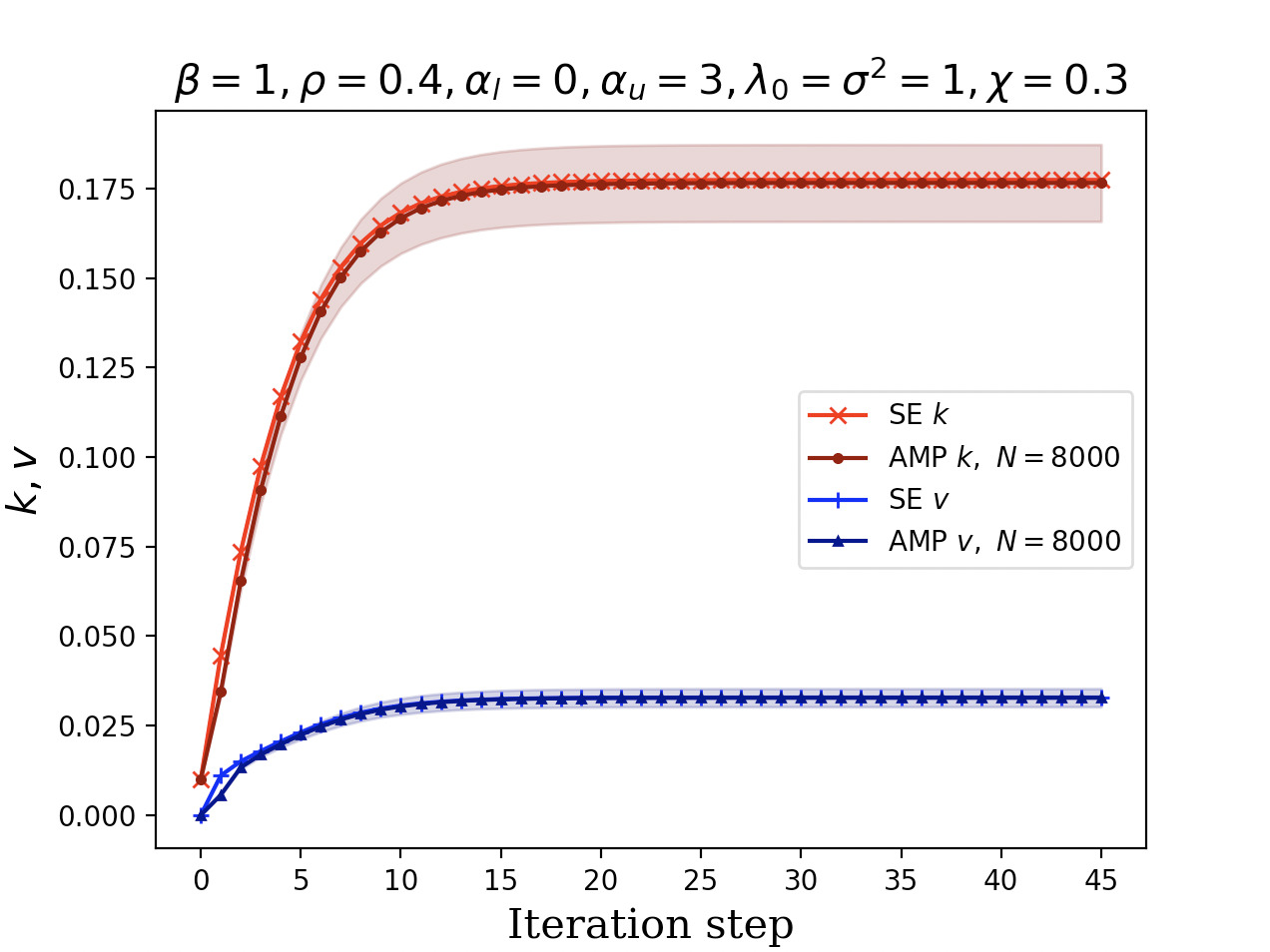}
\caption{Consistency check between AMP and SE in the Bayesian case. This is the counterpart of \Rfig{AMP_SE_v_k}, and the parameter values are set to those of the respective panels of that figure.
}
\Lfig{AMP_SE_v_k_BA}
\end{figure}

\subsubsection{Validity check of AMP through experiments using the off-the-shelf algorithms} \Lapp{VC_amp_gd}
To check the AMP’s effectiveness as a solver for obtaining the estimator in \Req{eq7} in the RMLE case, we conduct another numerical experiment using gradient descent (GD) that minimizes the objective function in the \textcolor{black}{right-hand} side of \Req{eq7}. For this, we define the error between the estimators from GD and AMP as $\Delta_{GD, AMP} = \lVert \hat{\bm{w}}_{GD}(\lambda) - \hat{\bm{w}}_{AMP}(\chi) \rVert_2/\lVert \hat{\bm{w}}_{GD}(\lambda) \rVert_2$ to quantify the consistency. To ensure an accurate comparison, several considerations must be carefully addressed. Firstly, the treatment for the parameter of GD and the order parameter of AMP requires special care when we compare the difference between $\hat{\bm{w}}_{GD}(\lambda)$ and $\hat{\bm{w}}_{AMP}(\chi)$. This is because the GD algorithm provides an estimator as a function of $\lambda$, $\hat{\bm{w}}_{\rm GD}(\lambda)$, while our AMP gives the estimator given $\chi$, $\hat{\bm{w}}_{\rm AMP}(\chi)$. To compare these two estimators, we use the $\lambda$-$\chi$ correspondence discussed in \Rsec{sub_Conv}. Secondly, AMP is an asymptotically exact algorithm and it does not precisely minimize the objective function at finite system size $N$. Therefore, a scaling analysis w.r.t. $N$ in the large $N$ limit is necessary for accurate comparison. Thirdly, The GD algorithm stops after a finite number of steps, which introduces some deviations from the true minimizer of the objective function. In our GD, the algorithm stops when the updated width $\lVert\hat{\bm{w}}_{GD}^{t+1}-\hat{\bm{w}}_{GD}^{t}\rVert_2/\lVert\hat{\bm{w}}_{GD}^{t+1}\rVert_2$ becomes smaller than a threshold value $\varepsilon_{GD}$, and the dependence on this threshold must be examined. Additionally, as the magnitude of the learning rate $\eta$ also affects the result, the dependence on $\eta$ must also be investigated.

\begin{figure}[htbp]
\centering
\includegraphics[width=0.49\textwidth]{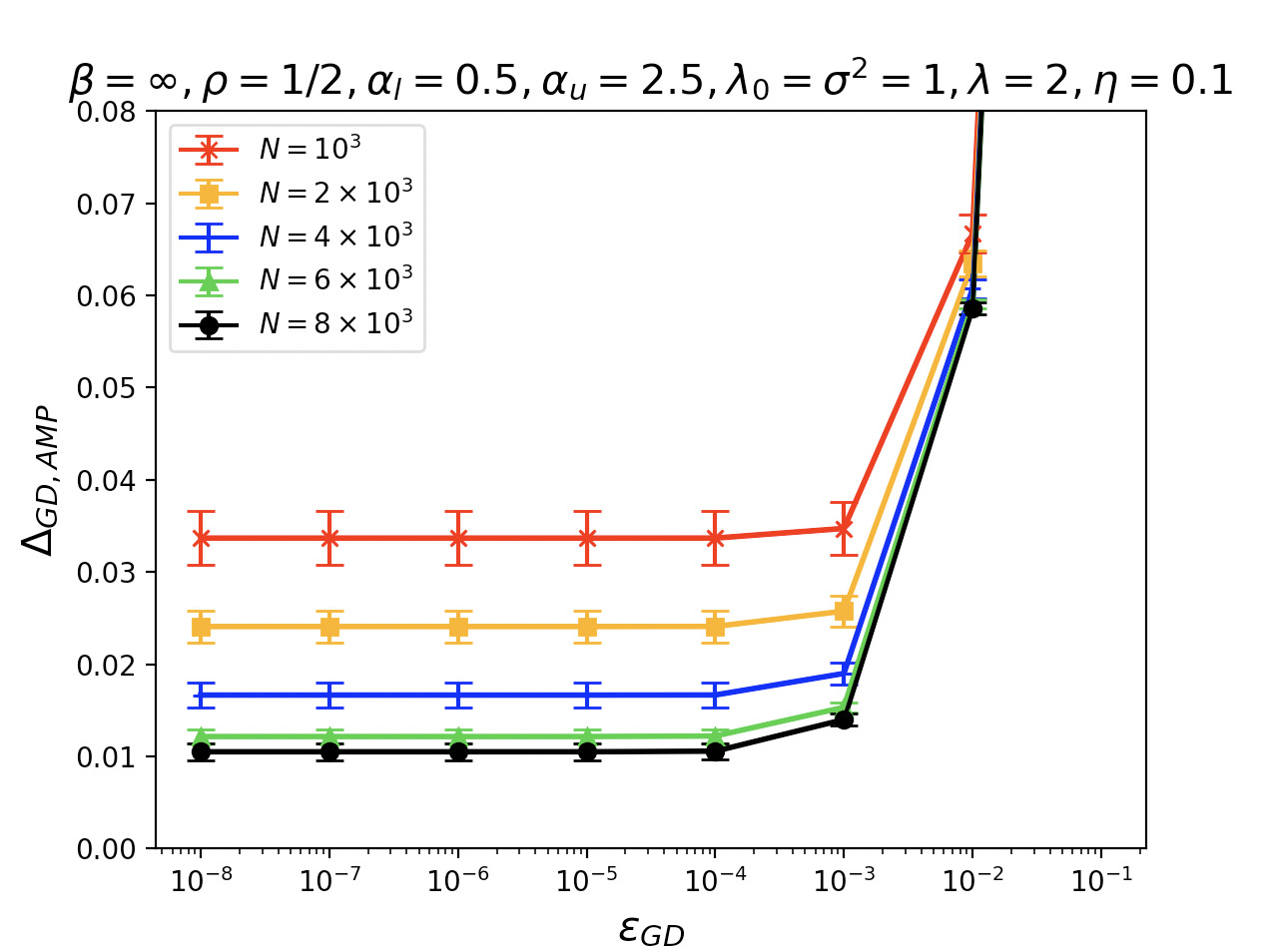}
\includegraphics[width=0.49\textwidth]{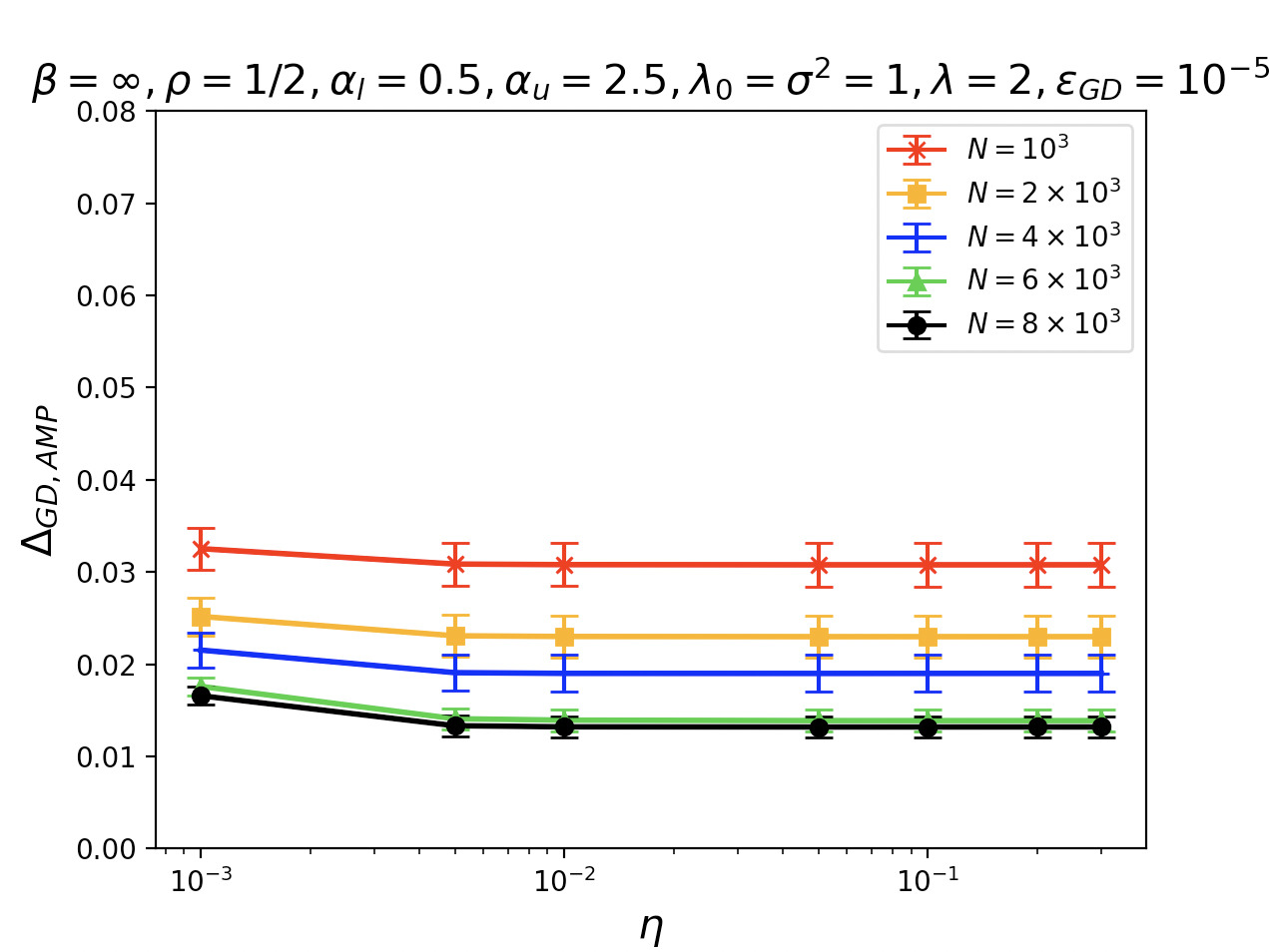}
\caption{Dependence of  $\Delta_{GD, AMP}$ on the threshold value ($\varepsilon_{GD}$) and learning rate ($\eta$). In the left panel, $\Delta_{GD, AMP}$ is plotted against $\varepsilon_{GD}$ with fixed $\eta=0.1$. Each colored curve with joined markers and error bars represents different system sizes $N$. In the right panel, $\Delta_{GD, AMP}$ is plotted against $\eta$ with fixed $\varepsilon_{GD}=10^{-5}$. Other parameters are $\rho=0.5$, $(\alpha_l,\alpha_u)=(0.5,2.5)$, $\lambda_0=\sigma^2=1$. The regularization parameter for GD is $\lambda=2$, and the corresponding $\chi$ for AMP is determined using the strategy discussed in \Rsec{sub_Conv}.}
\Lfig{GD_epsilon_eta}
\end{figure}

Based on the aforementioned considerations, we show the plots of   $\Delta_{GD, AMP}$ against $\varepsilon_{GD}$ and $\eta$ at several values of $N$ in \Rfig{GD_epsilon_eta}. In the left panel, the plot against $\varepsilon_{GD}$ with different values at $\eta=0.1$ is shown. When $\varepsilon_{GD} \leq 10^{-5}$, we can see that $\Delta_{GD, AMP}$ tends to be constant for all the examined system sizes, implying that the error due to the finiteness of $\varepsilon_{GD}$ is negligible in this region. Similarly, the right panel shows $\Delta_{GD, AMP}$ plotted against $\eta$ when $\varepsilon_{GD}$ is fixed at $10^{-5}$. $\Delta_{GD, AMP}$ remains constant when $\eta$ is large enough in the examined range, suggesting that the result is not sensitive against the choice of the learning rate in that region. According to these observations, we choose $\varepsilon_{GD}=10^{-5}$ and $\eta = 0.1$ as the appropriate values of these parameters for the large $N$ analysis below.

\begin{figure}[H]
\centering
\includegraphics[width=0.32\textwidth]{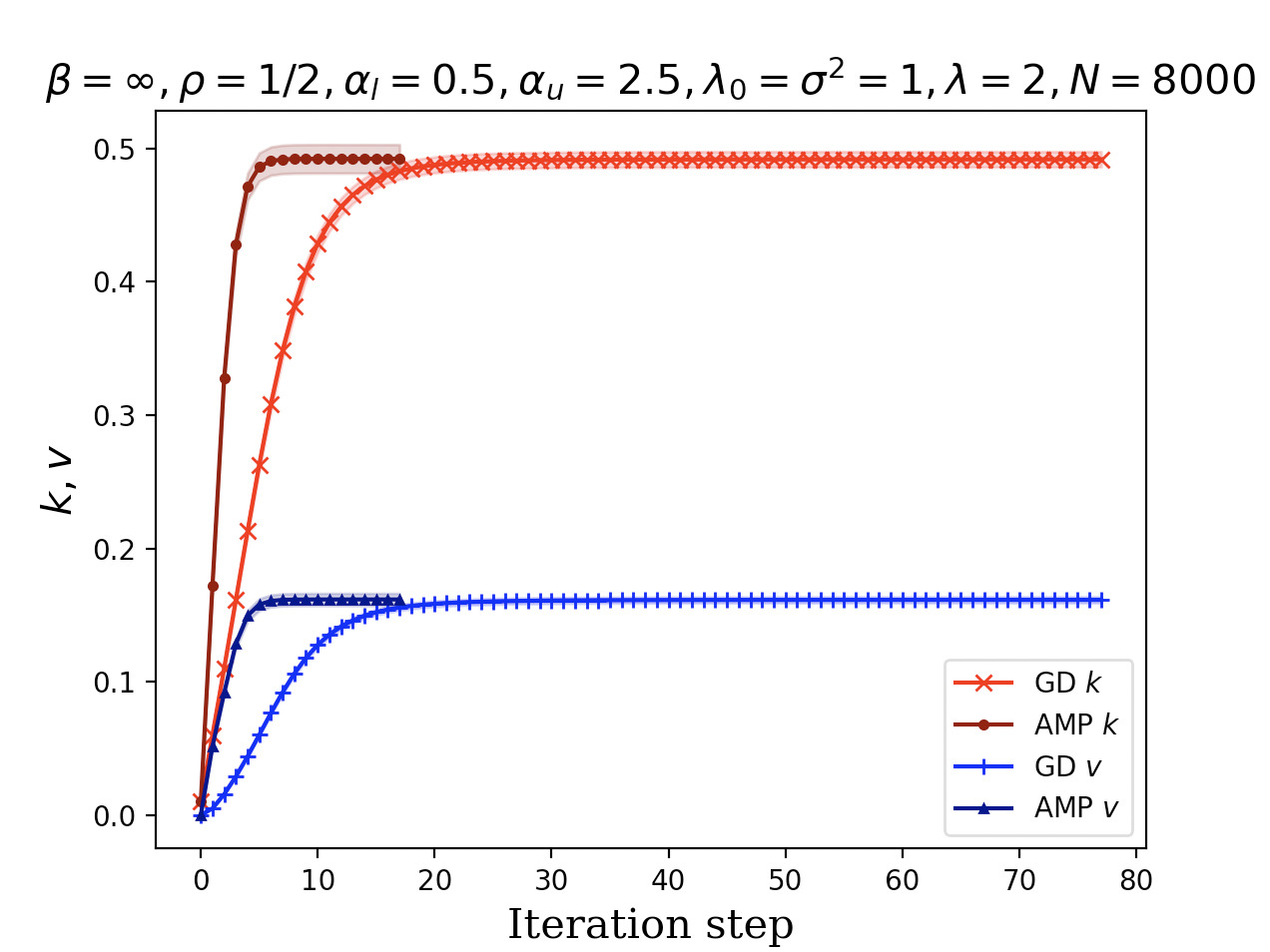}
\includegraphics[width=0.32\textwidth]{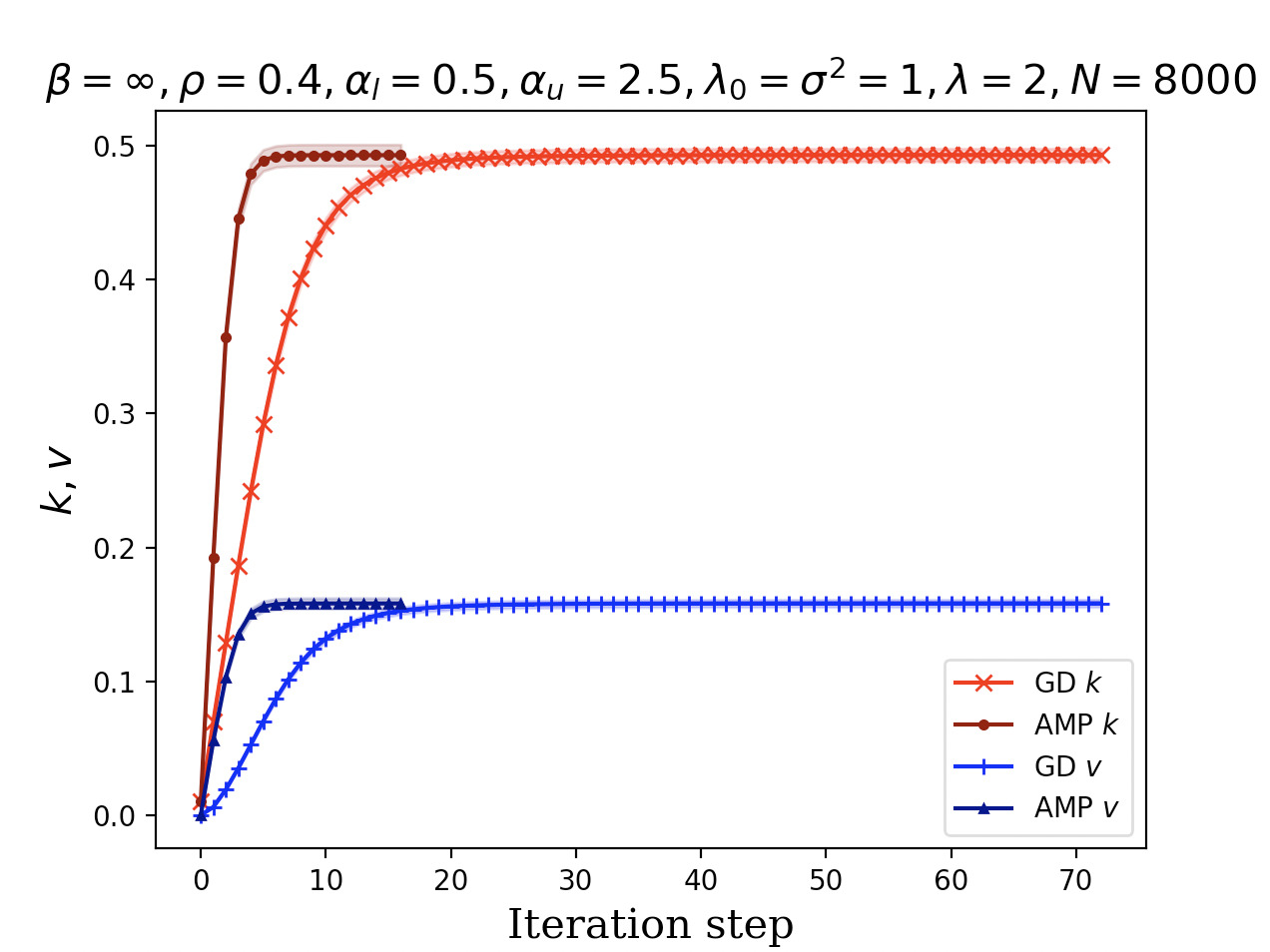}
\includegraphics[width=0.32\textwidth]{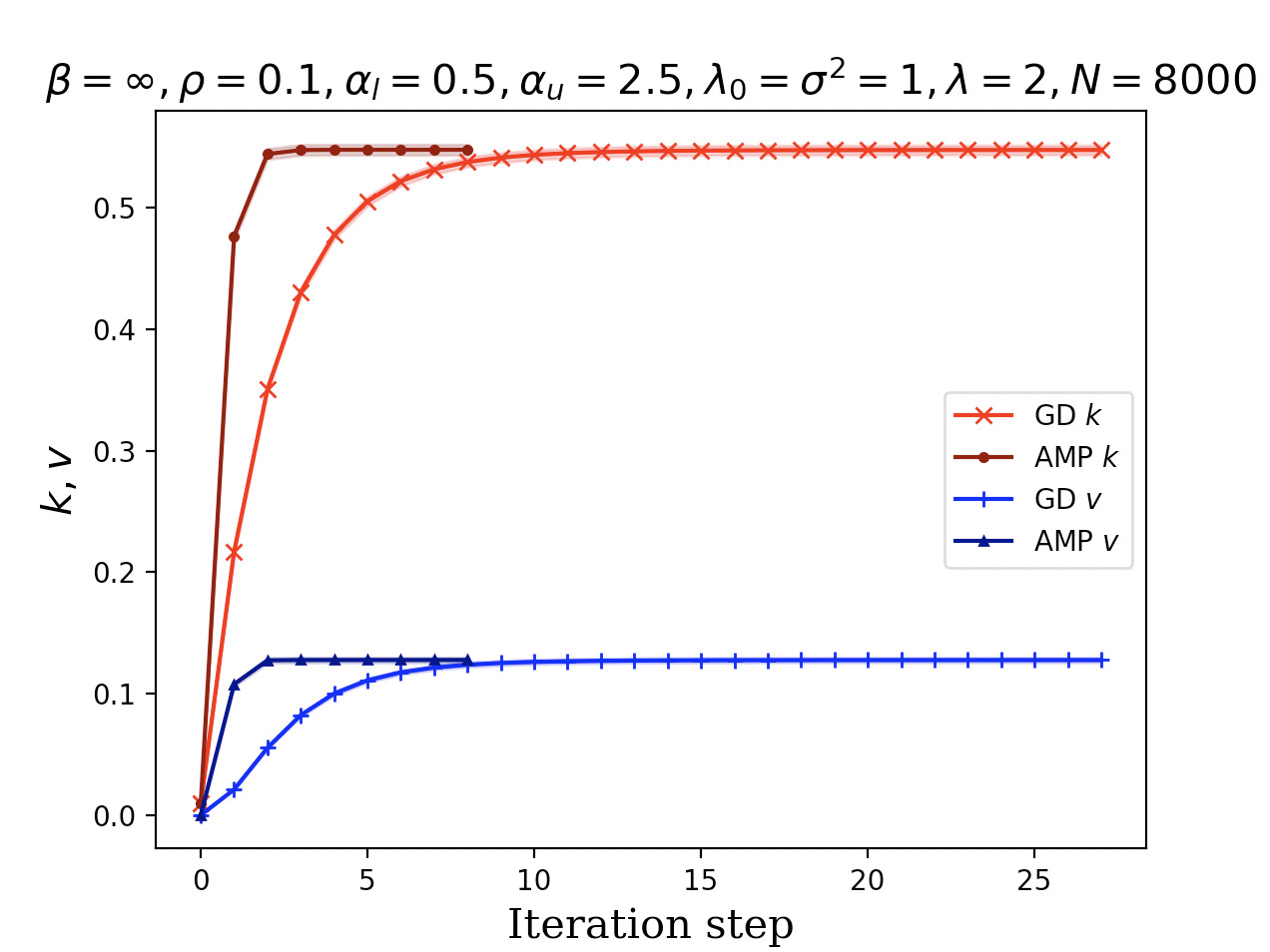}
\\
\includegraphics[width=0.32\textwidth]{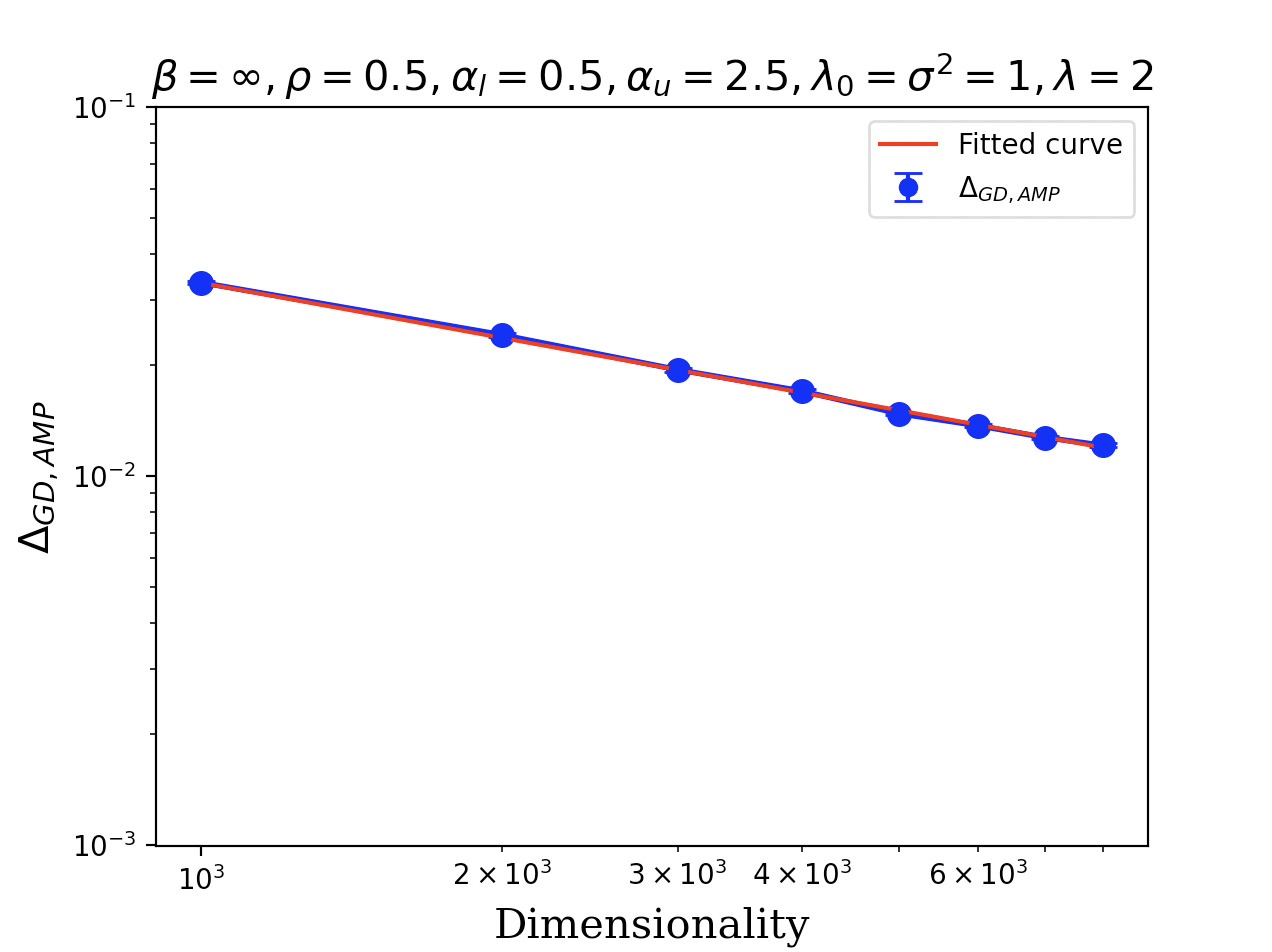}
\includegraphics[width=0.32\textwidth]{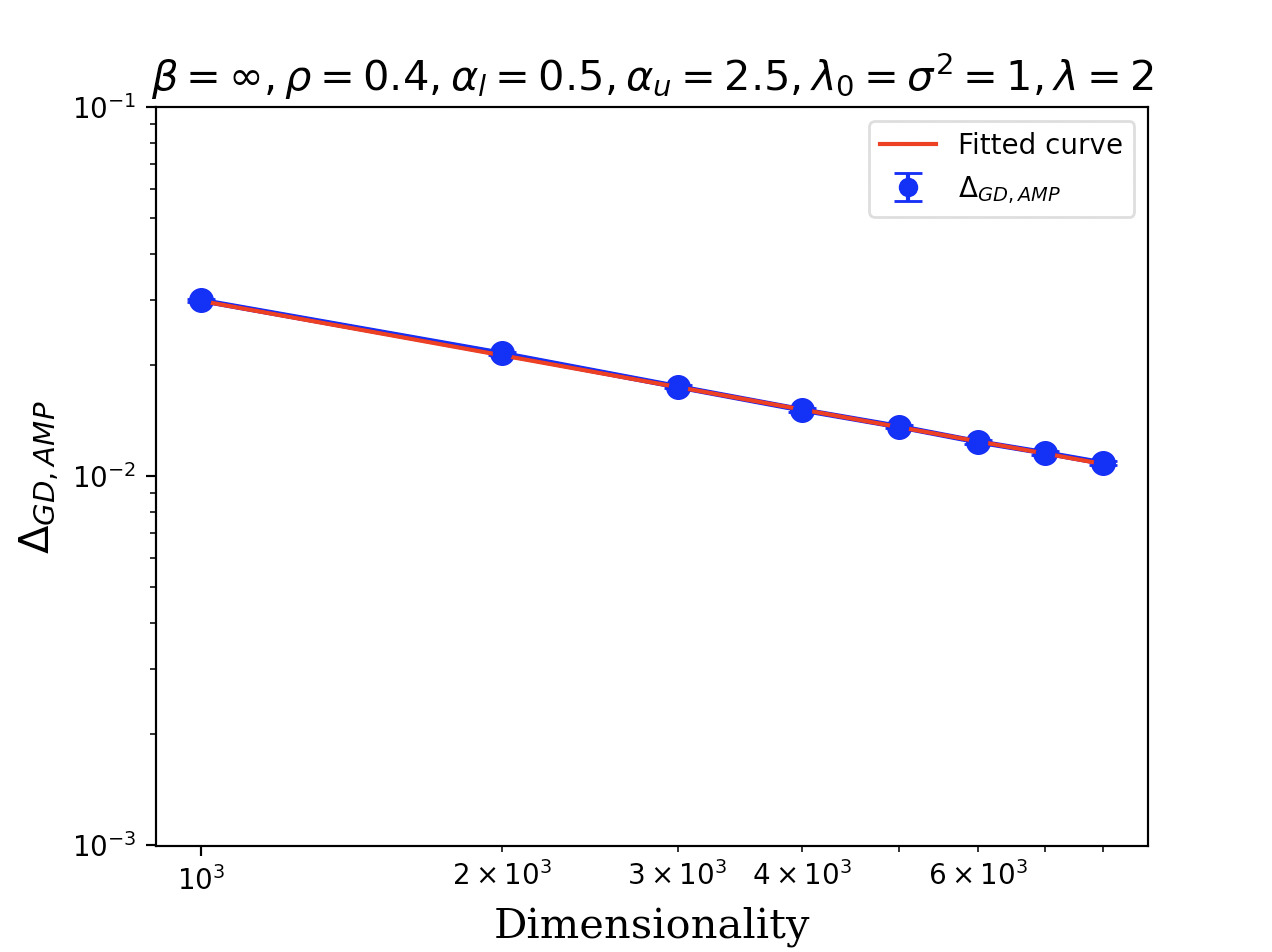}
\includegraphics[width=0.32\textwidth]{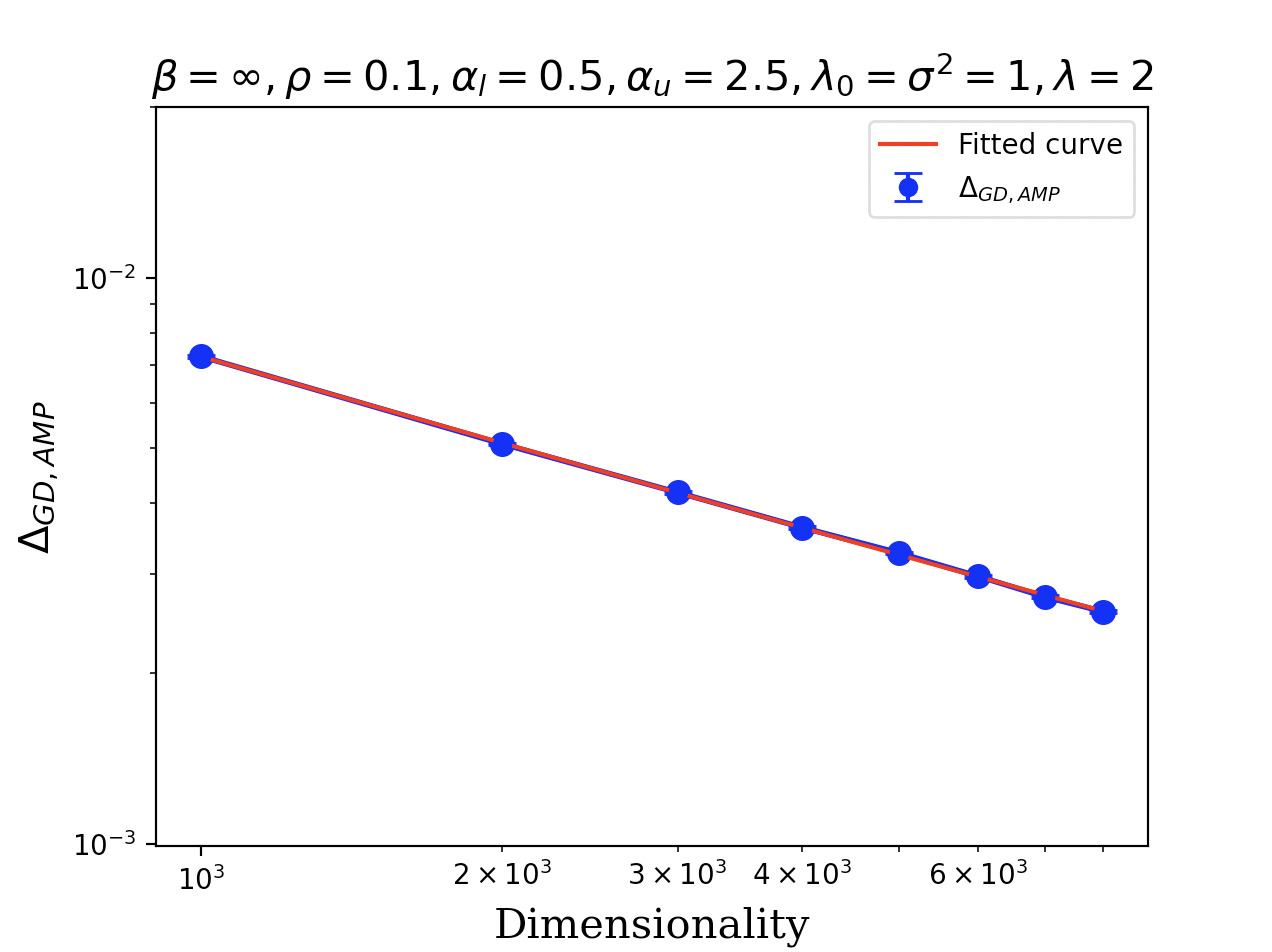}
\\
\includegraphics[width=0.32\textwidth]{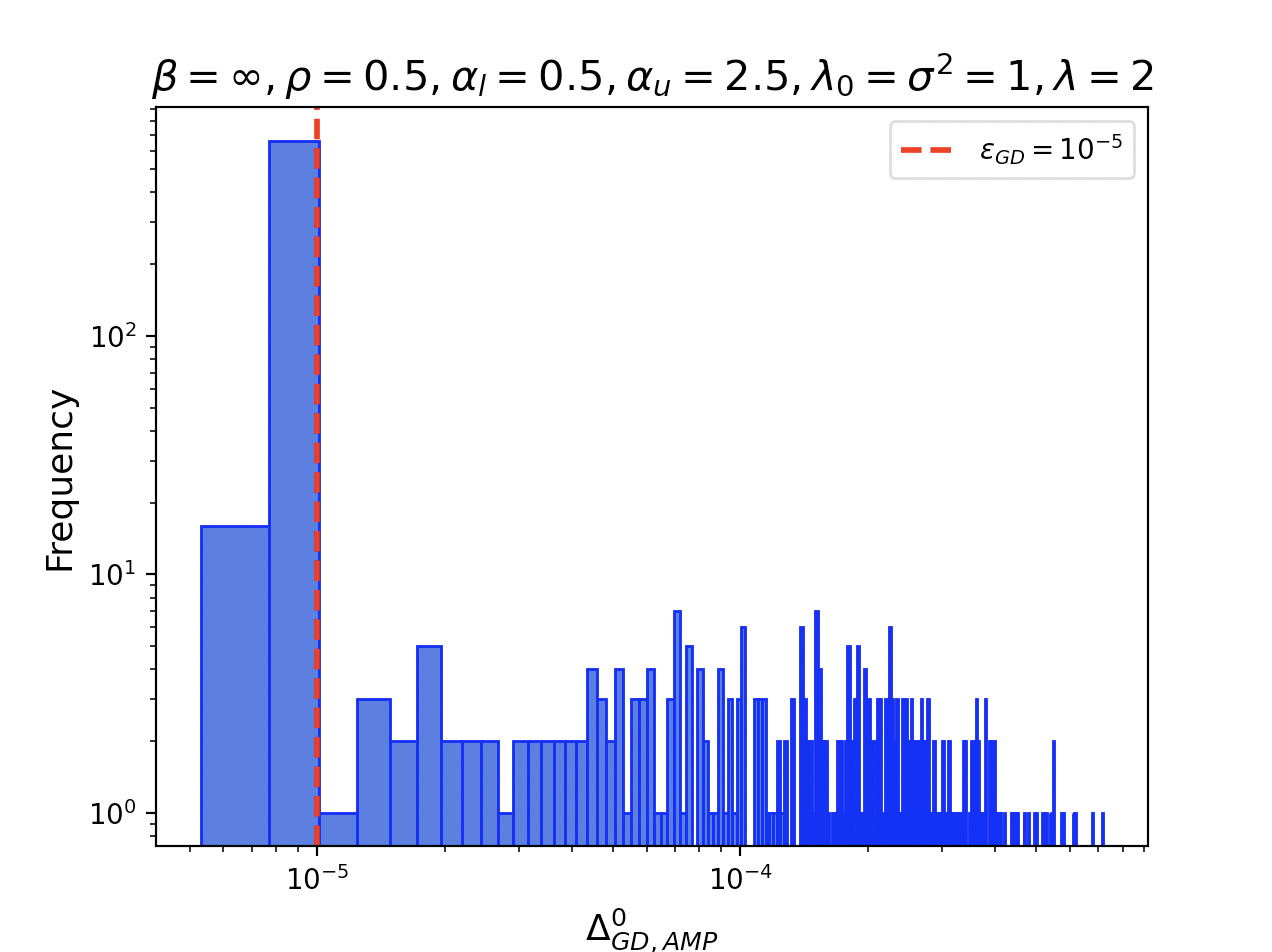}
\includegraphics[width=0.32\textwidth]{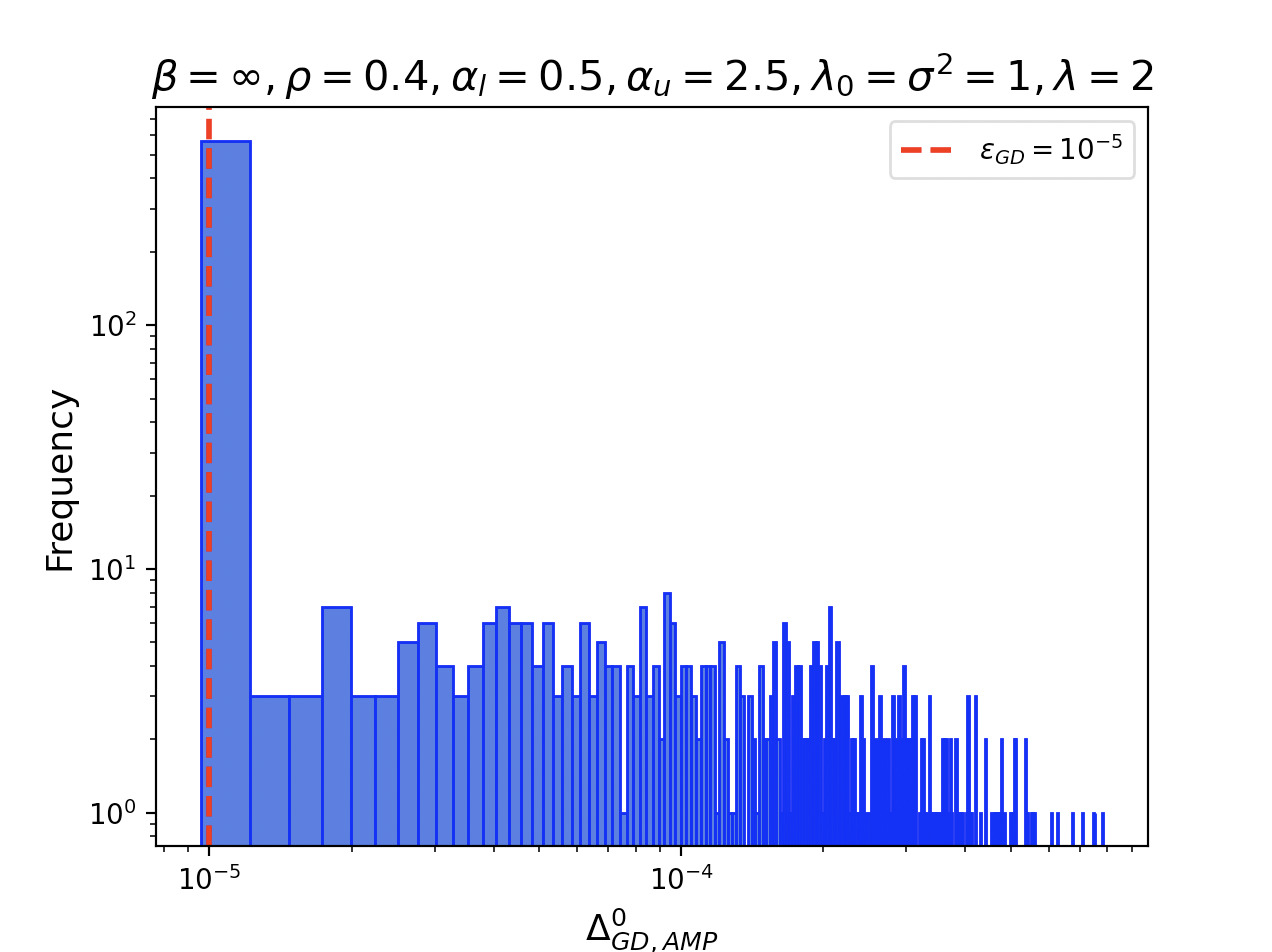}
\includegraphics[width=0.32\textwidth]{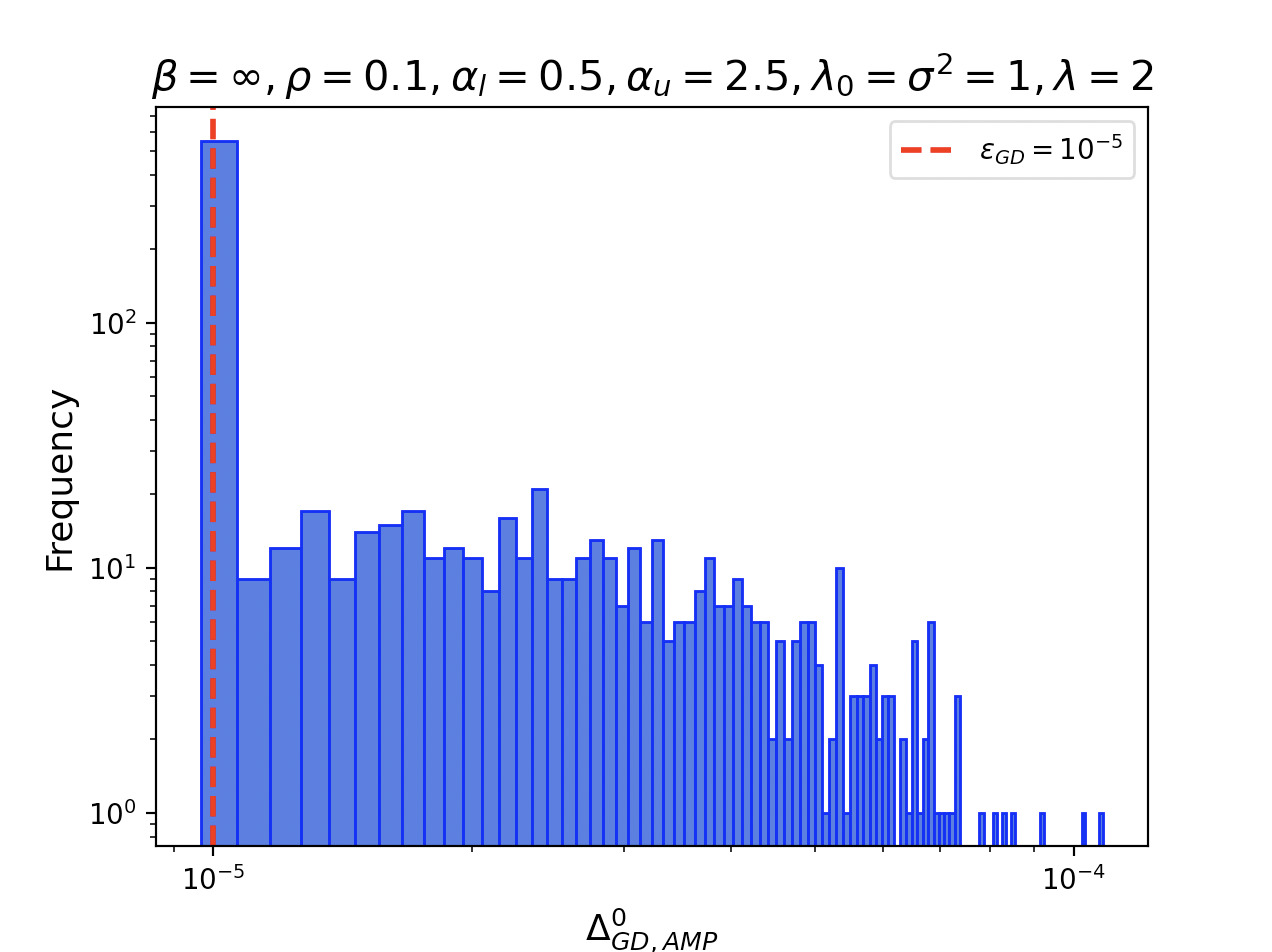}
\caption{\label{fig:ga_amp} Consistency check between AMP and GD in large $N$ region in the RMLE case: three different values of $\rho$, $0.5, 0.4, 0.1$ from left to right, are examined; other parameters are $\lambda_0=\sigma^2=1, \lambda=2$. In the upper panels, we plot the order parameters $k$ and $v$ of AMP and GD against the iteration step at $N=8000$. The joined markers $``\bullet,\scriptstyle\blacktriangle"$ represent the AMP results, while the joined markers $``\times,+"$ denote the GD results. The middle panels depict $\Delta_{GD, AMP}$ against the dimensionality $N$. The blue-line-joined marker represents the mean values of the $\Delta_{GD, AMP}$, with error bars, derived from 1000 experiments. The red curve represents the fitted curve for these blue points. In the bottom panels, we plot histograms of  $\Delta_{GD, AMP}^0$ and compare them with the red dashed line, $\varepsilon_{GD}$.} 
\Lfig{AMP_GA}
\end{figure}

Given the values of $\varepsilon_{GD}$ and $\eta$, we examine the order parameters and $\Delta_{GD, AMP}$ in the large $N$ region. \Rfig{AMP_GA} gives the result of the examination: the top panels show the plots of the order parameters $k$ and $v$ against the iteration step of GD and AMP at $N=8000$ with different $\rho$ values. Notably, the order parameters computed from the AMP and GD algorithms converge very close values to each other in the long time limit, suggesting consistency between them. For the convergence speed, AMP is clearly superior to GD. The middle panels display the plots of $\Delta_{GD, AMP}$ against $N$ in the log-log scale. The blue points with error bars represent the values of $\Delta_{GD, AMP}$ for finite $N$ from $1000$ to $8000$. The linear decreasing of $\Delta_{GD, AMP}$ is clearly seen, suggesting the power law. Based on this observation, to extract the behavior in the large $N$ limit, we fit the finite $N$ data points by $\Delta_{GD, AMP}(N) = \Delta_{GD, AMP}^0 + aN^{-d}$. The red curve represents the fitted curve: the fitted values of the parameters are $(\Delta_{GD, AMP}^0, a, d) = \big\{(1.0*10^{-5},1.0, 0.49), (1.6*10^{-5}, 0.88, 0.49), (1.0*10^{-5}, 0.23, 0.50)\big\}$ for $\rho = \{0.5, 0.4, 0.1\}$ respectively. In the ideal case where $\varepsilon_{GD}\to 0$ and $N\to\infty$, the AMP and GD results are expected to be identical, meaning that $\Delta_{GD, AMP}^0$ will vanish in that case. For the present situation with finite $\varepsilon_{GD}$, $\Delta_{GD, AMP}^0$ will thus be scaled by $\varepsilon_{GD}$. To validate this point, we compute the distribution of $\Delta_{GD, AMP}^0$ using the bootstrap method that resamples the 1000 samples used for computing the blue points. The result is shown in the bottom panels. The distribution of $\Delta_{GD, AMP}^0$ is clearly comparable with $\varepsilon_{GD}$, supporting the above expectation. These analyses strongly support the consistency of the AMP and GD algorithms.

For validating the Bayesian AMP as the RMLE case one should make a comparison with another off-the-shelf algorithm, and Markov chain Monte Carlo (MCMC) is a reasonable and almost the only choice. However, MCMC often requires many iterations and tends to be slow. Additionally, we observed that the replica symmetric (RS) solution in eqs.~(29-31) by \cite{tanaka2013statistical} is identical to our SE iteration \NReq{SE_BA} when $\rho=1/2$ and $\lambda_0=\lambda$. This agreement between the present study and the previous one \cite{tanaka2013statistical} supports the correctness of our AMP. Due to these reasons, we do not conduct a direct verification of AMP for $\beta=1$.

\bibliographystyle{unsrt}
\bibliography{JSTAT_major_revision_gu_obuchi}

\begin{thebibliography}{10}

\bibitem{zhu2005semi}
Xiaojin~Jerry Zhu.
\newblock Semi-supervised learning literature survey.
\newblock 2005.

\bibitem{books/mit/06/CSZ2006}
Olivier Chapelle, Bernhard Schölkopf, and Alexander Zien, editors.
\newblock {\em Semi-Supervised Learning}.
\newblock The MIT Press, 2006.

\bibitem{yalniz2019billion}
I~Zeki Yalniz, Herv{\'e} J{\'e}gou, Kan Chen, Manohar Paluri, and Dhruv
  Mahajan.
\newblock Billion-scale semi-supervised learning for image classification.
\newblock {\em arXiv preprint arXiv:1905.00546}, 2019.

\bibitem{zhang2020pushing}
Yu~Zhang, James Qin, Daniel~S Park, Wei Han, Chung-Cheng Chiu, Ruoming Pang,
  Quoc~V Le, and Yonghui Wu.
\newblock Pushing the limits of semi-supervised learning for automatic speech
  recognition.
\newblock {\em arXiv preprint arXiv:2010.10504}, 2020.

\bibitem{berthelot2019mixmatch}
David Berthelot, Nicholas Carlini, Ian Goodfellow, Nicolas Papernot, Avital
  Oliver, and Colin~A Raffel.
\newblock Mixmatch: A holistic approach to semi-supervised learning.
\newblock {\em Advances in neural information processing systems}, 32, 2019.

\bibitem{lelarge2019asymptotic}
Marc Lelarge and L{\'e}o Miolane.
\newblock Asymptotic {Bayes} risk for {Gaussian} mixture in a semi-supervised
  setting.
\newblock In {\em 2019 IEEE 8th International Workshop on Computational
  Advances in Multi-Sensor Adaptive Processing (CAMSAP)}, pages 639--643. IEEE,
  2019.

\bibitem{bishop2006pattern}
Christopher~M Bishop and Nasser~M Nasrabadi.
\newblock {\em Pattern recognition and machine learning}, volume~4.
\newblock Springer, 2006.

\bibitem{tanaka2013statistical}
Toshiyuki Tanaka.
\newblock Statistical-mechanics analysis of {Gaussian} labeled-unlabeled
  classification problems.
\newblock In {\em Journal of Physics: Conference Series}, volume 473, page
  012001. IOP Publishing, 2013.

\bibitem{permuter2006study}
Haim Permuter, Joseph Francos, and Ian Jermyn.
\newblock A study of {Gaussian} mixture models of color and texture features
  for image classification and segmentation.
\newblock {\em Pattern recognition}, 39(4):695--706, 2006.

\bibitem{mai2019high}
Xiaoyi Mai and Zhenyu Liao.
\newblock High dimensional classification via empirical risk minimization:
  Improvements and optimality.
\newblock {\em arXiv preprint arXiv:1905.13742}, 2019.

\bibitem{mignacco2020role}
Francesca Mignacco, Florent Krzakala, Yue Lu, Pierfrancesco Urbani, and Lenka
  Zdeborova.
\newblock The role of regularization in classification of high-dimensional
  noisy {Gaussian} mixture.
\newblock In {\em International conference on machine learning}, pages
  6874--6883. PMLR, 2020.

\bibitem{sur2019modern}
Pragya Sur and Emmanuel~J Cand{\`e}s.
\newblock A modern maximum-likelihood theory for high-dimensional logistic
  regression.
\newblock {\em Proceedings of the National Academy of Sciences},
  116(29):14516--14525, 2019.

\bibitem{oliver2018realistic}
Avital Oliver, Augustus Odena, Colin~A Raffel, Ekin~Dogus Cubuk, and Ian
  Goodfellow.
\newblock Realistic evaluation of deep semi-supervised learning algorithms.
\newblock {\em Advances in neural information processing systems}, 31, 2018.

\bibitem{libbrecht2015machine}
Maxwell~W Libbrecht and William~Stafford Noble.
\newblock Machine learning applications in genetics and genomics.
\newblock {\em Nature Reviews Genetics}, 16(6):321--332, 2015.

\bibitem{belkin2019reconciling}
Mikhail Belkin, Daniel Hsu, Siyuan Ma, and Soumik Mandal.
\newblock Reconciling modern machine-learning practice and the classical
  bias--variance trade-off.
\newblock {\em Proceedings of the National Academy of Sciences},
  116(32):15849--15854, 2019.

\bibitem{mei2022generalization}
Song Mei and Andrea Montanari.
\newblock The generalization error of random features regression: Precise
  asymptotics and the double descent curve.
\newblock {\em Communications on Pure and Applied Mathematics}, 75(4):667--766,
  2022.

\bibitem{gerace2022gaussian}
Federica Gerace, Florent Krzakala, Bruno Loureiro, Ludovic Stephan, and Lenka
  Zdeborov{\'a}.
\newblock Gaussian universality of linear classifiers with random labels in
  high-dimension.
\newblock {\em arXiv preprint arXiv:2205.13303}, 2022.

\bibitem{montanari2022universality}
Andrea Montanari and Basil~N Saeed.
\newblock Universality of empirical risk minimization.
\newblock In {\em Conference on Learning Theory}, pages 4310--4312. PMLR, 2022.

\bibitem{zhang2013non}
Pan Zhang, Florent Krzakala, Marc M{\'e}zard, and Lenka Zdeborov{\'a}.
\newblock Non-adaptive pooling strategies for detection of rare faulty items.
\newblock In {\em 2013 IEEE International Conference on Communications
  Workshops (ICC)}, pages 1409--1414. IEEE, 2013.

\bibitem{saade2016clustering}
Alaa Saade, Marc Lelarge, Florent Krzakala, and Lenka Zdeborov{\'a}.
\newblock Clustering from sparse pairwise measurements.
\newblock In {\em 2016 IEEE International Symposium on Information Theory
  (ISIT)}, pages 780--784. IEEE, 2016.

\bibitem{aubin2020generalization}
Benjamin Aubin, Florent Krzakala, Yue Lu, and Lenka Zdeborov{\'a}.
\newblock Generalization error in high-dimensional perceptrons: {Approaching}
  {Bayes} error with convex optimization.
\newblock {\em Advances in Neural Information Processing Systems},
  33:12199--12210, 2020.

\bibitem{bean2013optimal}
Derek Bean, Peter~J Bickel, Noureddine El~Karoui, and Bin Yu.
\newblock Optimal {M}-estimation in high-dimensional regression.
\newblock {\em Proceedings of the National Academy of Sciences},
  110(36):14563--14568, 2013.

\bibitem{donoho2016high}
David Donoho and Andrea Montanari.
\newblock High dimensional robust {M}-estimation: Asymptotic variance via
  approximate message passing.
\newblock {\em Probability Theory and Related Fields}, 166:935--969, 2016.

\bibitem{advani2016equivalence}
Madhu Advani and Surya Ganguli.
\newblock An equivalence between high dimensional {Bayes} optimal inference and
  {M}-estimation.
\newblock {\em Advances in Neural Information Processing Systems}, 29, 2016.

\bibitem{thrampoulidis2018precise}
Christos Thrampoulidis, Ehsan Abbasi, and Babak Hassibi.
\newblock Precise error analysis of regularized {M}-estimators in high
  dimensions.
\newblock {\em IEEE Transactions on Information Theory}, 64(8):5592--5628,
  2018.

\bibitem{montanari2012graphical}
Andrea Montanari, YC~Eldar, and G~Kutyniok.
\newblock Graphical models concepts in compressed sensing.
\newblock {\em Compressed Sensing}, pages 394--438, 2012.

\bibitem{donoho2009message}
David~L Donoho, Arian Maleki, and Andrea Montanari.
\newblock Message-passing algorithms for compressed sensing.
\newblock {\em Proceedings of the National Academy of Sciences},
  106(45):18914--18919, 2009.

\bibitem{donoho2010message}
David~L Donoho, Arian Maleki, and Andrea Montanari.
\newblock Message passing algorithms for compressed sensing:
  {I\hspace{-1.2pt}I.} {Analysis} and validation.
\newblock In {\em 2010 IEEE Information Theory Workshop on Information Theory
  (ITW 2010, Cairo)}, pages 1--5. IEEE, 2010.

\bibitem{bayati2011dynamics}
Mohsen Bayati and Andrea Montanari.
\newblock The dynamics of message passing on dense graphs, with applications to
  compressed sensing.
\newblock {\em IEEE Transactions on Information Theory}, 57(2):764--785, 2011.

\bibitem{matsushita2013low}
Ryosuke Matsushita and Toshiyuki Tanaka.
\newblock Low-rank matrix reconstruction and clustering via approximate message
  passing.
\newblock {\em Advances in Neural Information Processing Systems}, 26, 2013.

\bibitem{deshpande2014information}
Yash Deshpande and Andrea Montanari.
\newblock Information-theoretically optimal sparse {PCA}.
\newblock In {\em 2014 IEEE International Symposium on Information Theory},
  pages 2197--2201. IEEE, 2014.

\bibitem{celentano2020estimation}
Michael Celentano, Andrea Montanari, and Yuchen Wu.
\newblock The estimation error of general first order methods.
\newblock In {\em Conference on Learning Theory}, pages 1078--1141. PMLR, 2020.

\bibitem{yedidia2003understanding}
Jonathan~S Yedidia, William~T Freeman, Yair Weiss, et~al.
\newblock Understanding belief propagation and its generalizations.
\newblock {\em Exploring artificial intelligence in the new millennium},
  8(236-239):0018--9448, 2003.

\bibitem{kabashima2003cdma}
Yoshiyuki Kabashima.
\newblock A {CDMA} multiuser detection algorithm on the basis of belief
  propagation.
\newblock {\em Journal of Physics A: Mathematical and General}, 36(43):11111,
  2003.

\bibitem{tanaka2005approximate}
Toshiyuki Tanaka and Masato Okada.
\newblock Approximate belief propagation, density evolution, and statistical
  neurodynamics for {CDMA} multiuser detection.
\newblock {\em IEEE Transactions on Information Theory}, 51(2):700--706, 2005.

\bibitem{Thouless01031977}
P.~W.~Anderson D.~J.~Thouless and R.~G. Palmer.
\newblock Solution of ‘{Solvable} model of a spin glass’.
\newblock {\em The Philosophical Magazine: A Journal of Theoretical
  Experimental and Applied Physics}, 35(3):593--601, 1977.

\bibitem{mezard1987spin}
Marc M{\'e}zard, Giorgio Parisi, and Miguel~Angel Virasoro.
\newblock {\em Spin glass theory and beyond: {An} {Introduction} to the
  {Replica} {Method} and {Its} {Applications}}, volume~9.
\newblock World Scientific Publishing Company, 1987.

\bibitem{richardson2001design}
Thomas~J Richardson, Mohammad~Amin Shokrollahi, and R{\"u}diger~L Urbanke.
\newblock Design of capacity-approaching irregular low-density parity-check
  codes.
\newblock {\em IEEE transactions on information theory}, 47(2):619--637, 2001.

\bibitem{barkai1994statistical}
N~Barkai and Haim Sompolinsky.
\newblock Statistical mechanics of the maximum-likelihood density estimation.
\newblock {\em Physical Review E}, 50(3):1766, 1994.

\bibitem{lesieur2016phase}
Thibault Lesieur, Caterina De~Bacco, Jess Banks, Florent Krzakala, Cris Moore,
  and Lenka Zdeborov{\'a}.
\newblock Phase transitions and optimal algorithms in high-dimensional
  {Gaussian} mixture clustering.
\newblock In {\em 2016 54th Annual Allerton Conference on Communication,
  Control, and Computing (Allerton)}, pages 601--608. IEEE, 2016.

\bibitem{thrampoulidis2020theoretical}
Christos Thrampoulidis, Samet Oymak, and Mahdi Soltanolkotabi.
\newblock Theoretical insights into multiclass classification: A
  high-dimensional asymptotic view.
\newblock {\em Advances in Neural Information Processing Systems},
  33:8907--8920, 2020.

\bibitem{loureiro2021learning}
Bruno Loureiro, Gabriele Sicuro, C{\'e}dric Gerbelot, Alessandro Pacco, Florent
  Krzakala, and Lenka Zdeborov{\'a}.
\newblock Learning {Gaussian} mixtures with generalized linear models:
  {Precise} asymptotics in high-dimensions.
\newblock {\em Advances in Neural Information Processing Systems},
  34:10144--10157, 2021.

\bibitem{takahashi2024role}
Takashi Takahashi.
\newblock The role of pseudo-labels in self-training linear classifiers on
  high-dimensional {Gaussian} mixture data.
\newblock {\em arXiv preprint arXiv:2205.07739}, 2024.

\bibitem{jordan1999introduction}
Michael~I Jordan, Zoubin Ghahramani, Tommi~S Jaakkola, and Lawrence~K Saul.
\newblock An introduction to variational methods for graphical models.
\newblock {\em Machine learning}, 37:183--233, 1999.

\bibitem{wainwright2008graphical}
Martin~J Wainwright, Michael~I Jordan, et~al.
\newblock Graphical models, exponential families, and variational inference.
\newblock {\em Foundations and Trends{\textregistered} in Machine Learning},
  1(1--2):1--305, 2008.

\bibitem{kschischang2001factor}
Frank~R Kschischang, Brendan~J Frey, and H-A Loeliger.
\newblock Factor graphs and the sum-product algorithm.
\newblock {\em IEEE Transactions on information theory}, 47(2):498--519, 2001.

\bibitem{rose1990statistical}
Kenneth Rose, Eitan Gurewitz, and Geoffrey~C Fox.
\newblock Statistical mechanics and phase transitions in clustering.
\newblock {\em Physical review letters}, 65(8):945, 1990.

\bibitem{biehl1993statistical}
M~Biehl and A~Mietzner.
\newblock Statistical mechanics of unsupervised learning.
\newblock {\em Europhysics Letters}, 24(5):421, 1993.

\bibitem{barkai1993scaling}
N~Barkai, Hyunjune~Sebastian Seung, and Haim Sompolinsky.
\newblock Scaling laws in learning of classification tasks.
\newblock {\em Physical review letters}, 70(20):3167, 1993.

\bibitem{watkin1994optimal}
TLH Watkin and J-P Nadal.
\newblock Optimal unsupervised learning.
\newblock {\em Journal of Physics A: Mathematical and General}, 27(6):1899,
  1994.

\bibitem{iba1999nishimori}
Yukito Iba.
\newblock The {Nishimori} line and {Bayesian} statistics.
\newblock {\em Journal of Physics A: Mathematical and General}, 32(21):3875,
  1999.

\bibitem{nishimori2001statistical}
Hidetoshi Nishimori.
\newblock {\em Statistical physics of spin glasses and information processing:
  an introduction}.
\newblock Number 111. Clarendon Press, 2001.

\bibitem{tanaka2002statistical}
Toshiyuki Tanaka.
\newblock A statistical-mechanics approach to large-system analysis of {CDMA}
  multiuser detectors.
\newblock {\em IEEE Transactions on Information theory}, 48(11):2888--2910,
  2002.

\bibitem{rangan2011generalized}
Sundeep Rangan.
\newblock Generalized approximate message passing for estimation with random
  linear mixing.
\newblock In {\em 2011 IEEE International Symposium on Information Theory
  Proceedings}, pages 2168--2172. IEEE, 2011.

\bibitem{sakata2018approximate}
Ayaka Sakata and Yingying Xu.
\newblock Approximate message passing for nonconvex sparse regularization with
  stability and asymptotic analysis.
\newblock {\em Journal of Statistical Mechanics: Theory and Experiment},
  2018(3):033404, 2018.

\bibitem{takahashi2022macroscopic}
Takashi Takahashi and Yoshiyuki Kabashima.
\newblock Macroscopic analysis of vector approximate message passing in a
  model-mismatched setting.
\newblock {\em IEEE Transactions on Information Theory}, 68(8):5579--5600,
  2022.

\end{thebibliography}

\end{document}